\documentclass[10pt,twocolumn,letterpaper]{article}

\usepackage{iccv}
\usepackage{times}
\usepackage{epsfig}
\usepackage{graphicx}
\usepackage{amsmath}
\usepackage{amssymb}

\usepackage{subcaption}
\usepackage{mathrsfs}
\usepackage{booktabs}
\usepackage{multirow}
\usepackage{array}
\usepackage{comment}
\usepackage{caption}
\usepackage{rotate}
\usepackage[table]{xcolor}

\usepackage{enumitem}

\makeatletter
\renewcommand{\paragraph}{%
	\@startsection{paragraph}{4}%
	{\z@}{2.5ex \@plus 1ex \@minus .2ex}{-1em}%
	{\normalfont\normalsize\bfseries}%
}
\makeatother

\usepackage[pagebackref=true,breaklinks=true,letterpaper=true,colorlinks,bookmarks=false]{hyperref}
\iccvfinalcopy % *** Uncomment this line for the final submission

\def\iccvPaperID{7514} % *** Enter the ICCV Paper ID here

\begin{document}

%%%%%%%%% TITLE
\title{Adaptive Convolutions with Per-pixel Dynamic Filter Atom}

\author{Ze Wang$^1$, Zichen Miao$^1$, Jun Hu$^2$, and Qiang Qiu$^1$\\
Purdue University$^1$ \quad Facebook$^2$\\
{\tt\small \{zewang, miaoz, qqiu\}@purdue.edu \quad junhu2@fb.com}
}

\maketitle
%\ificcvfinal\thispagestyle{empty}\fi

%%%%%%%%% ABSTRACT
\begin{abstract}
Applying feature dependent network weights have been proved to be effective in many fields. However, in practice, restricted by the enormous size of model parameters and memory footprints, scalable and versatile dynamic convolutions with per-pixel adapted filters are yet to be fully explored. In this paper, we address this challenge by decomposing filters, adapted to each spatial position, over dynamic filter atoms generated by a light-weight network from local features. Adaptive receptive fields can be supported by further representing each filter atom over sets of pre-fixed multi-scale bases. As plug-and-play replacements to convolutional layers, the introduced adaptive convolutions with per-pixel dynamic atoms enable explicit modeling of intra-image variance, while avoiding heavy computation, parameters, and memory cost. Our method preserves the appealing properties of conventional convolutions as being translation-equivariant and parametrically efficient. We present experiments to show that, the proposed method delivers comparable or even better performance across tasks, and are particularly effective on handling tasks with significant intra-image variance.

\end{abstract}
\section{Introduction}
\label{intro}

\begin{figure*}[th]
	\resizebox{\linewidth}{0.15\linewidth}{
		\begin{subfigure}{0.25\linewidth}
			\vspace{8.6mm}
			\includegraphics[width=\linewidth]{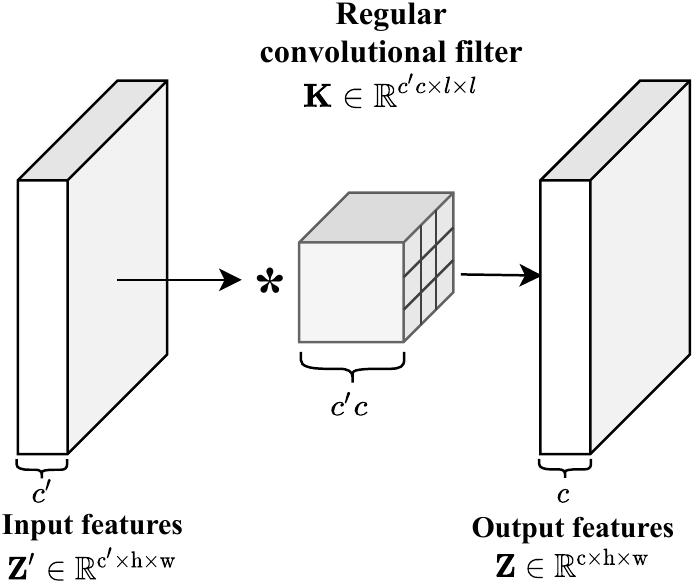}
			\caption{Standard convolutions}
			\label{fig:f0}
		\end{subfigure}
		\hspace{3mm}
		\begin{subfigure}{0.3\linewidth}
			\includegraphics[width=\linewidth]{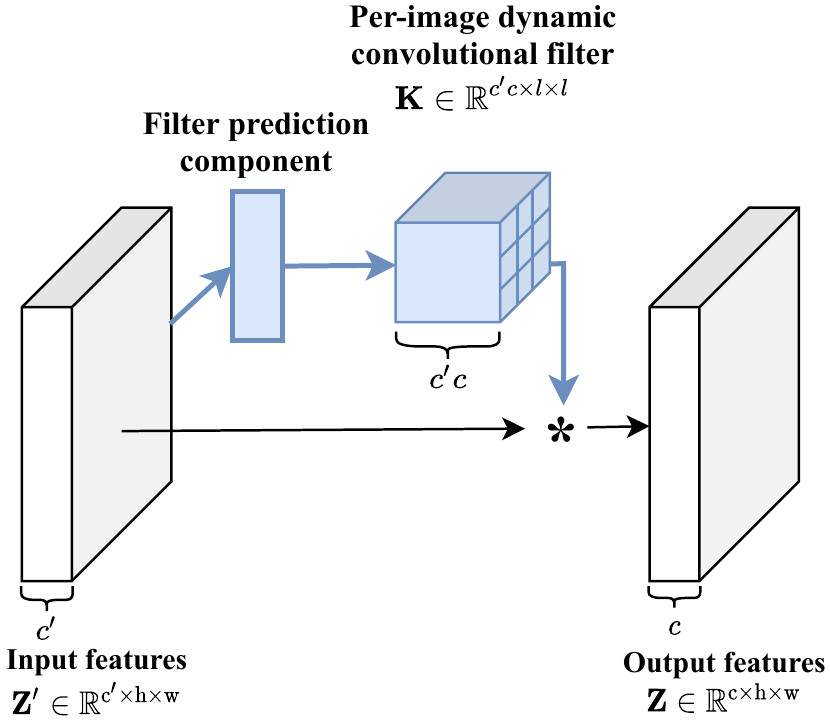}
			\caption{Per-image adaptive convolutions}
			\label{fig:f1}
		\end{subfigure}
	   \hspace{3mm}
		\begin{subfigure}{0.3\linewidth}
			\includegraphics[width=\linewidth]{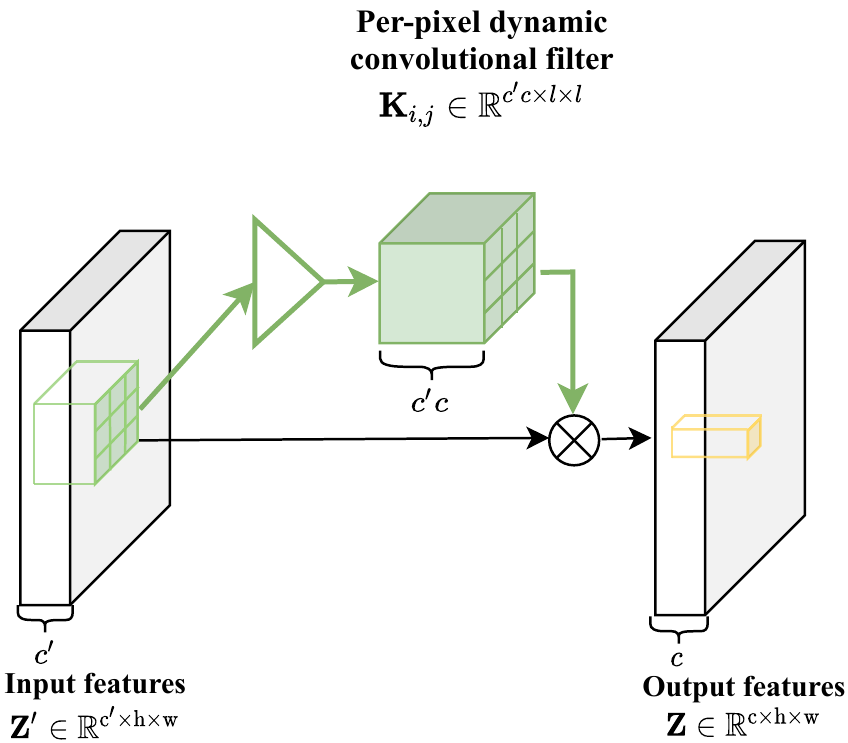}
			\caption{Per-pixel adaptive convolutions}
			\label{fig:f2}
		\end{subfigure}
}
	\caption{Convolutional layers. (a) Standard convolutions with filters shared across all samples and spatial position. (b) Per-image adaptive convolutions with per-image specific filters as in \cite{dycnn,dynamic,condconv}. (c) The proposed per-pixel adaptive convolutions dynamically generate filters conditioned on local feature patches, and explicitly model intra-image variance using per-pixel specific filters.}
	\label{fig:illus} 
\end{figure*}

The idea of data or context dependent network weights have long been studied in the research of neural networks. 
Many concepts, like fast weight \cite{fastweight,gated,schlag2021learning} and dynamic plasticity \cite{miconi2020backpropamine,diffpla} are developed to explicitly model the evolution of model parameters, and have demonstrated improved performance on sequential data. 
In \cite{fastweight,gated,schlag2021learning}, the parameters in these networks can be divided into two groups: slow weights that are learned through training with gradient descent, and fast weights that are generated on-the-fly depending on both slow weights and observed data. 

In recent years, similar idea has been extended to dynamic convolutions. 
Standard convolutions as in Figure~\ref{fig:f0} uses shared filters across all samples and all spatial position. Dynamic convolutions, as in Figure~\ref{fig:f1}, allow convolutional filters to be adapted to data in one-shot (as opposite to evolving through sequential observations). Dynamic filter networks (DFN) \cite{dynamic}, conditionally parameterized convolutions (CondConv) \cite{condconv}, dynamic convolutions with attention (DY-CNN) \cite{dycnn} are introduced to allow convolutional filters to be adapted to the current observed input, and explicitly modeling the inter-sample variance among images. 
While improvements on certain tasks have been shown, the flexibility comes at the cost of extra parameter and computation \cite{dycnn,condconv}, and sacrificing the translation equivariant property of CNNs \cite{dynamic}. More importantly, it is practically infeasible to extend such methods from per-image adapted filters to per-pixel adapted filters due to the prohibitive memory footprints of applying per-pixel adapted high-dimension filters as we will show in Section~\ref{cost}. 

In this paper, we enable CNNs with per-pixel adaptive convolutions as illustrated in Figure~\ref{fig:f2}, at any network layers to better model intra-image variance.
We introduce \textit{Adaptive Convolutions with Dynamic Atoms (ACDA)}, a versatile and scalable convolutional layer that allows per-pixel specific filters to be adaptively generated from each local input feature patch across spatial position.
To remedy the prohibitively high cost on generating and applying per-pixel adaptive filters in high dimensions, we decompose filters over dynamically generated low-dimensional filter atoms at each spatial location.
The adaptive filters can now be reconstructed by linearly combining these per-pixel specific dynamic atoms with cross-location shared compositional coefficients, as illustrated in Figure~\ref{fig:main}.
Most importantly, the decomposition enables a fast two-layer implementation of the adaptive convolutions as shown in Figure~\ref{fig:mainp}, which reduces the prohibitive computation and memory footprints of applying per-pixel specific convolutions to a level that matches standard convolutions, allowing our method becomes a versatile replacement to standard convolutions across layers in any CNNs.

To achieve adaptive receptive fields, we further decompose each filter atom over sets of multi-scale pre-fixed atom bases as in Figure~\ref{fig:atom}, for a two-level filter decomposition.
Now, instead of directly generating atoms, only per-pixel basis coefficients are required to be generated. 
The multi-scale atom bases and the generated basis coefficients reconstruct dynamic atoms, and allow the receptive field at each spatial position to be selectively decided from the local features. 
Meanwhile the adaptive filters are effectively regularized by the prefixed atom bases, which accelerates the learning of the large-size adaptive filter generation.
Importantly, our approach maintains and even reduces parameters and computation, and preserves appealing properties of CNNs including translation equivariant and weight sharing. 

We show empirically that, our approach can work as plug-and-play replacements to standard convolutional layers.
We demonstrate the effectiveness of the proposed method using image classification,
crowd counting, and real-world image restorations as example tasks that require handling significant intra-image variance.

\section{Related Work}

%\paragraph{Parameter predictions.}
\noindent \textbf{Parameter predictions.}
Instead of directly training parameters for deep networks, parameter predictions are discussed under various motivations.
Parameters in Hypernetworks \cite{hypernets} are formulated as the outputs of a network, which achieves parameter compression.
BasisGAN \cite{basisgan} adopts bases generators to model the space of parameters and achieve stochastic conditional image generations.
Predicting networks weights given samples \cite{parafromacts,probmeta} is widely used for few shot image classification. 

%\paragraph{Dynamic convolutions.}
\noindent \textbf{Dynamic convolutions.}
The idea of relaxing the strict weight sharing across spatial position of convolutions has been discussed in works like locally connected networks \cite{local} and dynamic filter networks \cite{dynamic}.
And applying locally feature dependent convolutional filters has demonstrated effectiveness on real-world image restorations \cite{realsr,kpn,unified}, where degradation models are non-uniformed across spatial position.
However, restricted by practical costs, the usages of adaptive convolutions are usually simplified based on certain tasks \cite{su2019pixel} and assumptions, e.g., the dynamic convolutions are applied only to the image domain \cite{realsr,kpn,jia2017super}, or shared across channels in deep features \cite{unified}, 
which prevent the methods from being extended to universal applications.
Dynamic convolution with attention is introduced in \cite{dycnn} by assembling multiple kernels through attention mechanism. The flexibility is restricted by the fixed number of filters.
Deformable convolutions \cite{deforme,zhu2019deformable} allow adapted kernel shape while fixing the values in kernels, thus are orthogonal to our efforts, and can potentially work together with the proposed framework of adaptive convolutions with dynamic filter atoms.

\newcommand{\feat}{\mathbf{Z}}
\newcommand{\conv}{\mathcal{F}}
\newcommand{\R}{\mathbb{R}}
\newcommand{\N}{\mathcal{N}}
\newcommand{\atom}{\mathbf{D}}
\newcommand{\coef}{\mathbf{A}}
\newcommand{\filter}{\mathbf{K}}
\newcommand{\bases}{\boldsymbol{\psi}}
\newcommand{\intercoef}{\boldsymbol{\alpha}}

\section{Method}
In this section, we first present 
per-pixel adaptive filter generations with filter decomposition over dynamic atoms, and then show a practical two-layer implementation to address the prohibitive memory footprints and accelerate the speed.
We then decompose filter atoms over sets of prefixed multi-scale bases to further enable dynamic receptive fields while maintaining parameter size.
We end this section with illustrative toy examples to show the advantages of our method on addressing intra-image variance, and preserving the translation equivariance property.

\subsection{Preliminary}
We denote scalars, vectors and tensors with lower-case, bold lower-case, and bold upper-case letters, e.g., $n$, $\mathbf{x}$, $\mathbf{X}$, respectively.
We use $\feat^\prime \in \R^{c^\prime \times h \times w}$ and $\feat \in \R^{c \times h \times w}$ to denote the input and output features of a typical convolutional layer with $c^\prime$ input and $c$ output channels, each of which has a spatial resolution of $h \times w$. 
The corresponding convolutional filter is denoted as $\filter \in \R^{c \times c^{\prime} \times l \times l}$, where $l$ is the kernel size.
We use $[i, j]$ to denote the spatial position of a feature map at the $i$-th row and the $j$-th column, and $\mathbf{z}_{i, j} = \mathbf{Z}[i,j] \in \R^{c}$ is the feature vector in $\feat$ at position $[i, j]$.
We use $\N_{\feat[i,j]}^{\delta}$ to denote the size-$\delta$ neighborhood region of $\feat[i, j]$, i.e., $\N_{\feat[i,j]}^{\delta} = \{\feat[i-u, j-v]\}_{-\delta \leq u \leq \delta, -\delta \leq v \leq \delta}$.
In the following discussions, without loss of generality, we assume convolutions with a stride of 1 and padding for consistent input and output resolutions.

\begin{figure}[t]
	\centering
	\includegraphics[width=\linewidth]{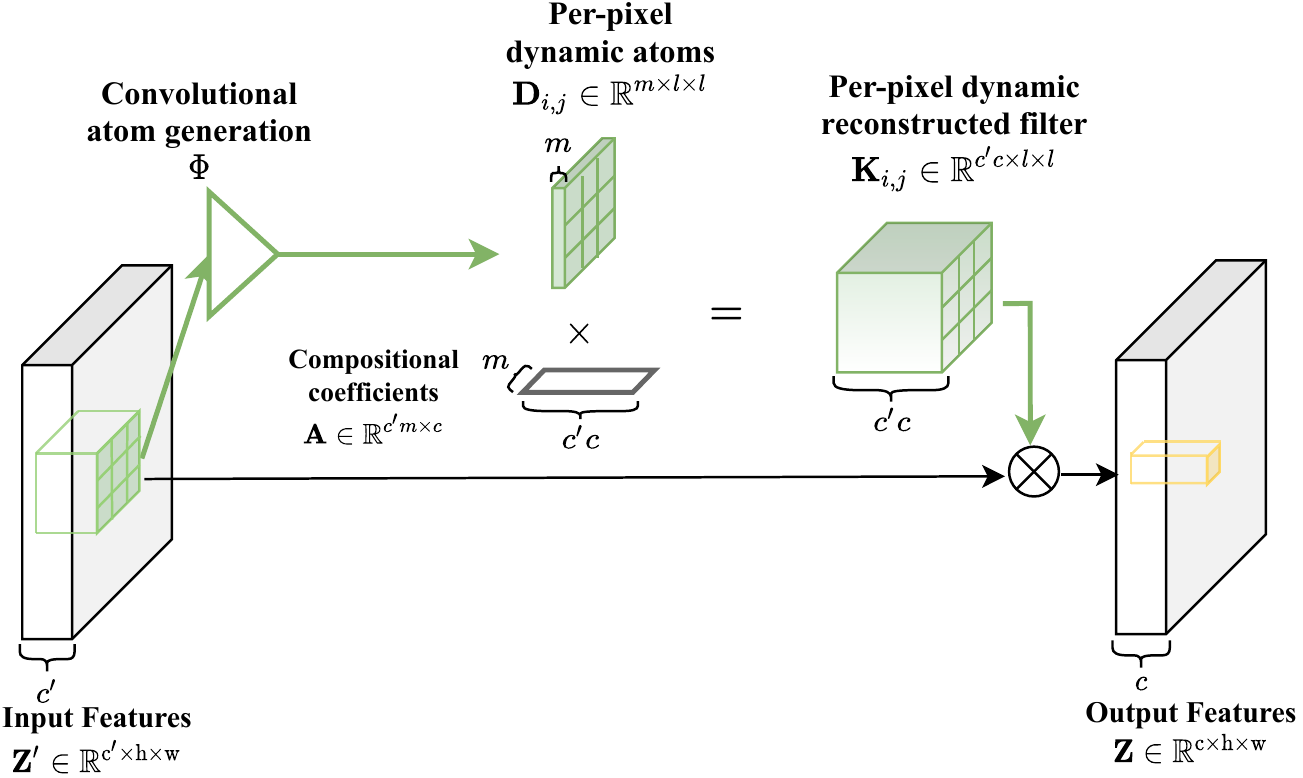}
	\caption{Adaptive convolution with per-pixel dynamic atoms. An atom generation network convolves with the input to generate the per-pixel dynamic filter atoms $\atom_{i,j}$, which then multiply with the cross-spatial shared compositional coefficients $\coef$ to reconstruct full-size adaptive filters $\filter_{i, j}$ across position. The predicted filters are then applied back to local regions centered around each spatial position.}
	\label{fig:main}
\end{figure}

\subsection{Adaptive Filter Generation}
\label{acda}
Conventional convolutional layers, as in Figure~\ref{fig:f0}, learn filters that are shared both across all spatial position of an input feature map, and among input features from different input images.
Formally, typical convolutions in CNNs at a specific position can be expressed as:
\begin{equation}
	\begin{aligned}
		\feat[i, j] & = \mathcal{F}(\N^{\delta}_{\feat[i, j]}; \filter)\\
		& = \sum_{u = -\delta}^{\delta}\sum_{v = -\delta}^{\delta} \filter[u, v] \cdot \feat[i-u, j-v],
	\end{aligned}
\end{equation}
where we use $\mathcal{F}$ to denote the convolution operation, and $\delta = \lfloor l / 2 \rfloor$. The convolutional filter $\filter$ is shared across spatial position of $\feat^\prime$, and remains fixed after training.

We aim at adaptively generating per-pixel convolutional filters conditioned on the corresponding neighborhood regions of input features, as illustrated in Figure~\ref{fig:f2}, to better handle intra-image variance.
Formally, the adaptive convolution is expressed as:
\begin{equation}
	\label{eq:dfn}
	\begin{aligned}
		\feat[i, j] & = \mathcal{F}(\N^{\delta}_{\feat[i,j]}; \filter_{i, j})\\
		& = \sum_{u = -\delta}^{\delta}\sum_{v = -\delta}^{\delta} \filter_{i,j}[u, v] \cdot \feat[i-u, j-v],
	\end{aligned}
\end{equation}
where $\filter$ here becomes a collection of $h \times w$ local filters, each of which is denoted as $\filter_{i, j}$, and generated as
\begin{equation}
	\begin{aligned}
		\filter_{i, j} = \Phi(\N_{\feat[i,j]}^{\delta^\prime}; \theta).
	\end{aligned}
\end{equation}
$\Phi$ here is the convolutional filter generation network parametrized by $\theta$, which is end-to-end trained.
Given a specific spatial position $[i, j]$ in the input features $\feat^\prime$, $\Phi$ takes as input the local region centered around $\feat[i, j]$, and predicts the local adaptive filter $\filter_{i, j} \in \R^{c \times c^\prime \times l \times l}$. The predicted filter $\filter_{i, j}$ is then applied back to a local region centered around $\feat[i, j]$, and outputs the vector at position $[i,j]$ in the output feature $\feat[i, j] \in \R^{c}$. 
Note that $\delta$ here does not necessarily equal to $\delta^\prime$, i.e., at each spatial position, the local neighborhood region fed into $\Phi$ can be either smaller or larger than the region the generated filter is applied back to.

CNNs nowadays usually have high-dimensional filters, which make direct filter generations infeasible considering the size of parameters and computation. 
Moreover, since now the convolutional filters are per-pixel specific, \textit{whose gradients need to be stored independently for backward computation}, this will lead to dramatically large memory footprints. For example, with a typical setting of $c^\prime=c = h =w =100$, and $l = 3$, a single layer will consume $26.8$GB memory for storing the per-pixel specific gradients, which is practically prohibitive. This is also the reason why existing per-sample specific adaptive convolutions \cite{dynamic,condconv,dycnn} can hardly be extended to per-pixel adaptations.

\paragraph{Filter atom decomposition.}
It is shown in \cite{qiu2018dcfnet} that a convolutional filter in a CNN can be decomposed as a linear combination of pre-fixed bases (visualized in Appendix Figure~\ref{fig:dcf}). 
We adopt filter decomposition as visualized in Figure~\ref{fig:main}, where a convolutional filter is decomposed as a linear combination of $m$ dynamic filter atoms $\atom \in \R^{m \times l \times l}$ as $\filter = \coef \atom$, $\coef \in \R^{c \times c^\prime \times m}$ is the composition coefficients.
The filter atoms at each spatial position is generated by
\begin{equation}
	\begin{aligned}
		\atom_{i, j} \in \R^{m \times l \times l} = \Phi(\N_{\feat[i,j]}^{\delta^\prime}; \theta).
	\end{aligned}
\end{equation}
We slightly abuse the notation and use $\Phi$ to denote
the convolutional atom generation network now.
After decomposition, the spatial patterns of the convolutional filters are decided by the filter atoms, which are very low-dimensional comparing to the filters.

\begin{figure}[t]
	\centering
	\includegraphics[width=\linewidth]{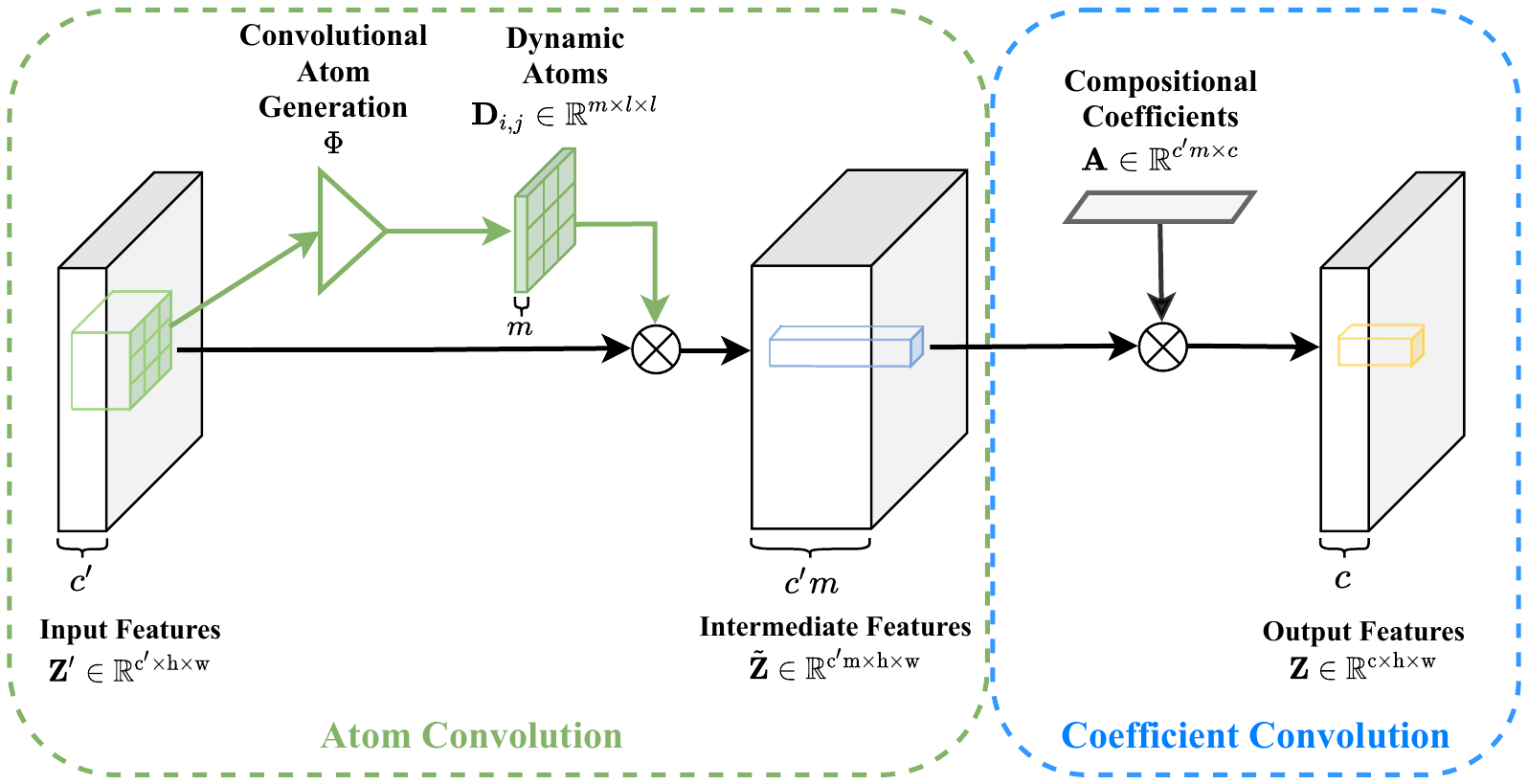}
	\caption{The efficient two-layer implementation of ACDA. The convolutional atom generation network first convolves with the input features to output dynamic atoms $\atom_{i,j}$ at each spatial position, which are then convolved spatially with the input features to output the intermediate atom outputs $\tilde{\feat}$. A linear transformation using the compositional coefficients $\coef$ is followed for the final outputs.}
	\label{fig:mainp}
\end{figure}

\paragraph{Two-layer implementation.}
Generating dynamic filter atoms of low dimensions significantly reduces the parameters and computation. 
Moreover, as visualized in Figure~\ref{fig:mainp}, the filter decomposition further allows the forward pass in (\ref{eq:dfn}) being decomposed into two convolutions, each of which involves multiplications between mild-size tensors only.
Specifically, given the generated dynamic atoms, the forward pass is now decomposed into two steps:

- First, in \textbf{atom convolution}, the input features with $c^\prime$ channels are convolved spatially only with each of the $m$ generated dynamic filter atoms, and output the intermediate features $\tilde{\feat}$ with $c^\prime m$ channels:
\begin{equation}
	\begin{aligned}
		\tilde{\feat}[i, j] \in \R^{c^\prime m} & = || \mathcal{F}(\N^{\delta}_{\feat[i,j]}; \atom_{i, j}[b])||_{b = \{1, \dots, m\}},
	\end{aligned}
\end{equation}
where $\atom_{i, j}[b]$ denotes the $b$-th generated filter atom at position $[i, j]$, and $||\cdot||_{b = \{1, \dots m\}}$ here denotes the channel-wise concatenation of the features. \\
- Second, in \textbf{coefficient convolution}, given the intermediate features $\tilde{\feat}$, the final outputs can be obtained by applying the composition coefficients $\coef \in \R^{c \times c^\prime m}$ and linearly combining the intermediate features:
\begin{equation}
	\begin{aligned}
		\feat \in \R^{c \times h \times w} = \coef \tilde{\feat}.
	\end{aligned}
\end{equation}

In practice, this step can be efficiently implemented by a $1 \times 1$ convolutions as $\coef$ now is a linear transform shared across spatial position.
Since all the involved operations are linear, this \textit{two-layer implementation} in Figure~\ref{fig:mainp} exactly equals to applying reconstructed full-size filters at each local position as in Figure~\ref{fig:main}, yet it can prevent the prohibitive memory footprints, reduce computation, and accelerate the speed as we will show in Section~\ref{cost}.

\begin{figure}
	\includegraphics[width=\linewidth]{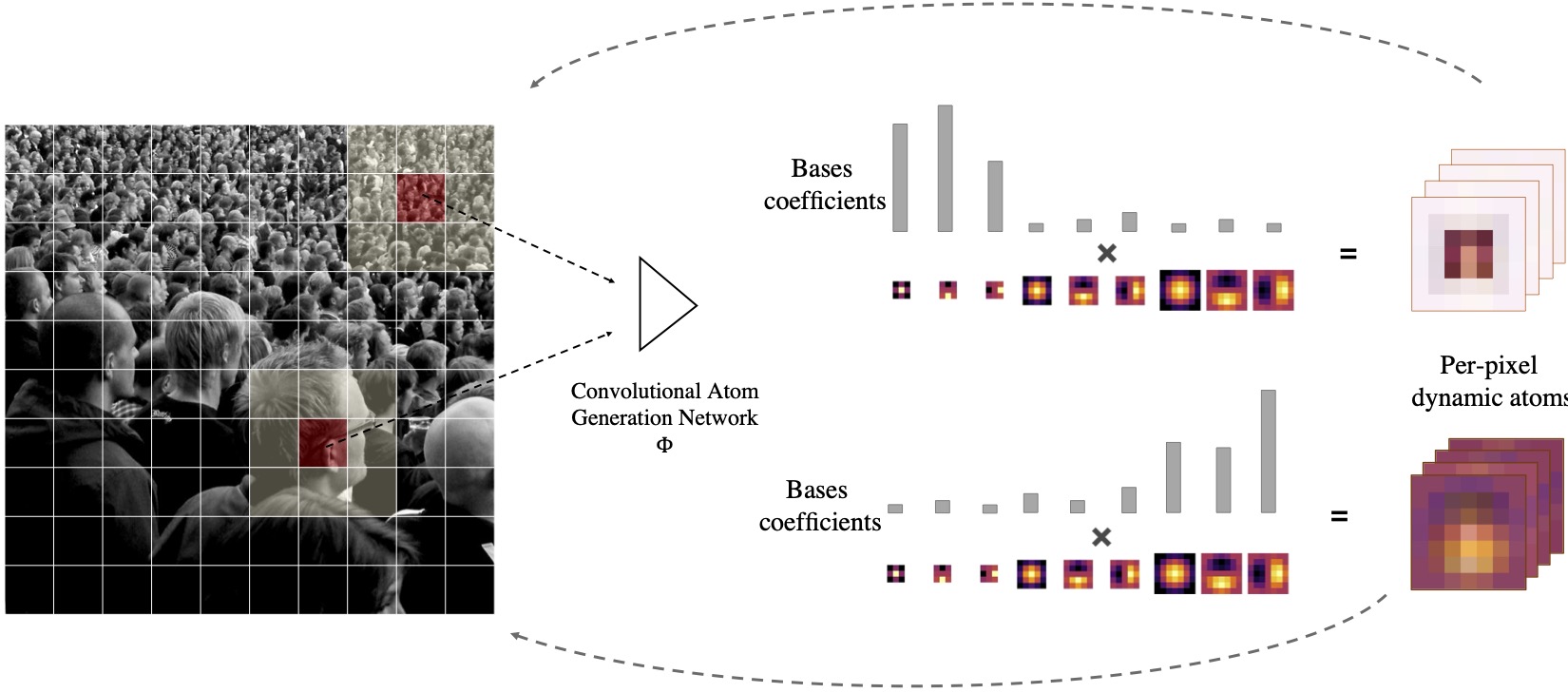}
	\caption{Atom generation with multi-scale Fourier-Bessel bases.
		The feature at target position (red points) and the neighborhood features (yellow points) are fed into the convolutional atom generation network $\Phi$. 
		At all spatial position,
		the basis coefficients $\intercoef \in \R^{m\times Sm^\prime}$ are generated, which are multiplied with $S$ sets of multi-scale Fourier-Bessel bases for the atoms $\atom_{i,j} \in \R^{m \times l \times l}$.}
	\label{fig:atom}
\end{figure}

\paragraph{Multi-scale atom bases for adaptive receptive fields.}
To the best of our knowledge, adaptive receptive fields in convolutional filters have seldom been discussed in previous works. 
Selectively deciding receptive fields at each spatial position can potentially benefit tasks with great intra-image scale variance, e.g., crowd counting, where the sizes of the concerned objects in a single image can vary significantly. 
However, the costs of generating large size dynamic atoms can grow quadratically w.r.t the kernel sizes. 
To achieve adaptive receptive fields without additional costs, we propose to further decompose the dynamic atoms over multi-scale pre-fixed bases. 
Specifically, we now decompose filter atoms over $S$ sets of pre-fixed bases, jointly denoted as $\bases = \{\bases_s \in \R^{m^\prime \times l_s \times l_s}\}_s^S$, each set contains $m^\prime$ basis elements at a certain spatial scale $l_s \times l_s$.
In practice, we adopt the multi-scale Fourier-Bessel bases as visualized in Appendix Figure~\ref{fig:fb}. As discussed in \cite{qiu2018dcfnet}, Fourier-Bessel bases can effectively regularize filters and prevent learning high-frequency noise.
As shown in Figure~\ref{fig:atom}, the convolutional atom generation network $\Phi$ now outputs the basis coefficients denoted as $\intercoef_{i,j} \in \R^{m \times Sm^\prime}$, which are multiplied with the sets of atom bases at different scales to reconstruct the dynamic filter atoms at each spatial position $\atom_{i,j} = \intercoef_{i,j} \bases$. 
The effective receptive field of $\filter_{i,j}$ is now decided by the predicted basis coefficients $\intercoef_{i, j}$, which decide the weights of atom bases $\bases$ at different scales when reconstructing dynamic filter atoms. 
Given the number of pre-fixed atom bases $m^\prime$, the atom bases allows filters with both different receptive fields and patterns to be applied across spatial position without increasing the cost of parameters.
Meanwhile, since the patterns of the filters at scales are all regularized by $\bases$, we consistently observe that the network can learn fast even with large scale, e.g., $7 \times 7$ filters being adaptively generated and applied.

\begin{figure}[h]
	\centering
	\includegraphics[width=0.9\linewidth]{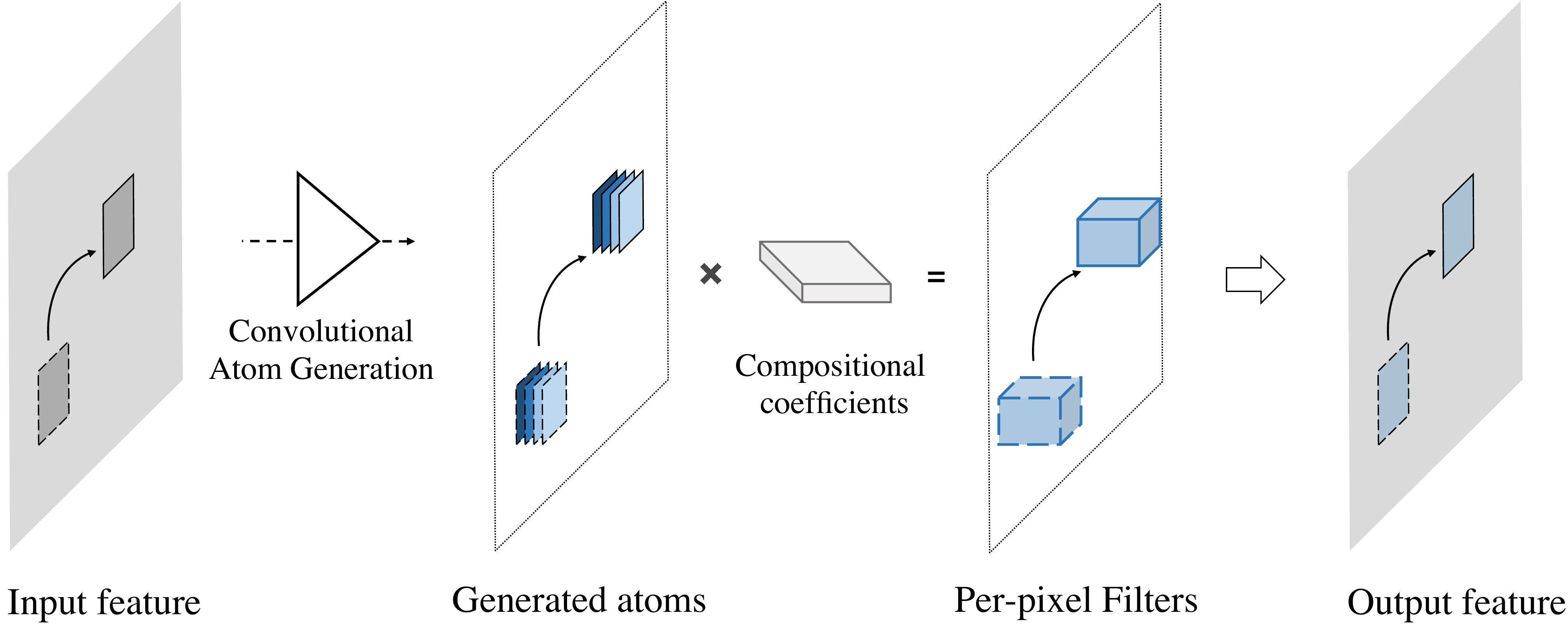}
	\caption{The translation equivariance property of the proposed method. The original and the translated position are denoted by dashed and solid lines, respectively.}
	\label{fig:shift}
\end{figure}

\paragraph{Translation equivariance.}
In our method, the convolutional atom generation network $\Phi$ and the compositional coefficients $\coef$ are both shared across spatial position. 
The \textit{weight sharing} in our method ensures that ACDA preserves the \textit{translation equivariance} property of standard convolutions. 
As illustrated in Figure~\ref{fig:shift}, a spatial translation to a local feature patch results in an equal shift to the corresponding dynamic atoms output from the convolutional atom generation network.
With the shared $\coef$, the adaptive filters and therefore the output feature patch are both exactly the spatially-shifted versions of the original spatial position.

\subsection{Toy Pattern Detection Examples}
To illustrate the advantages of the proposed method on handling intra-image variance and the translation-equivariant property preserved from standard convolutions, we present toy pattern detection examples as visualized in Figure~\ref{fig:toy}.
In Figure~\ref{fig:toy0}, we synthesize a single training sample by randomly placing pre-defined multi-scale patterns on a noisy background image. 
Specifically, we place 25 patterns at three scales, $3 \times 3$,  $5 \times 5$, and  $7 \times 7$, on a $100 \times 100$ noisy map.
The goal is to predict the groundtruth binary detection map as shown in Figure~\ref{fig:toy0}, where only the center position of the 25 patterns are marked as $1$, and $0$ otherwise. 
We train two single-layer networks, one with a single standard convolution, denoted as \textit{Conv}, and one with a single  proposed adaptive convolution layer with dynamic atoms, denoted as \textit{ACDA}. The kernel sizes of both layers are $7\times 7$.
Using the introduced synthesized input and groundtruth, both layers are trained with stochastic gradient decent (SGD) and mean square error (MSE) using learning rate $0.01$ till converge. 
As visualized in Figure~\ref{fig:toy0}, single standard convolutional layer is incapable of handling such diverse patterns within the image, and results in a high error rate. 
On the other hand, we observe that \textit{ACDA} can fast adapt to the diverse patterns, and generate adaptive filters at each position for detection. The accurate detection by \textit{ACDA} is reflected by both low error rate and appealing predicted map. 

\newcommand{\CTS}{0.32\linewidth}
\newcommand{\CHS}{\hspace{-6.0mm}}
\newcommand{\CHSS}{\hspace{-4.0mm}}

\begin{figure}[h]
	%	\centerin
	\begin{subfigure}[b]{\linewidth}
		\resizebox{\linewidth}{!}{%
			\begin{tabular}{c c c c}
				% 		\centering
				\includegraphics[width=\CTS]{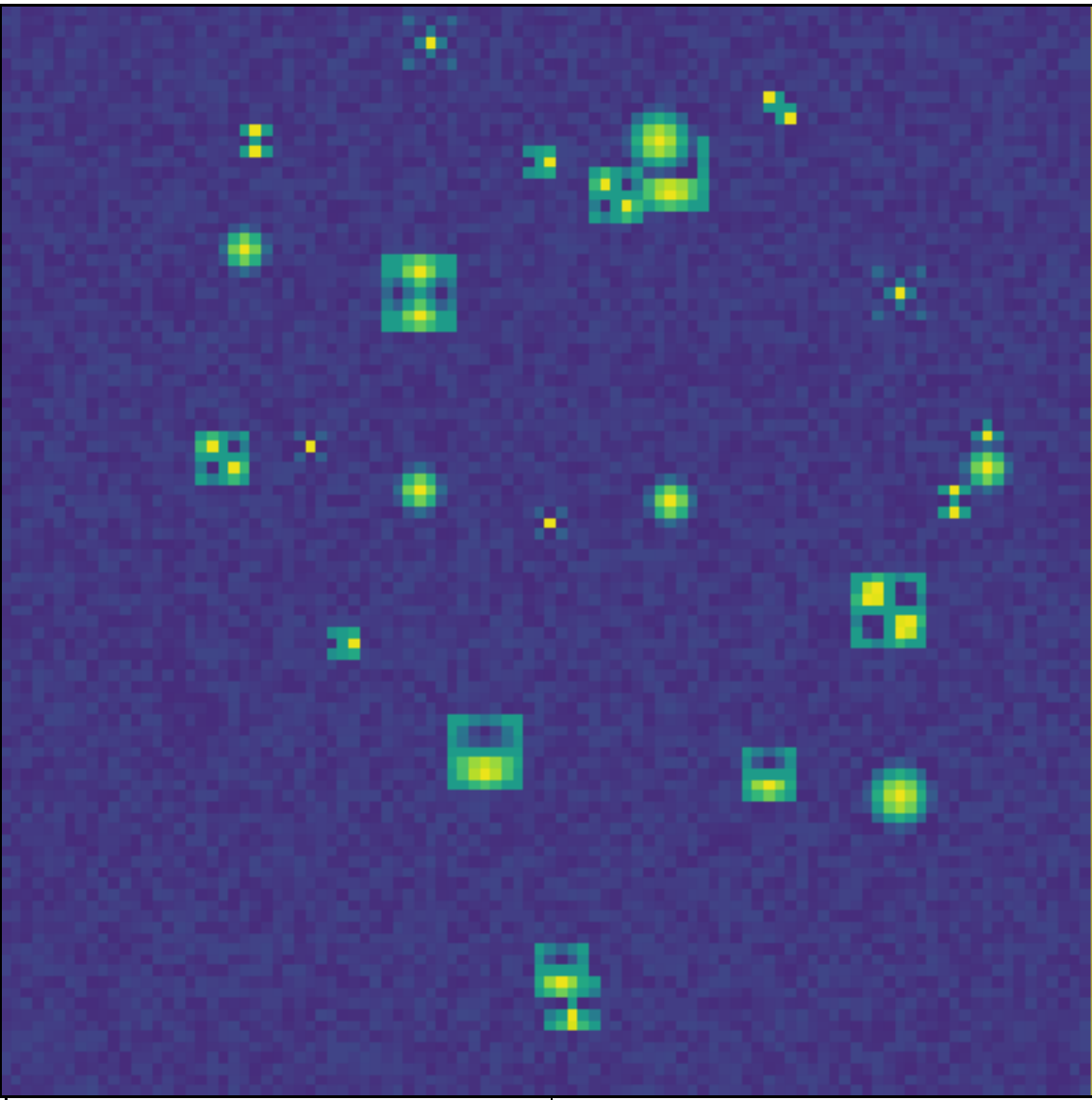}
				\CHSS
				&
				\CHSS
				\includegraphics[width=\CTS]{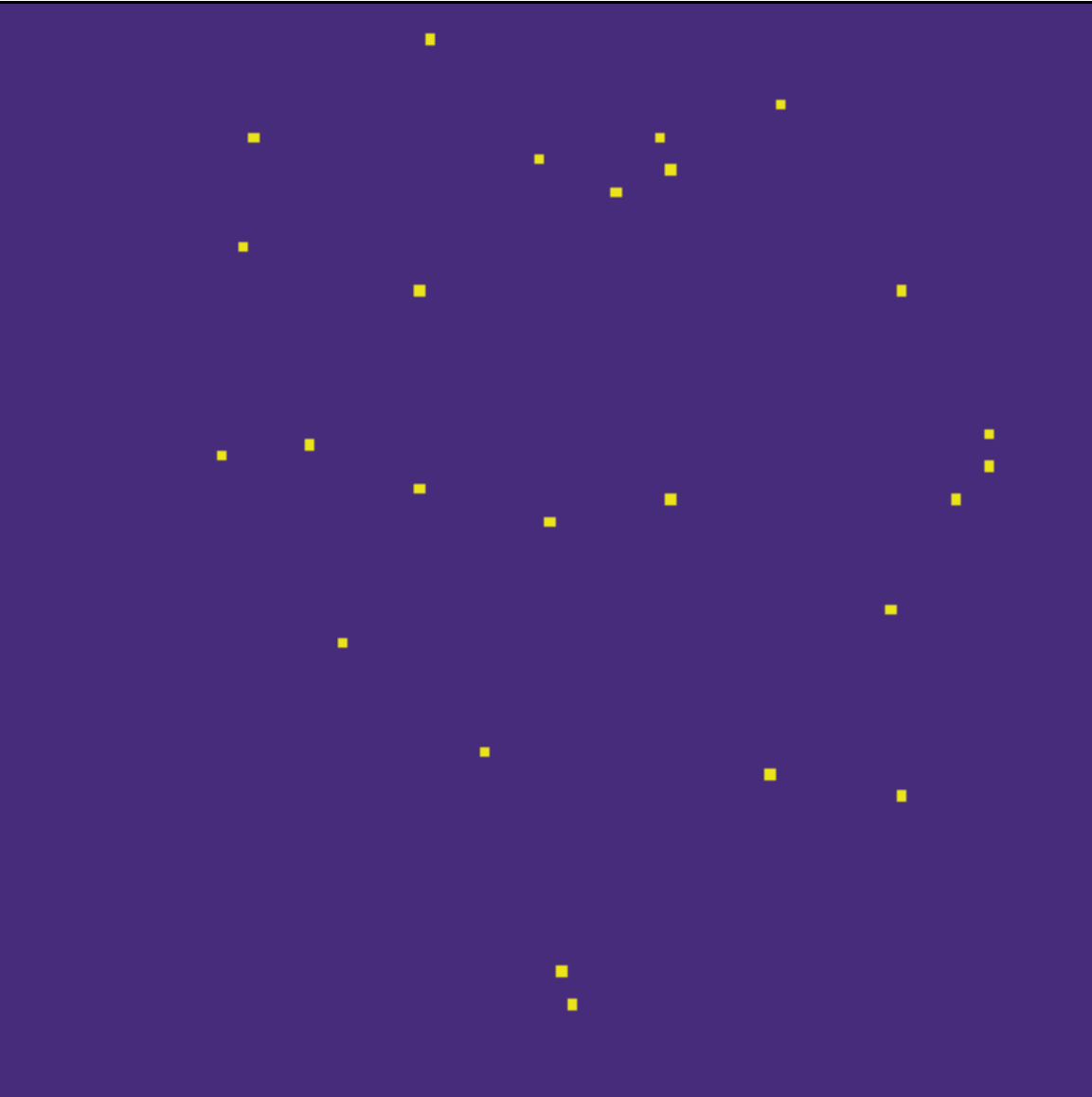}
				\CHSS
				&
				\CHSS
				\includegraphics[width=\CTS]{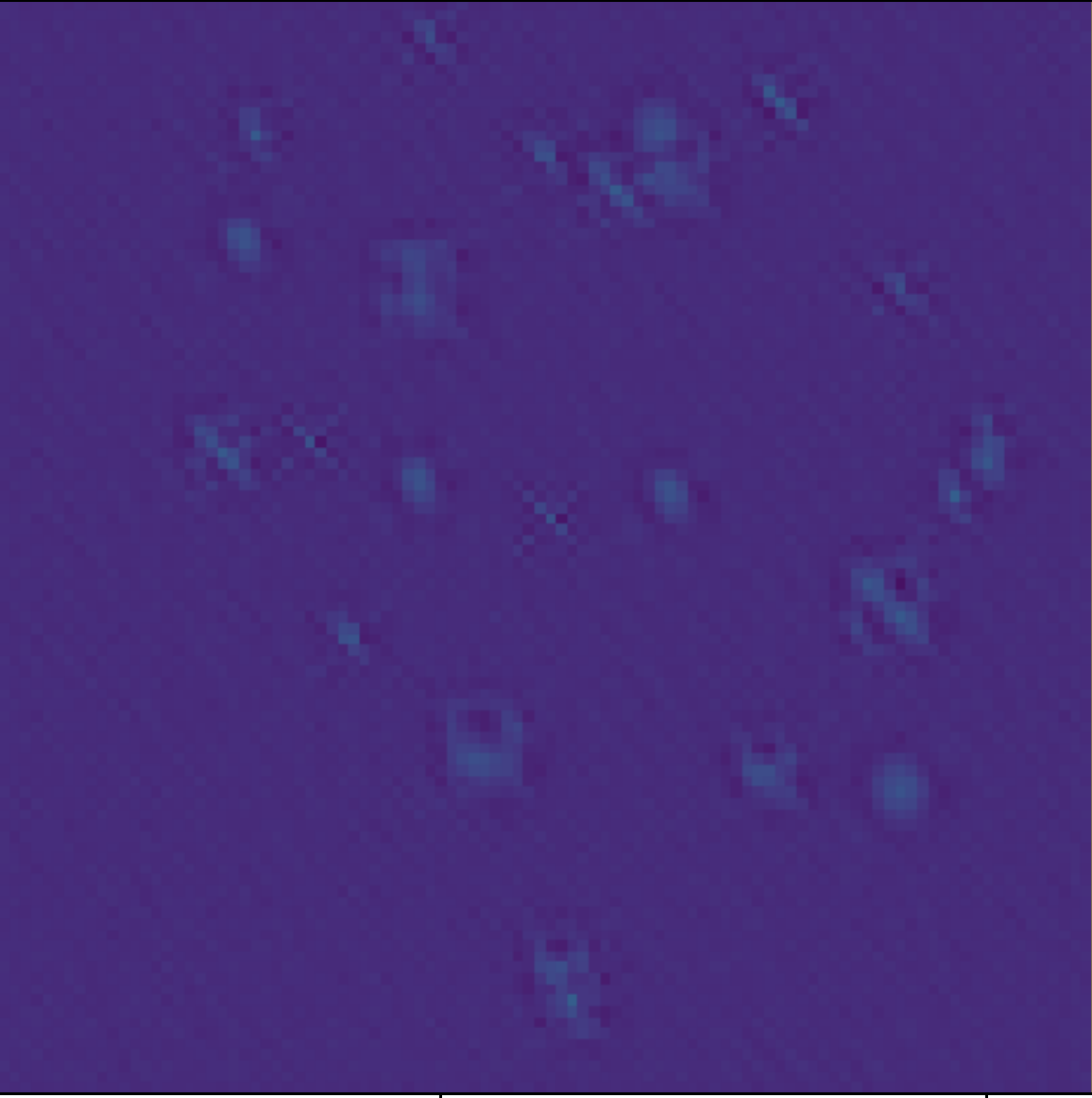}
				\CHSS
				&
				\CHSS
				\includegraphics[width=\CTS]{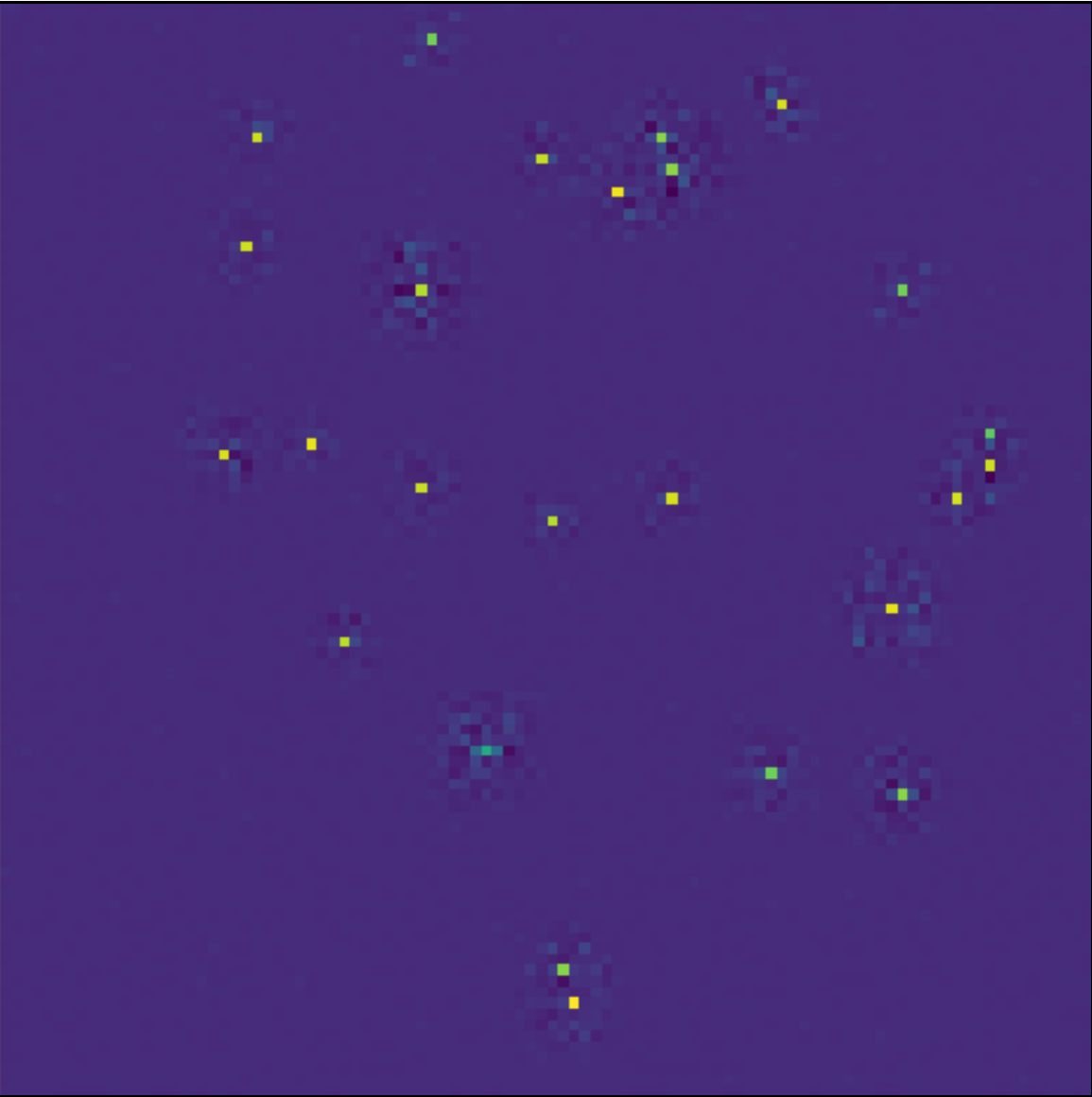}
				%    	\hspace{-2mm}
				\\
				Input & \hspace{-3mm}GT & \hspace{-3mm} \textit{Conv} output & \hspace{-3mm} \textit{ACDA} output \\
				&  & \hspace{-2mm} MSE $2.2 \times 10^{-3}$  &   \hspace{-3mm} MSE $2.3 \times 10^{-4}$ \\
		\end{tabular}}
% 		\vspace{-2mm}
		\caption{Toy pattern detection experiment. We train a single standard \textit{Conv} layer and a \textit{ACDA} layer to detect diverse patterns at scales. The ACDA layer clearly outperforms the standard convolutional layer, as the detection error of ACDA is of a magnitude lower than that of the standard convolutional layer.}
 		\vspace{3mm}
		\label{fig:toy0}
	\end{subfigure}
	\\
	\begin{subfigure}[b]{\linewidth}
		\resizebox{\linewidth}{!}{%
			\begin{tabular}{c c c c}
				\includegraphics[width=\CTS]{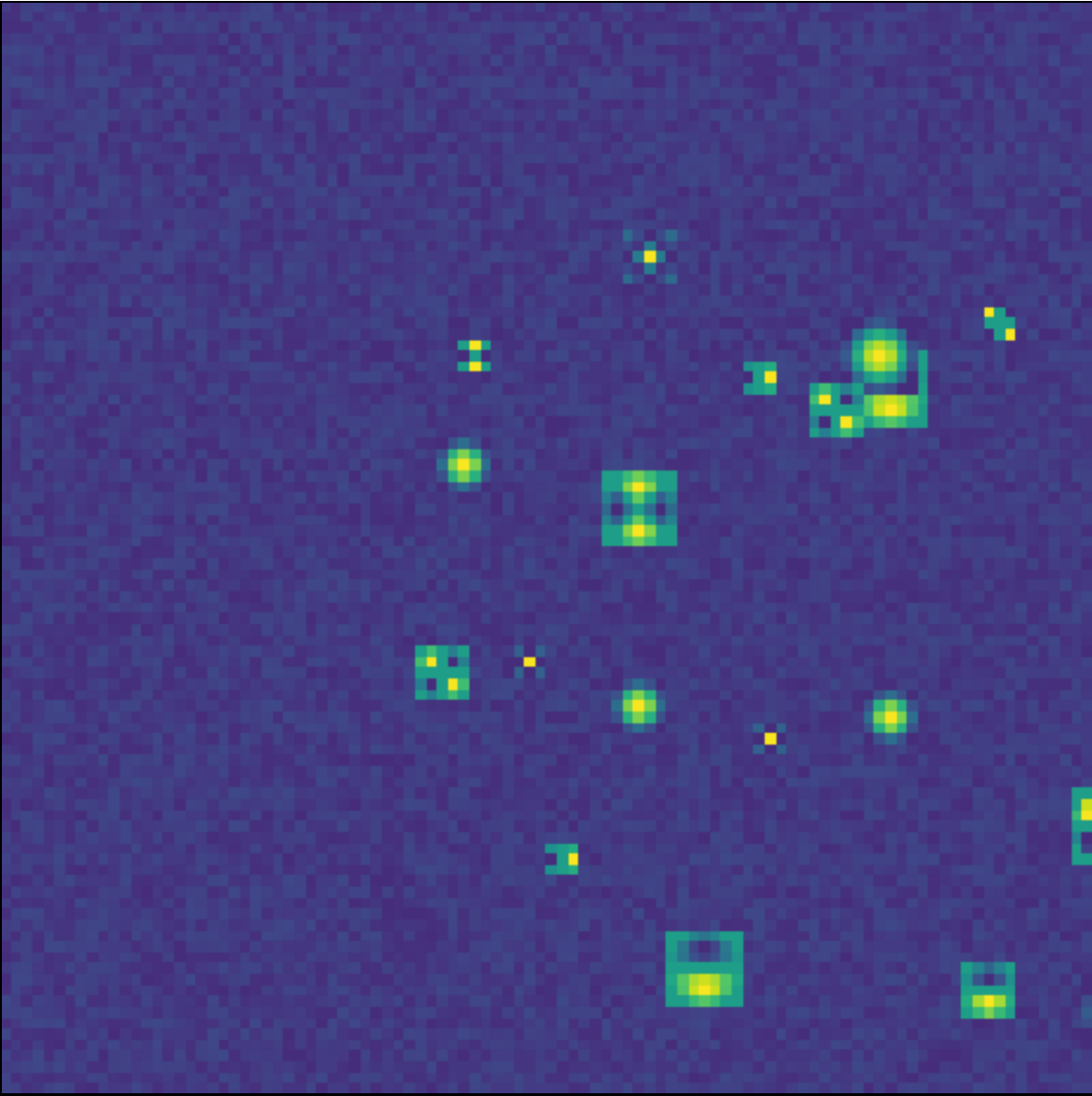}
				\CHSS
				&
				\CHSS
				\includegraphics[width=\CTS]{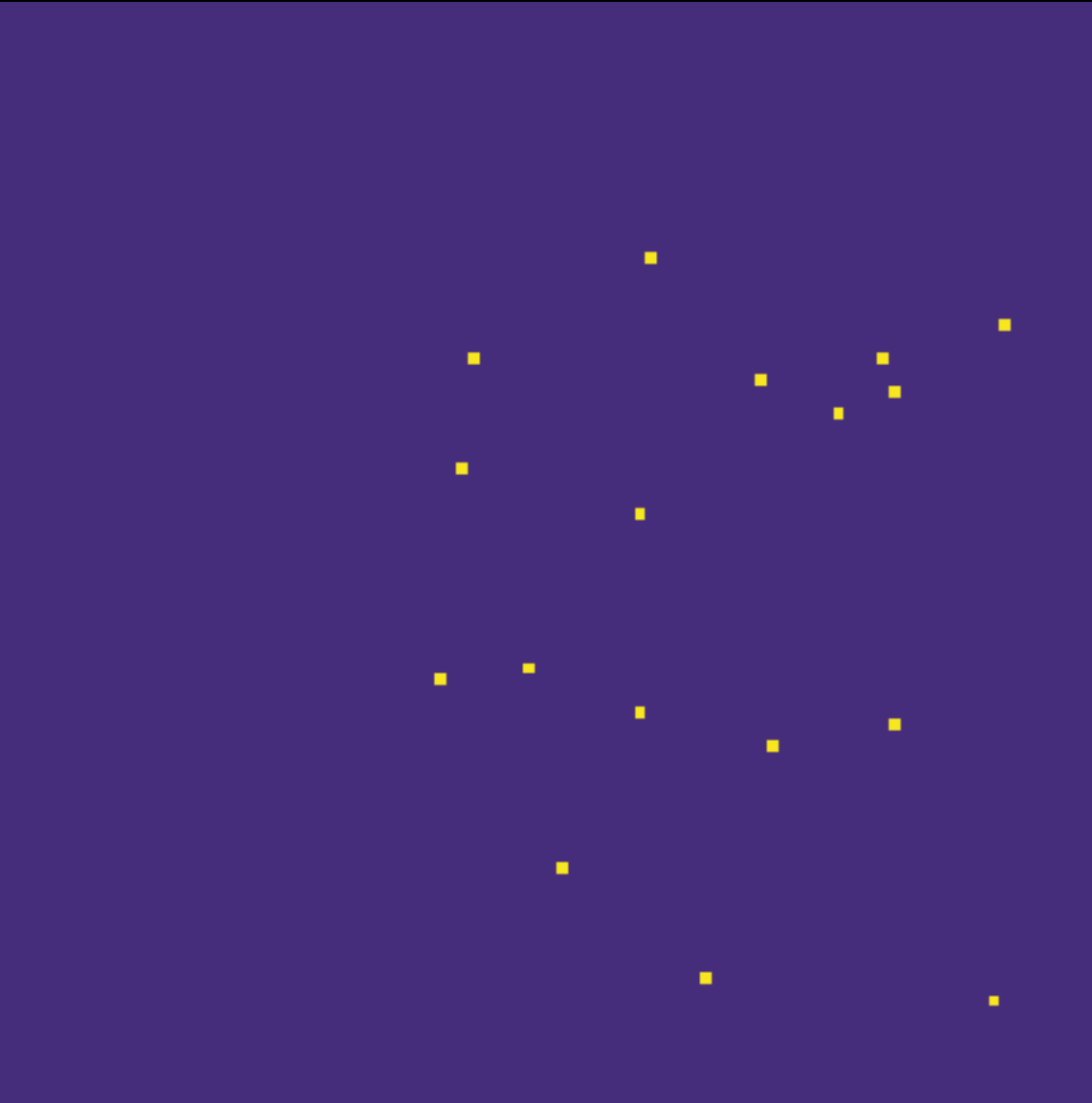}
				\CHSS
				&
				\CHSS
				\includegraphics[width=\CTS]{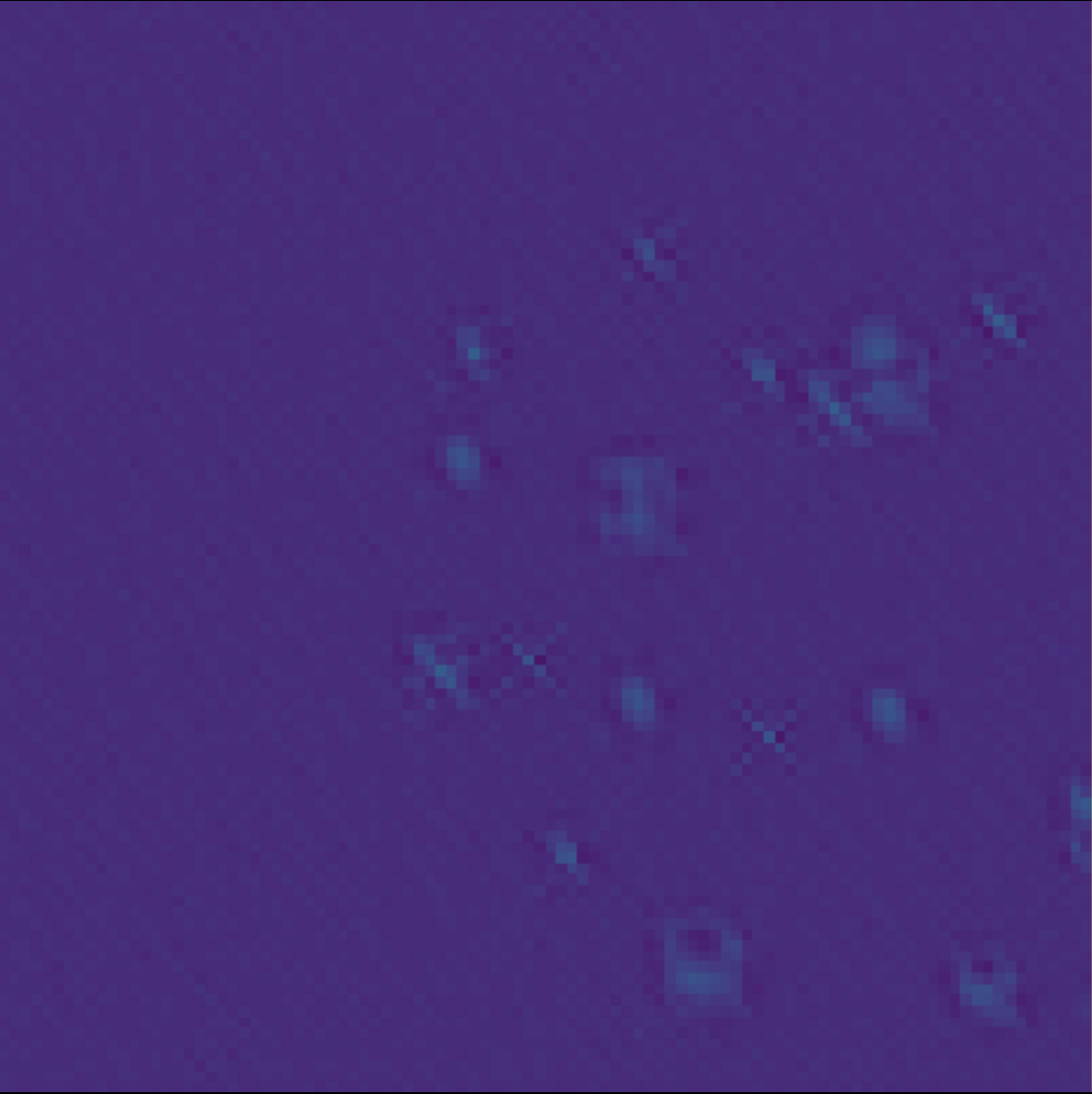}
				\CHSS
				&
				\CHSS
				\includegraphics[width=\CTS]{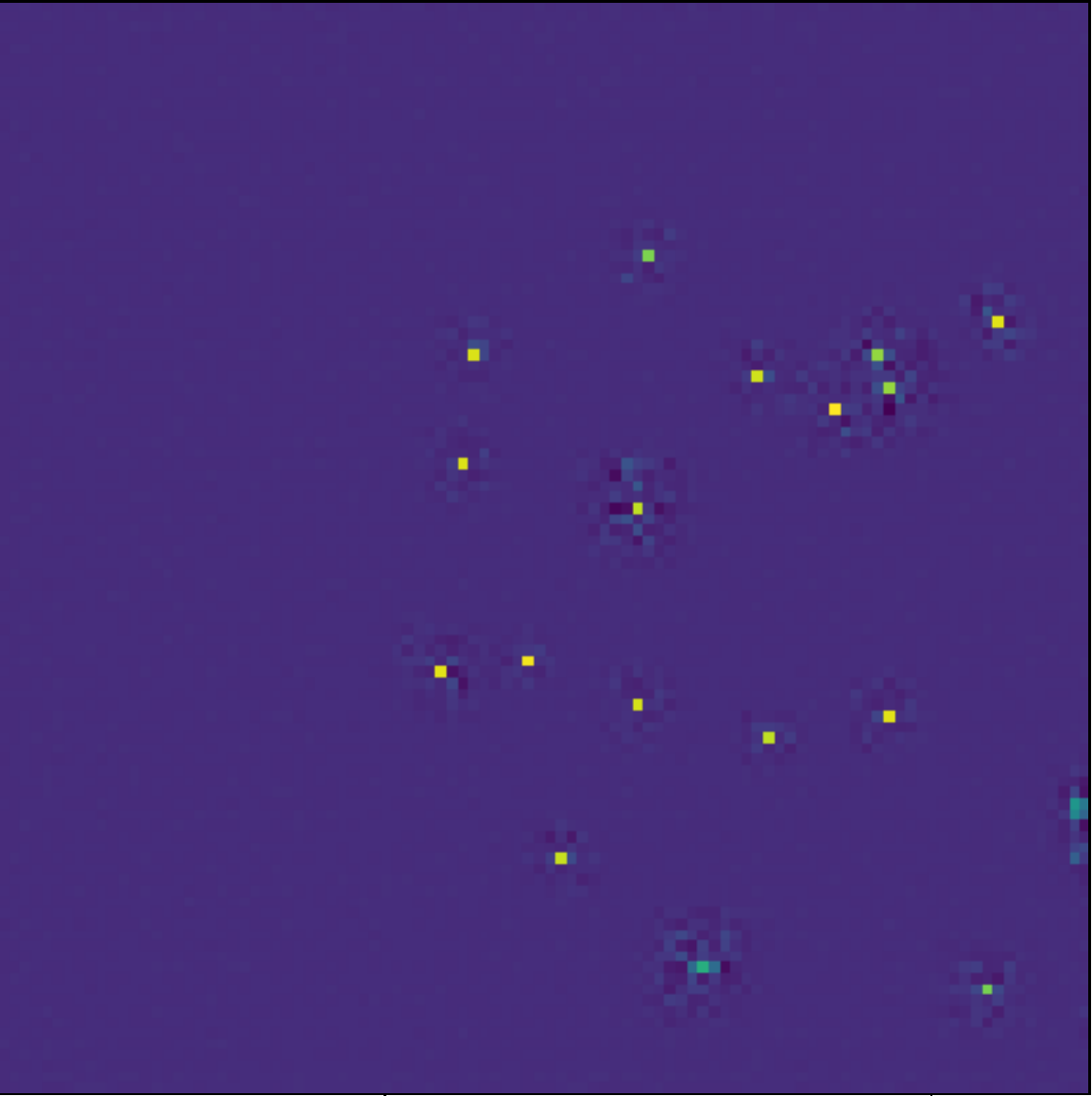}
				\hspace{-2mm}
				\\
				Input & \hspace{-3mm}GT & \hspace{-3mm} \textit{Conv} output & \hspace{-3mm} \textit{ACDA} output \\
		\end{tabular}}
		\caption{Translation equivariance. When testing with a shifted input image, the output predictions of both layers are exactly shifted versions of the ones in Figure~\ref{fig:toy0}.
		}
		\label{fig:toy1}
	\end{subfigure}
	
	\caption{Toy pattern detection experiments.}
	\label{fig:toy}
\end{figure}
We then test both layers with a shifted input as shown in Figure~\ref{fig:toy1}.
All the patterns in the image are shifted spatially to the bottom-right direction by 20 pixels. 
As shown in Figure~\ref{fig:toy1}, the outputs of both layers are exactly the shifted versions of those in Figure~\ref{fig:toy0}.
This is a clear demonstration that, thanks to the convolutional atom generation and shared coefficients, \textit{ACDA} preserve the appealing translation-equivariance property of standard convolutions.
These simple synthesized demonstrations show the flexibility and efficiency of \textit{ACDA}.
To further validate the effectiveness, we present real-world experiments in Section~\ref{exp}.

\section{Experiments}
\label{exp}
we present experimental results to fully validate the proposed approach, referred to as ACDA, on various applications. 
We start with image classification experiments, showing that by plug-and-playing the introduced ACDA into CNNs, 
comparable and improved performance can be obtained even with reduced channel numbers. 
The results indicate that the dynamic filters at convolutional layers can better handle image variance, thus alleviate the demand for learning filters with many channels for exhaustive feature matching.
We then move to real-world applications on crowd counting and image restorations, both of which involve significant intra-sample variance. 
We further demonstrate the advantages of ACDA through discussions on computation, memory, and parameters.

In all the experiments, if not otherwise specified, we adopt the same structure of atom generators as described in Appendix Section~\ref{phi}, with 3 sets of Fourier Bessel bases at scales from $3\times 3$ to $7 \times 7$. Note that although only three scales are used in each layer, the low cost allow ACDA layers to be stacked and achieve receptive fields with a very large range. E.g., stacking only two ACDA layers with three scales can achieve effective receptive fields range from $5\times 5$ to $13 \times 13$. 
We provide ablation study in Appendix Section~\ref{ablation} to validate the selections of hyperparameters.

\begin{table}[h]
	\centering
	\small
	\vspace{2mm}
	\caption{Illustrative image classification performance on CIFAR-10 and CIFAR-100. Top-1 error rates are reported.}
	\begin{tabular}{c|c|c}
		\toprule 
		Methods &  CIFAR-10 & CIFAR-100 \\
		\midrule
		LeNet \cite{lenet} & 24.74 & 56.60 \\
		\textbf{LeNet + ACDA}   & \textbf{16.27} (34.2\%$\downarrow$) & \textbf{49.53} (12.5\%$\downarrow$) \\
		\bottomrule
	\end{tabular}
	\label{tab:lenet}
\end{table}

\subsection{Image Classification}
Before diving into large scale experiments with customized networks, we start with a simple illustrative experiment with LeNet \cite{lenet}  and CIFAR.
LeNet is a tiny network architecture with only two convolutional layers.
We show in Table~\ref{tab:lenet} that, by only replacing the two convolutional layers in LeNet with the proposed adaptive convolutional layers, significant accuracy improvements are observed, indicating the extra expressiveness achieved by the proposed ACDA.

We then demonstrate the effective and scalable properties of ACDA by performing experiments on ImageNet. 
Without heavily tuning the network structure, we construct a simple architecture based upon deep residual networks (ResNet) \cite{resnet}.
We empirically observe that adopting ACDA in shallow layers in a CNN does not significantly influence the network performance, therefore we leave those layers unchanged.
In deep layers, we build \textit{dynamic bottleneck blocks} following the bottleneck blocks introduced in \cite{resnet}. 
Details on the network configurations are presented in Appendix Section~\ref{adres}.
We construct  architectures of adaptive ResNets with different sizes
and show in Table~\ref{tab:imagenet} that, while significantly reducing the parameters, networks with ACDA can deliver comparable performance as standard CNNs. And comparing to dynamic convolutions \cite{dycnn,condconv}, ACDA enjoys clear advantages on parameters.
Although building compact networks is not our primary focus in this paper, comparisons against state-of-the-art compact network architecture MobileNet-V3 \cite{mobilev3} show that ACDA can achieve comparable or even better performance comparing to those heavily tuned architectures.  

\begin{table}[]
	\begin{center} 
		\small
		\centering
		
		\caption{Image classification performance on ImageNet. 
			We report both parameter size as well as Top-1 and Top-5 error rates.}
		%			Results reproduced by our implementations are marked with $*$.}
		\label{tab:imagenet}
		\resizebox{0.9\linewidth}{!}{
		\begin{tabular}{c |c | c c }
			\toprule
			Methods & Parameters & Top-1 & Top-5 \\
			\midrule
			ResNet-18 & 11.69M & 30.24 & 10.92\\
			ResNet-34 & 21.28M & 26.70 & 8.58\\
			ResNet-50 & 25.56M & 23.85 & 7.13\\
			\midrule
			MobileNet-V3 small \cite{mobilev3} & 2.9M & 32.6 & 13.6\\
			\midrule
			CondConv-EfficientNet \cite{condconv} &13.3M  & 22.8 & - \\
			CondConv-ResNet-50 \cite{condconv} & - & 22.3 & - \\
			DY-MobileNetV3 small \cite{dycnn} & 4.8M & 29.7 & 11.3\\
			\midrule
			\textbf{Ad-ResNet-s} & 3.85M & 28.81 & 9.62\\
			\textbf{Ad-ResNet-m} & 9.83M & 25.81 & 7.92\\
			\textbf{Ad-ResNet-l} & 18.19M & 23.22 & 6.74\\
			\bottomrule
		\end{tabular}}
%		\vspace{-6mm}
	\end{center}
\end{table}

\subsection{Crowd Counting}
Crowd counting, aiming at counting the total number of particular objects (typically pedestrians), poses challenges on learning based methods due to the significantly large variance on the object sappearance.
We conduct experiments on crowd counting by simply adopting the networks with ACDA trained with ImageNet as feature extractor, and replace the final linear layer with few standard transposed convolutional layers for recovering resolution. 
We follow the simplest practice and generate groundtruth heatmaps of each object marked by a \textbf{fixed-size} Gaussian kernel, and directly train the networks with a mean square error (MSE) loss.
Without bells and whistles, networks with ACDA achieve state-of-the-art results on large scale datasets, UCF-QNRF \cite{ucf} and ShanghaiTech \cite{shanghai} subset A.
Both datasets are collected from various sources so that contain significant variance reflected by large diversity of viewing angles, image qualities, and etc.

\begin{table}[]
	\begin{center} 
		\small
		\centering
		\caption{Comparisons on large-scale crowd counting datasets. Numbers of parameters are reported in millions.}
		\label{tab:cc}
		\resizebox{\linewidth}{!}{
			\begin{tabular}{c |c c | c c c}
				\toprule
				Datasets & \multicolumn{2}{c|}{SHTech-A} & \multicolumn{2}{c}{UCF-QNRF}\\
				\midrule
				Metrics & MAE & MSE & MAE & MSE \\
				\midrule
				MCNN \cite{single} & 110.2 & 173.2 &  277 & 426 \\
				Switch-CNN \cite{switch} & 90.4 & 135.0 & 228 & 445 \\
				SCNet \cite{wangcc} & 71.9 & 117.9 & - & - \\
				ic-CNN \cite{iccnn} & 68.5 & 116.2 & - & - \\
				SANet \cite{sanet} & 67.0 & 104.5 & - & - \\
				CL-CNN \cite{compositional_cc} & - & - & 132 & 191\\
				PACNN \cite{perspective} & 62.4 & 102.0 & - & - \\
				SFCN \cite{synthetic} (38.60M)& 64.8 & 107.5 & 102 & 171  \\
				CAN \cite{context} (18.10M) & 62.3 & 100.0 & 107 & 183 \\
				Wan \textit{et al.} \cite{addense} & 64.7 & 97.1 &  101 & 176\\
				BL \cite{bl} (21.50M)& 62.8 & 101.8 & 88.7 & 154.8 \\
				\midrule
				CondConv-s \cite{condconv} (14.8M) & 68.44  & 112.96 & 117 & 182 \\
				CondConv-l \cite{condconv} (25.5M) & 63.82  & 104.23  & 109 & 179 \\
				\midrule
				Baseline (4.78M)& 67.73 & 110.12 & 124.77 & 210.05 \\
				\textbf{Ad-ResNet-s} (4.87M) & 57.88 & 91.27 & 99.22 & 182.13 \\
				\textbf{Ad-ResNet-m }(11.87M) & 56.04 & 89.76 & 96.08 & 176.87 \\
				\bottomrule
		\end{tabular}}
%	\vspace{-6mm}
	\end{center}
\end{table}

\begin{figure}

	\resizebox{\linewidth}{!}{%
		\small
		\begin{tabular}{c c c}
			\hspace{-2mm}
			\includegraphics[width=\CTS]{./figs/results/crowd/000000_180input.jpg}
			\CHSS
			&
			\CHSS
			\includegraphics[width=\CTS]{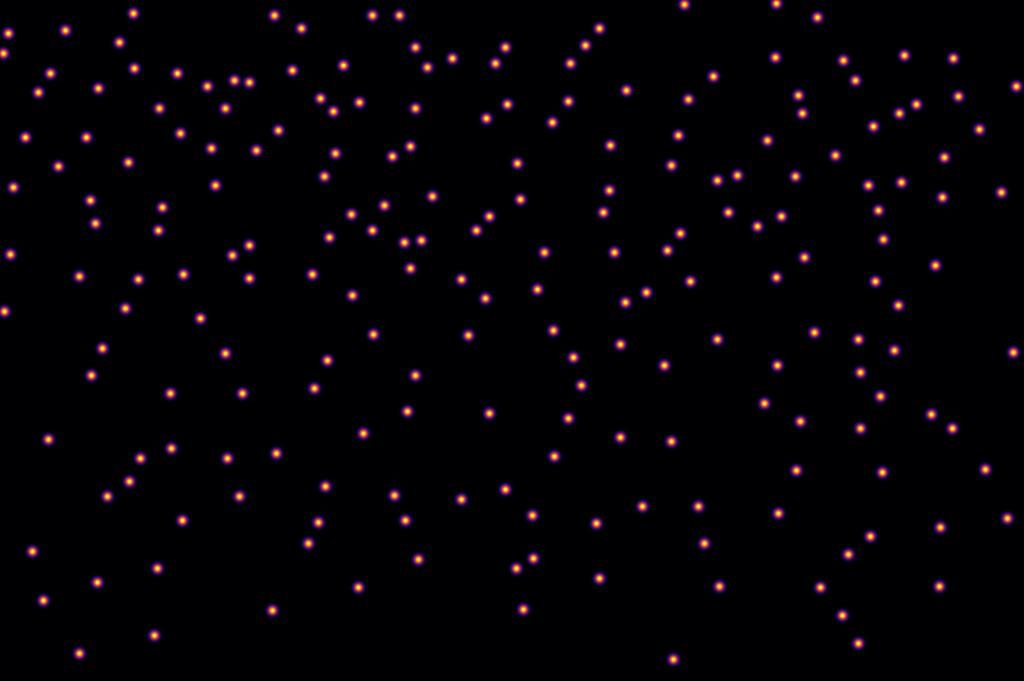}
			\CHSS
			&
			\CHSS
			\includegraphics[width=\CTS]{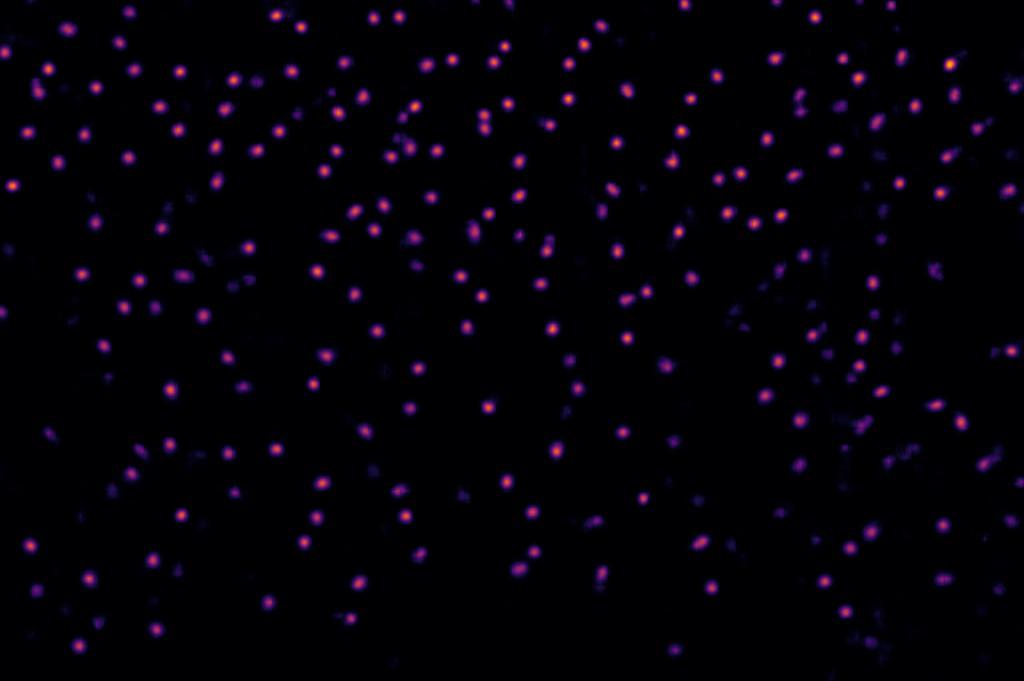}
			\hspace{-2mm}
			\\
			\vspace{1mm}
			
			Input & \hspace{-3mm}GT: 215 & Prediction: 213 \\
			
			\hspace{-2mm}
			\includegraphics[width=\CTS]{./figs/results/crowd/000000_01input.jpg}
			\CHSS
			&
			\CHSS
			\includegraphics[width=\CTS]{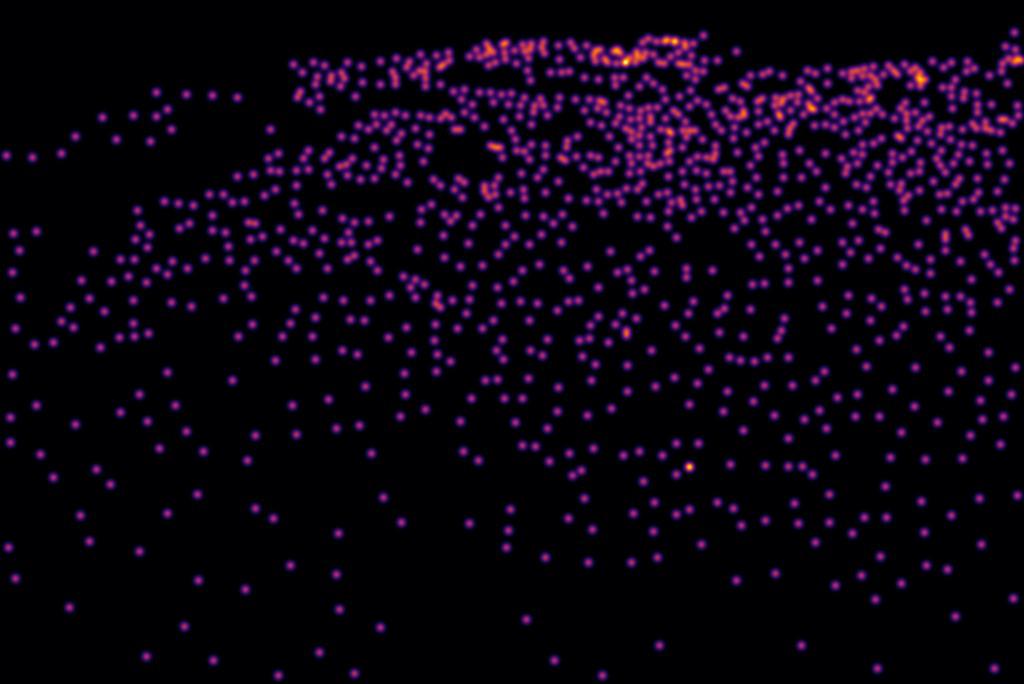}
			\CHSS
			&
			\CHSS
			\includegraphics[width=\CTS]{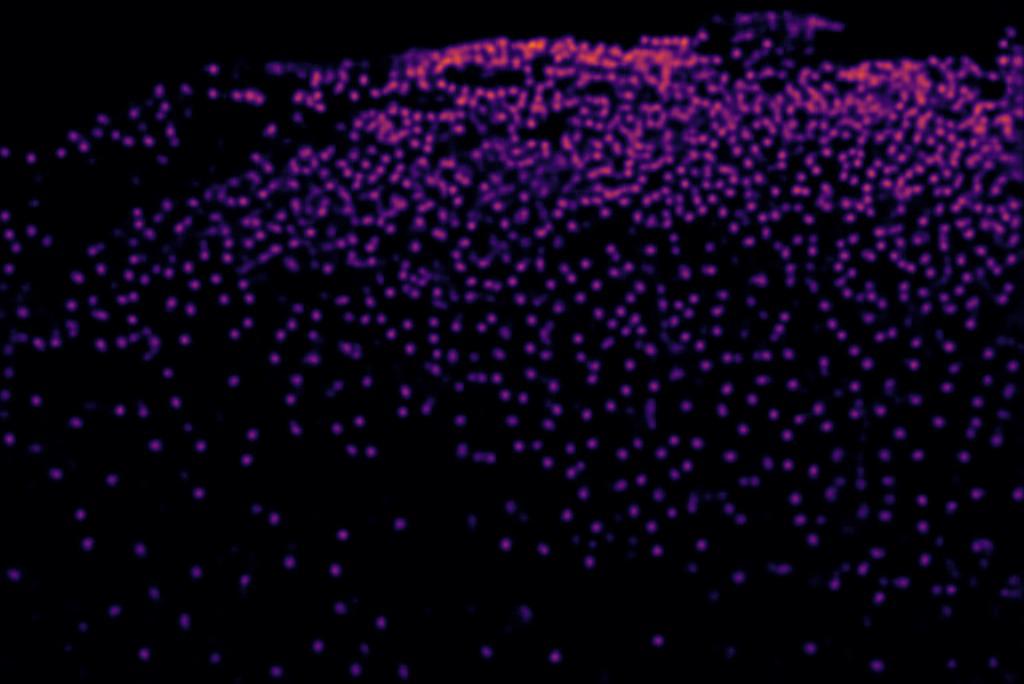}
			\hspace{-2mm}
			\\
			
			Input & \hspace{-3mm}GT: 1110 & Prediction: 1196 \\
			
	\end{tabular}}
% 	\vspace{-2mm}
	\caption{Qualitative results on crowd counting.}
	\label{fig:viscc}
\end{figure}

Following the normal protocol, we report results with mean absolute error (MAE) and mean square error (MSE) in Table~\ref{tab:cc}, and compare the results against various state-of-the-art methods that adopts customized network architectures \cite{single,switch,wangcc,sanet} and loss functions as well as training strategies \cite{iccnn,compositional_cc,perspective,context,bl}.
The baseline performance is obtained by training a same network architecture as Ad-ResNet with standard convolutional layers only.
Networks with ACDA achieve significant improvements over state-of-the-art methods on the ShanghaiTech-A dataset.
To further demonstrate the advantages of per-pixel specific adaptive filters with ACDA over per-image specific adaptive filters, we adopt the official models of CondConv \cite{condconv} and train them on crowd counting by appending a light weight decoder network. 
The results indicate that, the per-pixel adaptive filters can deliver much better performance comparing to both standard and per-image adaptive filters, with even reduced network scales.
We present a visualization of basis coefficient heatmaps in Figure~\ref{fig:coef} to validate the unique advantages of ACDA on addressing scale variance.
It is clearly shown when processing images with significant intra-image scale variance, ACDA can selectively decide the effective receptive fields at each position based on the target object sizes, by adjusting the weights of multi-scale atom bases.
Qualitative results are in Figure~\ref{fig:viscc} and Appendix Section~\ref{supp_cc}.

\newcommand{\TS}{0.24\linewidth}
\newcommand{\HS}{\hspace{-6.0mm}}
\newcommand{\HSS}{\hspace{-4.0mm}}

\begin{figure}
	\resizebox{\linewidth}{!}{%
		\begin{tabular}{c | c c c}
			\includegraphics[width=\TS]{./figs/results/coef/000000_177input.jpg}
			\HS
			&
			% 	\HS
			\includegraphics[width=\TS]{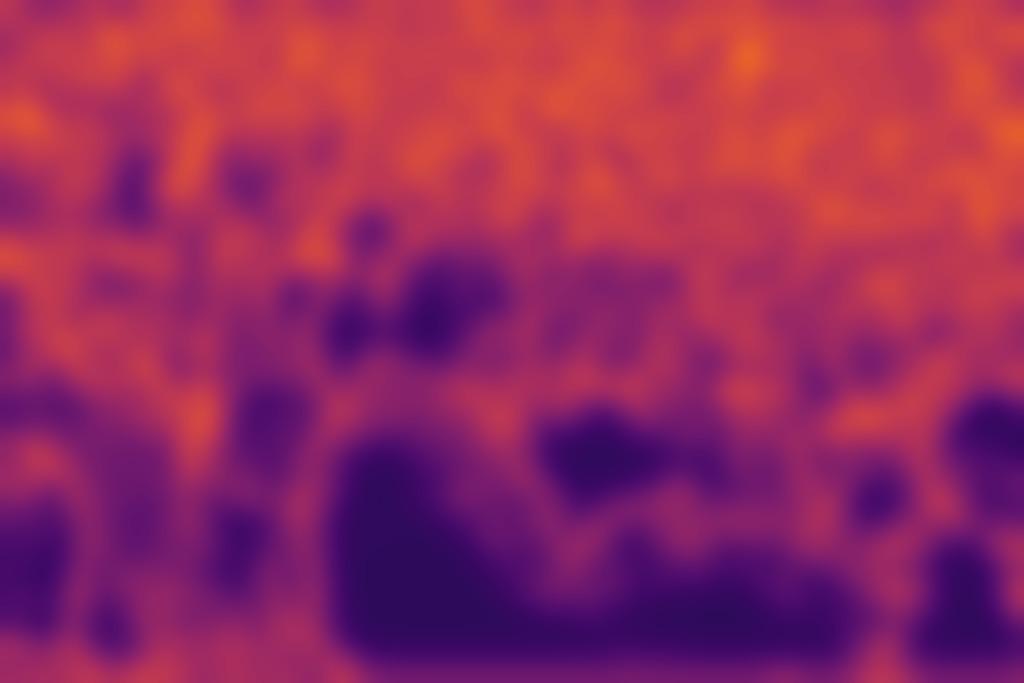}
			\HSS
			&
			%	\HS
			\HSS
			\includegraphics[width=\TS]{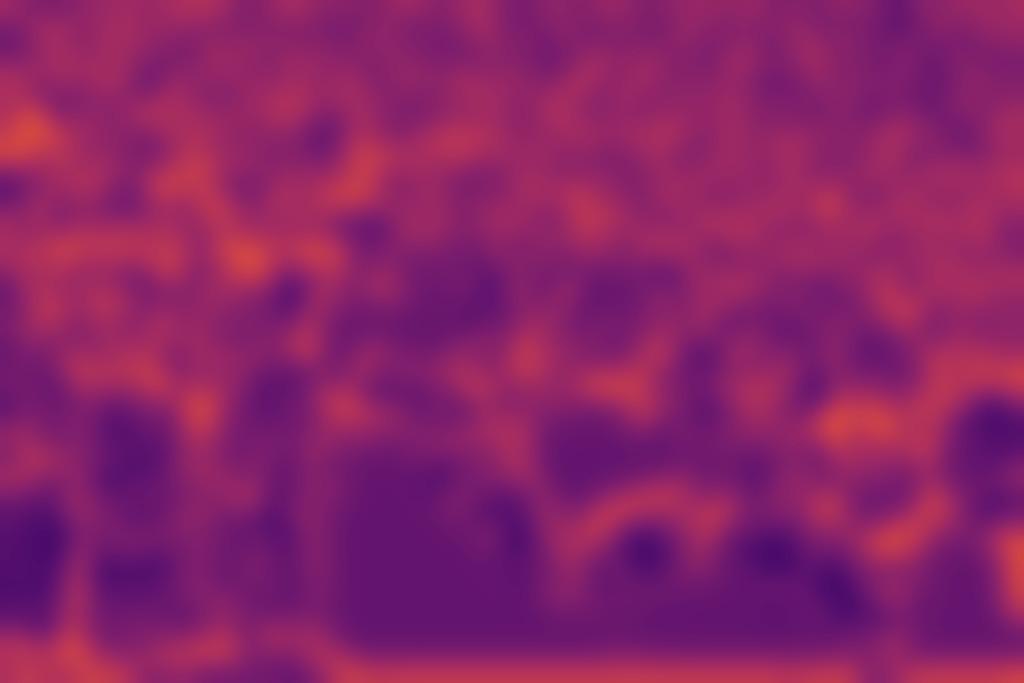}
			\HSS
			&
			\HSS
			\includegraphics[width=\TS]{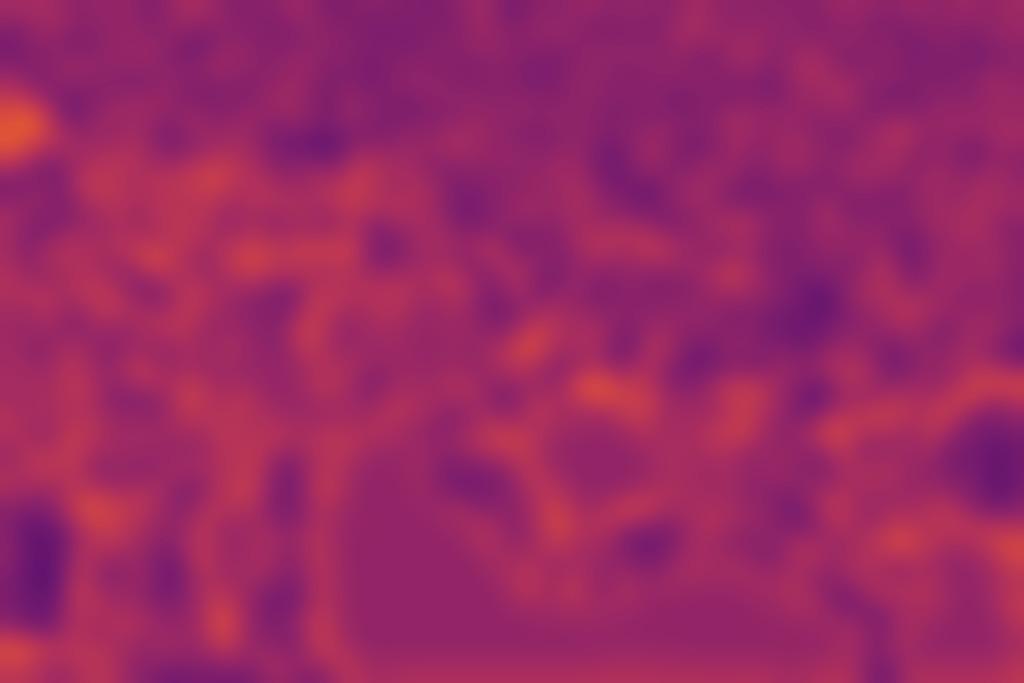}
			\hspace{-2mm}
			\\

			\hspace{-2mm}
			\includegraphics[width=\TS]{./figs/results/coef/000000_178input_.jpg}
			\HS
			&
			% 	\HS
			\includegraphics[width=\TS]{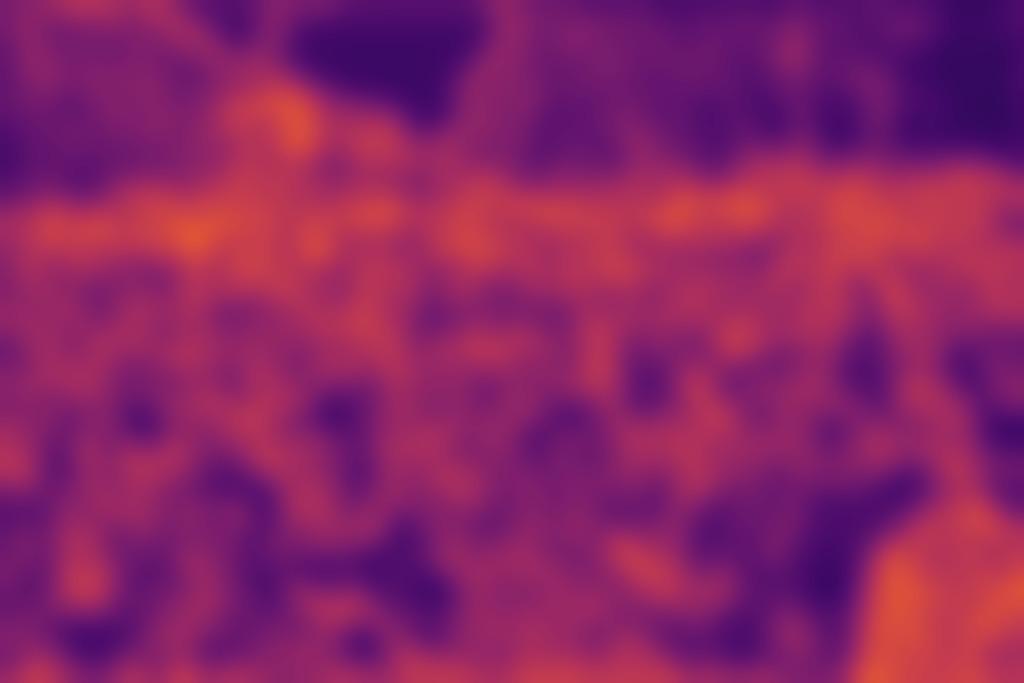}
			\HSS
			&
			%	\HS
			\HSS
			\includegraphics[width=\TS]{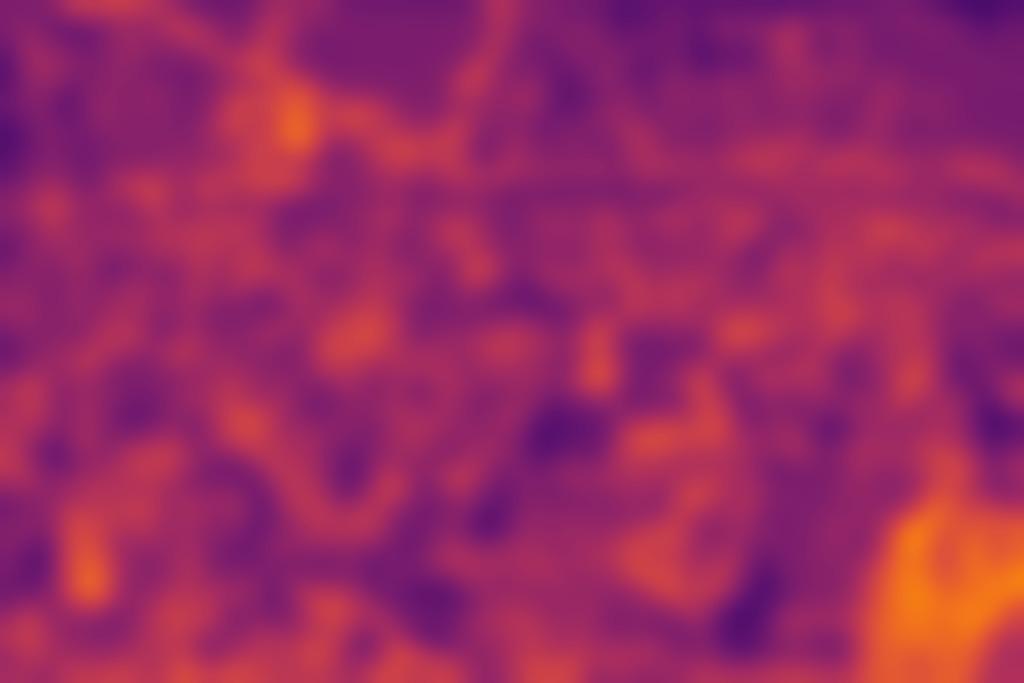}
			\HSS
			&
			\HSS
			\includegraphics[width=\TS]{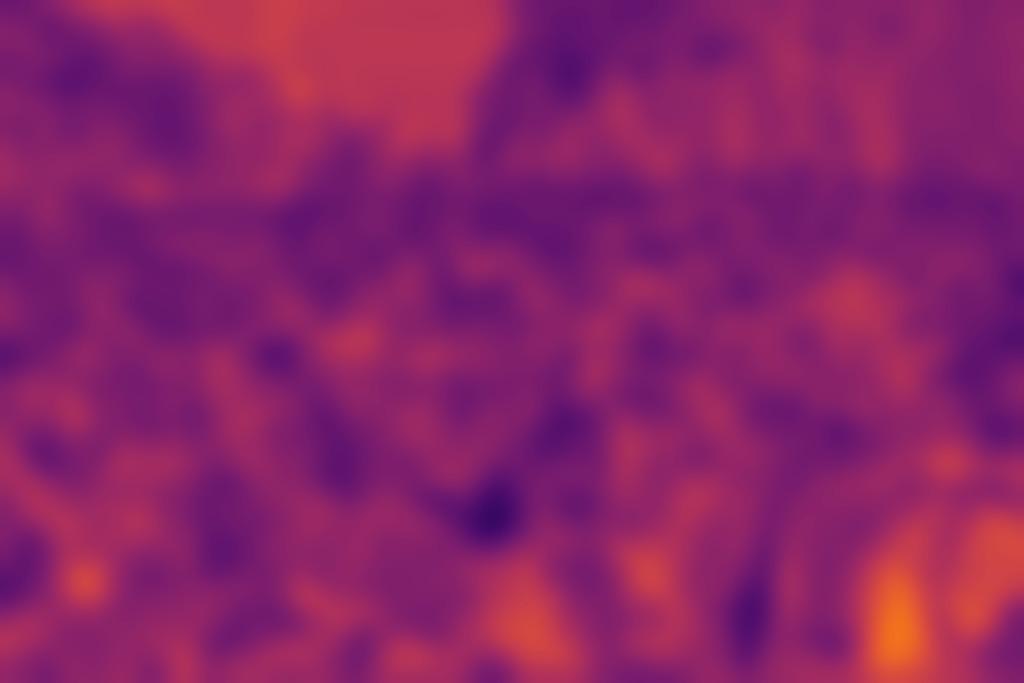}
			\hspace{-2mm}
			\\
			
			Input & $3\times3$ & \hspace{-5mm}$5\times5$ & \hspace{-2mm}$7\times 7$ \\

	\end{tabular}}
	\caption{Visualizations of the atom basis coefficients heatmaps (lighter the higher). Our per-pixel adaptive convolutions tend to adopt large kernel sizes, i.e., with $7\times 7$ atom bases, when the objects in the target regions have large spatial sizes, i.e., the closer objects. While $3\times3$ bases are preferred when targeting on regions with dense objects.}
	\label{fig:coef}
\end{figure}

%\paragraph{Real-world image restorations.}
\subsection{Real-world Image Restorations}
In real-world image restorations, the degradation model can be highly non-uniform across spatial position of an image and locally feature dependent. 
We adopt ACDA to recover real-world degradation in a fully non-linear manner.
We perform experiments on real-world SR dataset RealSR \cite{realsr} and real-world denoising dataset SIDD \cite{sidd}.

\begin{table}[t]
	\begin{center} 
		\centering
		\caption{Comparisons on the RealSR real-world super-resolution dataset. $l$ is the kernel size.}
		\label{tab:realsr}
		\resizebox{\linewidth}{!}{
			\begin{tabular}{c |c c c | c c c}
				\toprule
				Metrics & \multicolumn{3}{c|}{PSNR} & \multicolumn{3}{c}{SSIM} \\
				\midrule
				Scales & $\times 2$ & $\times 3$ & $\times 4$ & $\times 2$ & $\times 3$ & $\times 4$ \\
				\midrule
				\midrule
				\multicolumn{7}{c}{\textit{RealSR 2019}}\\
				\midrule
				Bicubic & 32.61 & 29.34 & 27.99 & 0.907 & 0.841 & 0.806\\
				\midrule
				VDSR \cite{kim2016accurate} & 33.64 & 30.14 & 28.63 & 0.917 & 0.856 & 0.821\\
				SRResNet \cite{ledig2017photo} & 33.69 & 30.18 & 28.67 & 0.919 & 0.859 & 0.824 \\
				RCAN \cite{zhang2018image} & 33.87 & 30.40 & 28.88 & 0.922 & 0.862 & 0.826 \\
				\midrule
				% DPS & 33.71 & 30.20 & 28.69 & 0.919 & 0.859 & 0.824\\
				KPN, $l=5$ & 33.75 & 30.26 & 28.74 & 0.920 & 0.860 & 0.826 \\
				KPN, $l=19$ \cite{kpn} & 33.86 & 30.39 & 28.90 & 0.924 & 0.864 & 0.830\\
				LP-KPN, $l=5$ \cite{realsr} & 33.90 & 30.42 & 28.92 & 0.927 & 0.868 & 0.834\\
				\midrule
				\textbf{ACDA} & 33.98 & 30.62 & 28.97 & 0.929 & 0.871 & 0.937 \\
				% \bottomrule
				\midrule
				\midrule
				\multicolumn{7}{c}{\textit{RealSR Final}}\\
				\midrule
				KPN, $l=5$ \cite{kpn} & 33.41 & 30.47 & 28.80 & 0.913 & 0.860 & 0.826 \\
				KPN, $l=19$ \cite{kpn} & 33.45 & 30.57 & 28.99 & 0.914 & 0.864 & 0.832\\
				LP-KPN, $l=5$ \cite{realsr} & 33.49	& 30.60	& 29.05	& 0.917	& 0.865	& 0.834\\
				\midrule
				%        Baseline $l=5$ & \\
				\textbf{ACDA}  & 33.54 & 30.73 & 29.28 &  0.918 & 0.868 & 0.836\\
				\bottomrule
		\end{tabular}}
%		\vspace{-7mm}
	\end{center}
\end{table}

\paragraph{Real-world single-image super resolution.}
We adopt an extremely simple network architecture modified from the network used in \cite{realsr}.
We use two convolution layers and pixel shuffle downsampling layers to reduce the feature resolution, and $8$ consecutive \textit{dynamic bottleneck block}s to process the intermediate features.
Finally, two convolution layers with pixel shuffle upsampling layers are followed to restore the feature resolution and output the final predictions.
Despite being extremely simple, the adopted network is shown to be effective on handling real-world SR with spatially non-uniform and potentially feature dependent degradation models.
We show comparisons on two versions of the RealSR \cite{realsr} dataset, \textit{RealSR 2019} , and a larger \textit{RealSR final}. 
We compare ACDA against state-of-the-art methods using standard convolutions \cite{kim2016accurate,ledig2017photo,zhang2018image} and methods that adopt simplified adaptive convolutions \cite{realsr,kpn}.
We follow the standard practice in \cite{realsr} and train the network with random cropped image patches, and simple mean squared error (MSE) as the loss function.
The results and comparisons are presented in Table~\ref{tab:realsr}.
Following \cite{enhanced,realsr,zhang2018image}, we use PSNR and SSIM \cite{ssim} indices on the Y channel in the YCbCr space as the metrics of performance evaluations.
While methods based on simplified adaptive convolutions achieve good results on restoring non-uniformed degradation, they use only combinations of linear corrections. 
We show that endowing networks with local adaptive filters in a fully non-linear way can further boost performance. 
The experimental results and our methods shed the light on future real-world SR methods that rely on little assumptions on local degradation. 
Qualitative results and comparisons are shown in Figure~\ref{fig:sr0} and Appendix Section~\ref{supp_sr}.

\newcommand{\sTS}{0.32\linewidth}
\newcommand{\sHS}{\hspace{-6.0mm}}
\newcommand{\sHSS}{\hspace{-4.0mm}}

\begin{figure}
	\centering
	\resizebox{\linewidth}{!}{%
		\small
			\vspace{2mm}
		\begin{tabular}{c c c c}
			\includegraphics[width=\CTS]{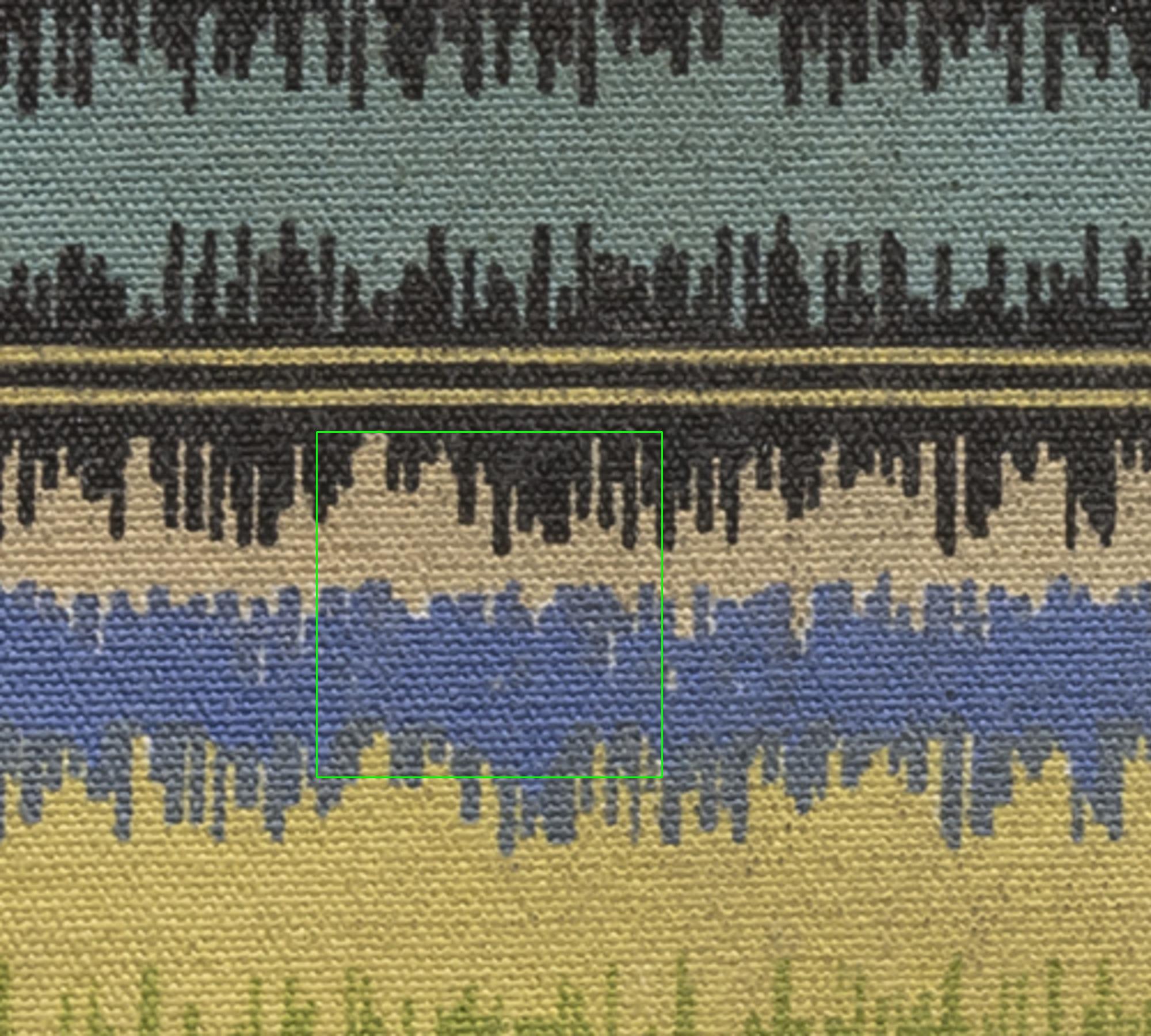}
			\sHSS
			&
			\sHSS
			\includegraphics[width=\CTS]{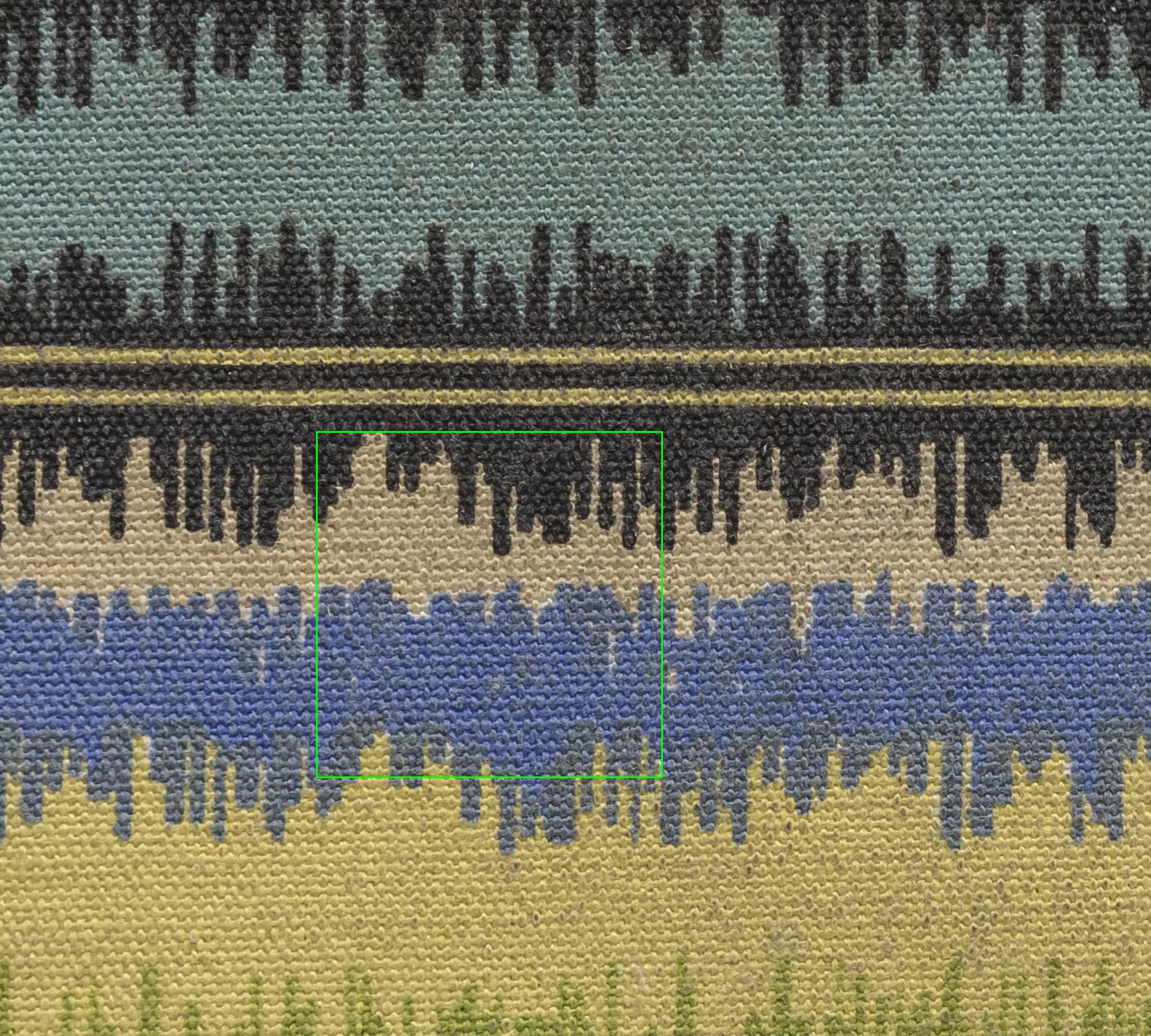}
			\sHSS
			&
			\sHSS
			\includegraphics[width=\CTS]{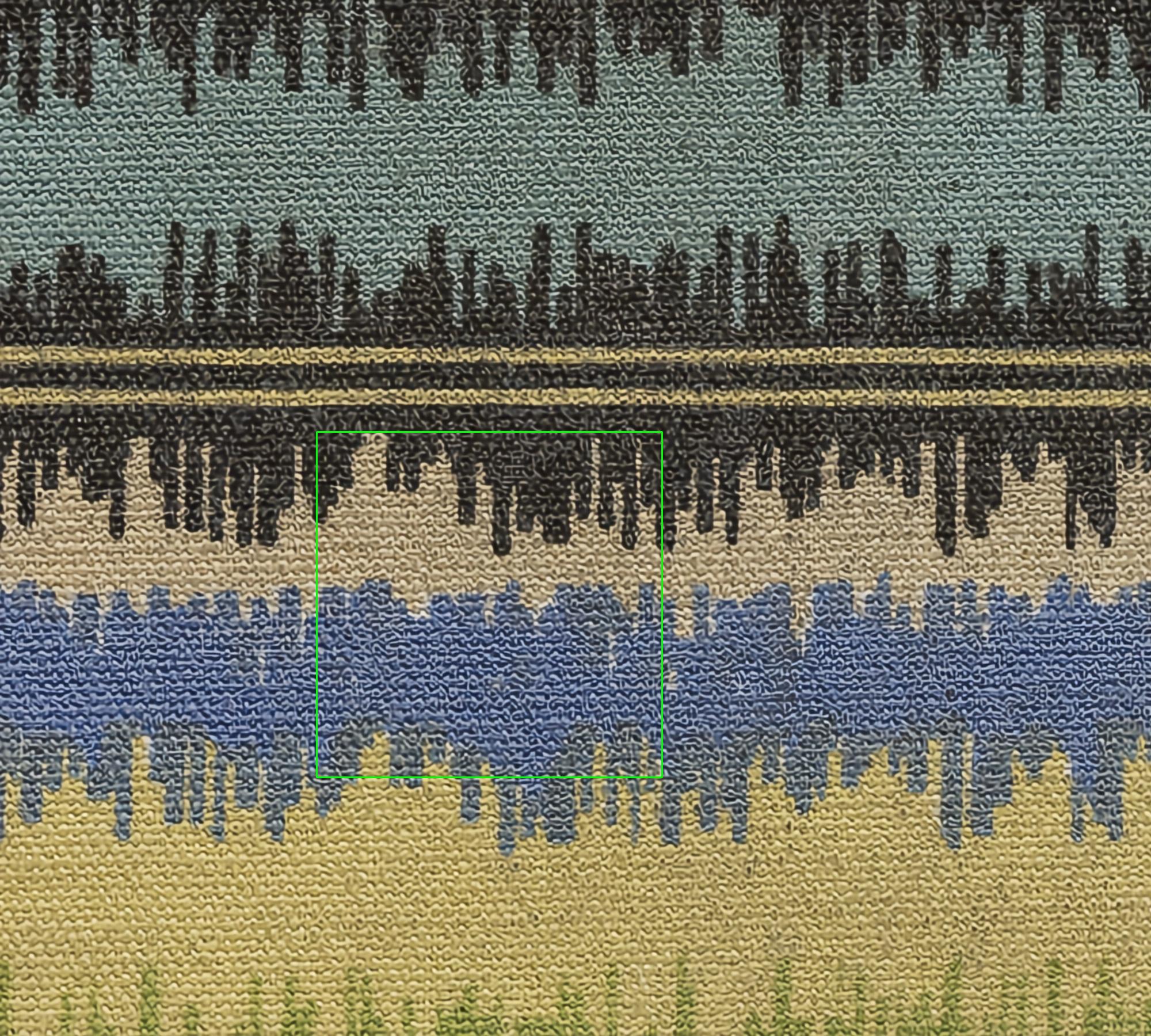}
			\sHSS
			&
			\sHSS
			\includegraphics[width=\CTS]{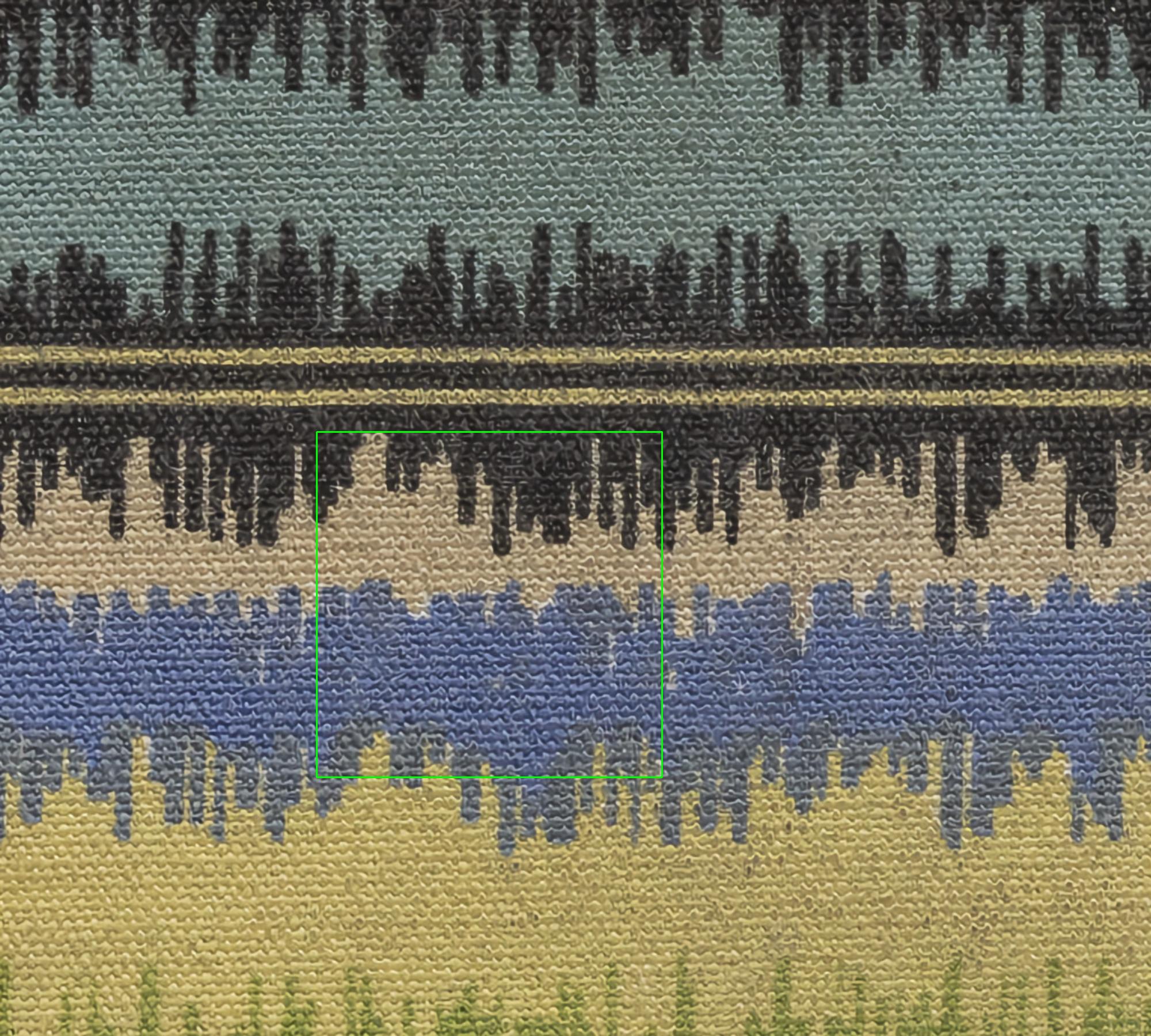}
			\hspace{-2mm}
			\\
			%			\vspace{1mm}
			\includegraphics[width=\CTS]{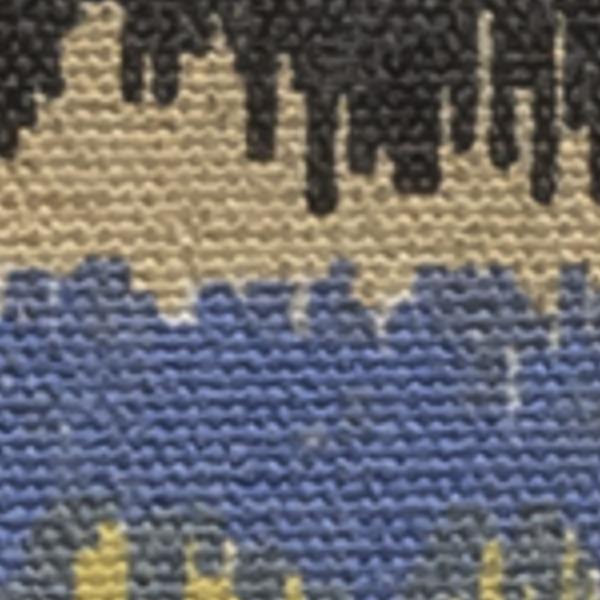}
			\sHSS
			&
			\sHSS
			\includegraphics[width=\CTS]{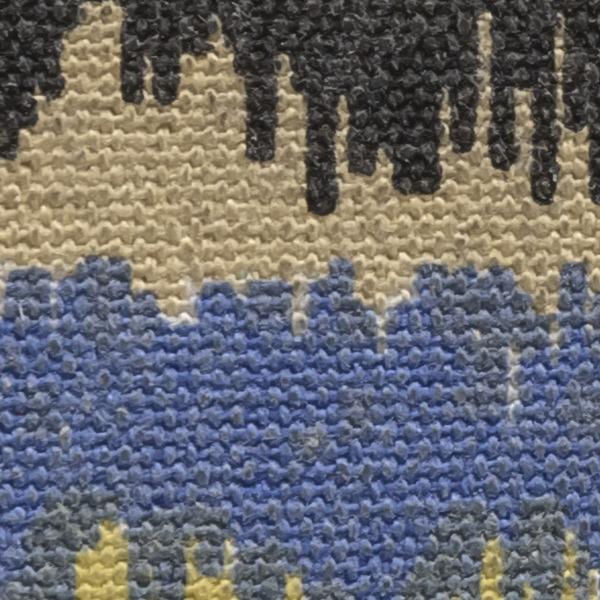}
			\sHSS
			&
			\sHSS
			\includegraphics[width=\CTS]{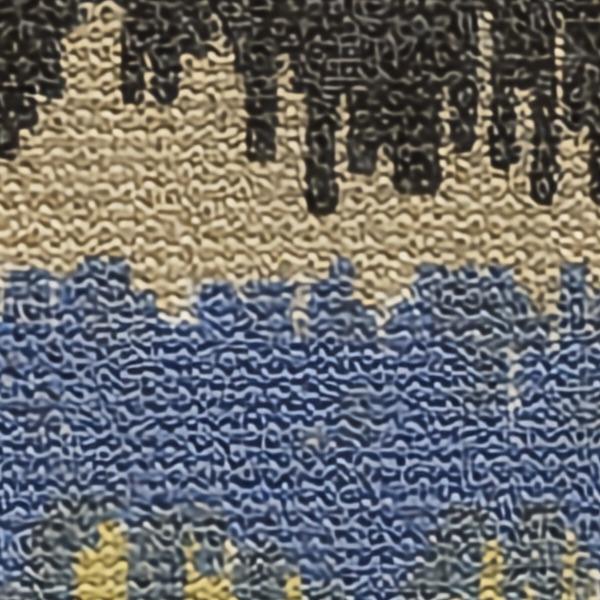}
			\sHSS
			&
			\sHSS
			\includegraphics[width=\CTS]{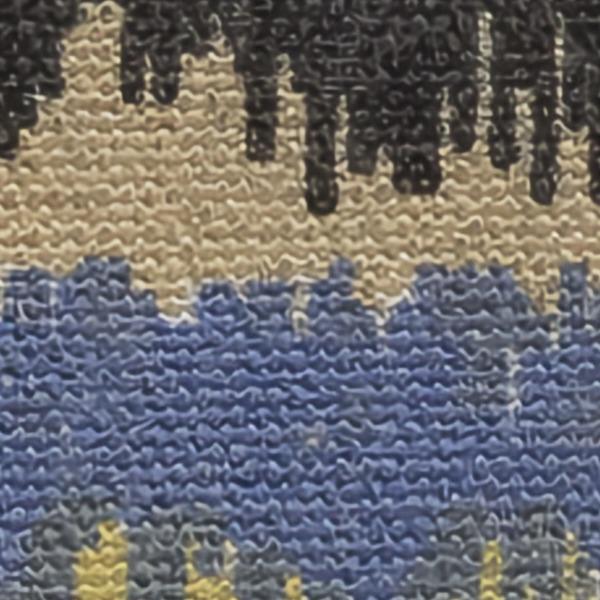}
			\hspace{-2mm}
			\\
			
			Low resolution & \hspace{-2mm} High resolution&  \hspace{-2mm} LP-KPN & \hspace{-2mm} Ours \\
			& &  \hspace{-2mm} PSNR 22.20 & \hspace{-2mm} PSNR  27.56 \\
	\end{tabular}}
	\caption{Qualitative comparisons against LP-KPN. LP-KPN suffers from strong artifacts. ACDA produces more faithful results. We believe the adaptive convolutions operated in deep features help better capture image semantics, thus prevent over-sharp results with strong artifacts.}
%	\vspace{-2mm}
	\label{fig:sr0}
\end{figure}

\paragraph{Real-world image denoising.}
We then perform real-world image denoising with the SIDD \cite{sidd} dataset.
Similar to real-world SR, real-world denoising aims at restoring high-quality images given noisy inputs with spatially non-uniform non-i.i.d. real-world noise.
The noise can be locally feature-dependent, thus adaptive convolutions a naturally good tool for modeling the intra-image variance. 
We present the quantitative results in Table~\ref{tab:sidd}. Qualitative comparisons are presented in Appendix Section~\ref{supp_dn}.
Baseline performance is obtained by using the same network architecture as ACDA with standard convolutions only.
As shown in Table~\ref{tab:sidd}, although the dynamic atom generation network introduces additional parameters, the overall size is smaller in ACDA comparing to the baseline thanks to the atom bases decomposition.
We present results on ACDA by using both simple MSE training loss (ACDA + MSE) and the state-of-the-art variational denoising framework proposed in VDNet\cite{vdn} (ACDA + VDNet).
ACDA delivers improvements over baseline models with fewer parameters.

\begin{table}[h]
	\begin{center} 
		\small
		\centering
				\vspace{2mm}
		\caption{Comparisons on the RealSR real-world super-resolution dataset.}
		\label{tab:sidd}
		\begin{tabular}{c |c c c | c c c}
			\toprule
			Metrics & PSNR & SIDD \\
			\midrule
			DnCNN-B \cite{dncnn} & 38.41 & 0.909\\
			CBDNet \cite{cbdnet} & 38.68 & 0.901\\
			VDNet \cite{vdn} with UNet (7.70M) & 39.28 & 0.909\\
			\midrule
			Baseline + MSE (2.28M) &  38.74 & 0.902\\
			\textbf{ACDA + MSE }(1.97M)  & 38.96 & 0.905 \\
			\textbf{ACDA + VDNet} (1.97M) & 39.32 & 0.912  \\
			\bottomrule
		\end{tabular}
%		\vspace{-3mm}
	\end{center}
\end{table}

\subsection{Discussions on Efficiency}
\label{cost}
\noindent \textbf{Computation and parameter.}
A regular convolution needs $c^\prime hw \cdot c (1+l^2)$ FLOPS, while \textit{ACDA} needs $\underbrace{c^\prime m (1+l^2)}_{\text{atom conv}} + \underbrace{c^\prime m c}_{\text{coefficient}} + \underbrace{c^\prime h w  d (1+l_a^2) + d h w Sm^\prime (1+l_b^2)}_{\text{atom generation}}$, where $d=64$, $l_a=1$, and $l_b=3$ following Section~\ref{adres}. 
For more straightforward comparisons, we present in Table~\ref{tab:para} comparisons on parameter size and FLOPs between standard convolution (denoted as Conv) and ACDA.
The numbers are obtained by calculating the parameters and computation in a single layer with a typical setting: 256 input and output channels, and $100 \times 100$ feature resolution.
We report comparisons with three kernel sizes from $3\times 3$ to $7 \times 7$.
As shown in Table~\ref{tab:para}, ACDA have clear advantages on both parameter size and computation in all settings. 
When using large kernel sizes, the advantages become more superior thanks to the atom bases.
The comparisons between the numbers of ACDA (conv only) and ACDA (+ atom generation) show that the atom generation only introduces small overhead under the typical settings.

\begin{table}[]
	\begin{center} 
		\small
		\centering
		
		\caption{Comparisons between standard convolution (Conv) and ACDA on computation (FLOPs) and parameters (Params). All numbers are reported in millions. 
			ACDA (conv only) denotes the two stage convolutions only, and  ACDA (+ atom generation) denote the entire process of atom generations and convolutions.}
		\label{tab:para}
		\resizebox{\linewidth}{!}{
			\begin{tabular}{c | c |c c c }
				\toprule
				\multicolumn{2}{c|}{Kernel size} & $3\times3$ & $5\times5$ &$ 7\times 7$\\
				\midrule
				\multirow{3}{*}{\rotatebox{90}{FLOPs}} & Conv & 5,900.8 & 16,386.6 & 32,115.2\\
				& ACDA (conv only) & 4,073.0 & 4,318.7 & 4,687.4 \\
				& ACDA (+ atom generation) & 4,311.2 & 4,557.0 & 4,925.7\\
				\midrule
				\multirow{3}{*}{\rotatebox{90}{Params}} & Conv  & 0.59 &  1.64 & 3.12\\
				& ACDA (conv only) &0.39  & 0.39 & 0.39 \\
				& ACDA (+ atom generation) & 0.41 & 0.43 &0.45 \\
				\bottomrule
		\end{tabular}}
		%}
	\vspace{-2mm}
	\end{center}
\end{table}

\begin{table}[]
	\begin{center} 
		\small
		\centering
% 		\vspace{2mm}
		\caption{Comparisons on training memory and time. }
		\label{tab:memo}
		\resizebox{\linewidth}{!}{
			\begin{tabular}{c | c c c}
				\toprule
				Methods & Params & Memory & Time\\
				\midrule
				Conv & 4.60M & 3.7GB $\times$ 4 & 0.53s\\
				DFN &2.32M & OOM & - \\
				ACDA (two-layer implementation)&2.28M & 4.6GB $\times$ 4 & 0.46s	\\		
				
				\bottomrule
		\end{tabular}}
		\vspace{-2mm}
	\end{center}
\end{table}

\noindent \textbf{Memory and speed.}
We then perform comparisons on memory and speed. 
Ad-ResNet-s, and construct baseline (denoted as Conv) by using standard convolutions only, thus Conv and ACDA in Table~\ref{tab:memo} have the same architectures. 
We train the networks on ImageNet with standard settings, and report the training memory consumption and speed of one iteration. 
The proposed ACDA achieves higher speed without significantly increase the memory footprints comparing to standard convolutions. 
We further present a comparison by adopting per-pixel dynamic filter networks (DFN \cite{dynamic}), where a light-weight network is used to generation adaptive filters that are directly applied to the feature maps.
In practice, training DFN consistently results in out-of-memory (OOM) error, and similar impractical costs are also observed when using CondConv \cite{condconv} and DY-CNN \cite{dycnn} for per-pixel adaptive filters.  
The results indicate that, without the proposed two-layer implementation in our ACDA framework, applying pixel-wise adaptive convolutions is prohibitive as it involves the multiplications between very high dimensional tensors. 
The proposed ACDA successfully addresses this challenge by decomposing the prohibitive multiplication into two mild-size multiplications as quantitatively validated in Table~\ref{tab:memo}.  
And a faster speed is observed thanks to the reduced computation.

\section{Conclusion}
In this paper, we introduced adaptive convolutions with dynamic filter atoms, plug-and-play replacements to convolution layers to better model intra-image variance.
The convolutional filters in ACDA are adaptively generated from local feature.
We decomposed adaptive filters over dynamically generated  atoms to significantly save in parameter and memory.
We further decomposed atoms over multi-scale bases for adaptive receptive fields.
We empirically validated our approach on image classification, crowd counting, and real-world image restorations. 

\section{Acknowledgements}
This work was supported by the DARPA TAMI program.

{\small
	\bibliographystyle{ieee_fullname}
	\bibliography{egbib}

\begin{thebibliography}{10}\itemsep=-1pt

\bibitem{sidd}
Abdelrahman Abdelhamed, Stephen Lin, and Michael~S Brown.
\newblock A high-quality denoising dataset for smartphone cameras.
\newblock In {\em Proceedings of the IEEE Conference on Computer Vision and
  Pattern Recognition}, pages 1692--1700, 2018.

\bibitem{fastweight}
Jimmy Ba, Geoffrey Hinton, Volodymyr Mnih, Joel~Z Leibo, and Catalin Ionescu.
\newblock Using fast weights to attend to the recent past.
\newblock {\em Advances in neural information processing systems}, 2016.

\bibitem{local}
Joan Bruna, Wojciech Zaremba, Arthur Szlam, and Yann LeCun.
\newblock Spectral networks and locally connected networks on graphs.
\newblock {\em arXiv preprint arXiv:1312.6203}, 2013.

\bibitem{realsr}
Jianrui Cai, Hui Zeng, Hongwei Yong, Zisheng Cao, and Lei Zhang.
\newblock Toward real-world single image super-resolution: A new benchmark and
  a new model.
\newblock In {\em Proceedings of the IEEE International Conference on Computer
  Vision}, pages 3086--3095, 2019.

\bibitem{sanet}
Xinkun Cao, Zhipeng Wang, Yanyun Zhao, and Fei Su.
\newblock Scale aggregation network for accurate and efficient crowd counting.
\newblock In {\em Proceedings of the European Conference on Computer Vision
  (ECCV)}, pages 734--750, 2018.

\bibitem{dycnn}
Yinpeng Chen, Xiyang Dai, Mengchen Liu, Dongdong Chen, Lu Yuan, and Zicheng
  Liu.
\newblock Dynamic convolution: Attention over convolution kernels.
\newblock In {\em Proceedings of the IEEE/CVF Conference on Computer Vision and
  Pattern Recognition}, pages 11030--11039, 2020.

\bibitem{deforme}
Jifeng Dai, Haozhi Qi, Yuwen Xiong, Yi Li, Guodong Zhang, Han Hu, and Yichen
  Wei.
\newblock Deformable convolutional networks.
\newblock In {\em Proceedings of the IEEE international conference on computer
  vision}, pages 764--773, 2017.

\bibitem{probmeta}
Jonathan Gordon, John Bronskill, Matthias Bauer, Sebastian Nowozin, and
  Richard~E Turner.
\newblock Meta-learning probabilistic inference for prediction.
\newblock {\em arXiv preprint arXiv:1805.09921}, 2018.

\bibitem{cbdnet}
Shi Guo, Zifei Yan, Kai Zhang, Wangmeng Zuo, and Lei Zhang.
\newblock Toward convolutional blind denoising of real photographs.
\newblock In {\em Proceedings of the IEEE Conference on Computer Vision and
  Pattern Recognition}, pages 1712--1722, 2019.

\bibitem{hypernets}
David Ha, Andrew Dai, and Quoc~V Le.
\newblock Hypernetworks.
\newblock {\em arXiv preprint arXiv:1609.09106}, 2016.

\bibitem{resnet}
Kaiming He, Xiangyu Zhang, Shaoqing Ren, and Jian Sun.
\newblock Deep residual learning for image recognition.
\newblock In {\em Proceedings of the IEEE conference on computer vision and
  pattern recognition}, pages 770--778, 2016.

\bibitem{mobilev3}
Andrew Howard, Mark Sandler, Grace Chu, Liang-Chieh Chen, Bo Chen, Mingxing
  Tan, Weijun Wang, Yukun Zhu, Ruoming Pang, Vijay Vasudevan, et~al.
\newblock Searching for mobilenetv3.
\newblock In {\em Proceedings of the IEEE International Conference on Computer
  Vision}, pages 1314--1324, 2019.

\bibitem{ucf}
Haroon Idrees, Muhmmad Tayyab, Kishan Athrey, Dong Zhang, Somaya Al-Maadeed,
  Nasir Rajpoot, and Mubarak Shah.
\newblock Composition loss for counting, density map estimation and
  localization in dense crowds.
\newblock In {\em Proceedings of the European Conference on Computer Vision
  (ECCV)}, pages 532--546, 2018.

\bibitem{compositional_cc}
Haroon Idrees, Muhmmad Tayyab, Kishan Athrey, Dong Zhang, Somaya Al-Maadeed,
  Nasir Rajpoot, and Mubarak Shah.
\newblock Composition loss for counting, density map estimation and
  localization in dense crowds.
\newblock In {\em Proceedings of the European Conference on Computer Vision
  (ECCV)}, pages 532--546, 2018.

\bibitem{jia2017super}
Xu Jia, Hong Chang, and Tinne Tuytelaars.
\newblock Super-resolution with deep adaptive image resampling.
\newblock {\em arXiv preprint arXiv:1712.06463}, 2017.

\bibitem{dynamic}
Xu Jia, Bert De~Brabandere, Tinne Tuytelaars, and Luc~V Gool.
\newblock Dynamic filter networks.
\newblock In {\em Advances in neural information processing systems}, pages
  667--675, 2016.

\bibitem{kim2016accurate}
Jiwon Kim, Jung Kwon~Lee, and Kyoung Mu~Lee.
\newblock Accurate image super-resolution using very deep convolutional
  networks.
\newblock In {\em Proceedings of the IEEE conference on computer vision and
  pattern recognition}, pages 1646--1654, 2016.

\bibitem{lenet}
Yann LeCun, L{\'e}on Bottou, Yoshua Bengio, and Patrick Haffner.
\newblock Gradient-based learning applied to document recognition.
\newblock {\em Proceedings of the IEEE}, 86(11):2278--2324, 1998.

\bibitem{ledig2017photo}
Christian Ledig, Lucas Theis, Ferenc Husz{\'a}r, Jose Caballero, Andrew
  Cunningham, Alejandro Acosta, Andrew Aitken, Alykhan Tejani, Johannes Totz,
  Zehan Wang, et~al.
\newblock Photo-realistic single image super-resolution using a generative
  adversarial network.
\newblock In {\em Proceedings of the IEEE conference on computer vision and
  pattern recognition}, pages 4681--4690, 2017.

\bibitem{enhanced}
Bee Lim, Sanghyun Son, Heewon Kim, Seungjun Nah, and Kyoung Mu~Lee.
\newblock Enhanced deep residual networks for single image super-resolution.
\newblock In {\em Proceedings of the IEEE conference on computer vision and
  pattern recognition workshops}, pages 136--144, 2017.

\bibitem{context}
Weizhe Liu, Mathieu Salzmann, and Pascal Fua.
\newblock Context-aware crowd counting.
\newblock In {\em Proceedings of the IEEE Conference on Computer Vision and
  Pattern Recognition}, pages 5099--5108, 2019.

\bibitem{bl}
Zhiheng Ma, Xing Wei, Xiaopeng Hong, and Yihong Gong.
\newblock Bayesian loss for crowd count estimation with point supervision.
\newblock In {\em Proceedings of the IEEE International Conference on Computer
  Vision}, pages 6142--6151, 2019.

\bibitem{miconi2020backpropamine}
Thomas Miconi, Aditya Rawal, Jeff Clune, and Kenneth~O Stanley.
\newblock Backpropamine: training self-modifying neural networks with
  differentiable neuromodulated plasticity.
\newblock {\em International Conference on Learning Representations}, 2019.

\bibitem{diffpla}
Thomas Miconi, Kenneth Stanley, and Jeff Clune.
\newblock Differentiable plasticity: training plastic neural networks with
  backpropagation.
\newblock In {\em International Conference on Machine Learning}, pages
  3559--3568. PMLR, 2018.

\bibitem{kpn}
Ben Mildenhall, Jonathan~T Barron, Jiawen Chen, Dillon Sharlet, Ren Ng, and
  Robert Carroll.
\newblock Burst denoising with kernel prediction networks.
\newblock In {\em Proceedings of the IEEE Conference on Computer Vision and
  Pattern Recognition}, pages 2502--2510, 2018.

\bibitem{parafromacts}
Siyuan Qiao, Chenxi Liu, Wei Shen, and Alan~L Yuille.
\newblock Few-shot image recognition by predicting parameters from activations.
\newblock In {\em Proceedings of the IEEE Conference on Computer Vision and
  Pattern Recognition}, pages 7229--7238, 2018.

\bibitem{qiu2018dcfnet}
Qiang Qiu, Xiuyuan Cheng, Robert Calderbank, and Guillermo Sapiro.
\newblock {DCFNet}: Deep neural network with decomposed convolutional filters.
\newblock {\em International Conference on Machine Learning}, 2018.

\bibitem{iccnn}
Viresh Ranjan, Hieu Le, and Minh Hoai.
\newblock Iterative crowd counting.
\newblock In {\em Proceedings of the European Conference on Computer Vision
  (ECCV)}, pages 270--285, 2018.

\bibitem{switch}
Deepak~Babu Sam, Shiv Surya, and R~Venkatesh Babu.
\newblock Switching convolutional neural network for crowd counting.
\newblock In {\em 2017 IEEE Conference on Computer Vision and Pattern
  Recognition (CVPR)}, pages 4031--4039. IEEE, 2017.

\bibitem{schlag2021learning}
Imanol Schlag, Tsendsuren Munkhdalai, and J{\"u}rgen Schmidhuber.
\newblock Learning associative inference using fast weight memory.
\newblock In {\em International Conference on Learning Representations}, 2021.

\bibitem{gated}
Imanol Schlag and J{\"u}rgen Schmidhuber.
\newblock Gated fast weights for on-the-fly neural program generation.
\newblock In {\em NeurIPS Metalearning Workshop}, 2017.

\bibitem{perspective}
Miaojing Shi, Zhaohui Yang, Chao Xu, and Qijun Chen.
\newblock Revisiting perspective information for efficient crowd counting.
\newblock In {\em Proceedings of the IEEE Conference on Computer Vision and
  Pattern Recognition}, pages 7279--7288, 2019.

\bibitem{su2019pixel}
Hang Su, Varun Jampani, Deqing Sun, Orazio Gallo, Erik Learned-Miller, and Jan
  Kautz.
\newblock Pixel-adaptive convolutional neural networks.
\newblock In {\em IEEE Conference on Computer Vision and Pattern Recognition},
  pages 11166--11175, 2019.

\bibitem{addense}
Jia Wan and Antoni Chan.
\newblock Adaptive density map generation for crowd counting.
\newblock In {\em Proceedings of the IEEE International Conference on Computer
  Vision}, pages 1130--1139, 2019.

\bibitem{synthetic}
Qi Wang, Junyu Gao, Wei Lin, and Yuan Yuan.
\newblock Learning from synthetic data for crowd counting in the wild.
\newblock In {\em Proceedings of the IEEE conference on computer vision and
  pattern recognition}, pages 8198--8207, 2019.

\bibitem{ssim}
Zhou Wang, Alan~C Bovik, Hamid~R Sheikh, and Eero~P Simoncelli.
\newblock Image quality assessment: from error visibility to structural
  similarity.
\newblock {\em IEEE transactions on image processing}, 13(4):600--612, 2004.

\bibitem{basisgan}
Ze Wang, Xiuyuan Cheng, Guillermo Sapiro, and Qiang Qiu.
\newblock Stochastic conditional generative networks with basis decomposition.
\newblock In {\em ICLR}, 2020.

\bibitem{wangcc}
Ze Wang, Zehao Xiao, Kai Xie, Qiang Qiu, Xiantong Zhen, and Xianbin Cao.
\newblock In defense of single-column networks for crowd counting.
\newblock {\em arXiv preprint arXiv:1808.06133}, 2018.

\bibitem{unified}
Yu-Syuan Xu, Shou-Yao~Roy Tseng, Yu Tseng, Hsien-Kai Kuo, and Yi-Min Tsai.
\newblock Unified dynamic convolutional network for super-resolution with
  variational degradations.
\newblock In {\em Proceedings of the IEEE Conference on Computer Vision and
  Pattern Recognition}, pages 12496--12505, 2020.

\bibitem{condconv}
Brandon Yang, Gabriel Bender, Quoc~V Le, and Jiquan Ngiam.
\newblock Condconv: Conditionally parameterized convolutions for efficient
  inference.
\newblock {\em Advances in neural information processing systems}, 2019.

\bibitem{vdn}
Zongsheng Yue, Hongwei Yong, Qian Zhao, Deyu Meng, and Lei Zhang.
\newblock Variational denoising network: Toward blind noise modeling and
  removal.
\newblock In {\em Advances in neural information processing systems}, pages
  1690--1701, 2019.

\bibitem{dncnn}
Kai Zhang, Wangmeng Zuo, Yunjin Chen, Deyu Meng, and Lei Zhang.
\newblock Beyond a gaussian denoiser: Residual learning of deep cnn for image
  denoising.
\newblock {\em IEEE Transactions on Image Processing}, 26(7):3142--3155, 2017.

\bibitem{zhang2018image}
Yulun Zhang, Kunpeng Li, Kai Li, Lichen Wang, Bineng Zhong, and Yun Fu.
\newblock Image super-resolution using very deep residual channel attention
  networks.
\newblock In {\em Proceedings of the European Conference on Computer Vision
  (ECCV)}, pages 286--301, 2018.

\bibitem{shanghai}
Yingying Zhang, Desen Zhou, Siqin Chen, Shenghua Gao, and Yi Ma.
\newblock Single-image crowd counting via multi-column convolutional neural
  network.
\newblock In {\em Proceedings of the IEEE conference on computer vision and
  pattern recognition}, pages 589--597, 2016.

\bibitem{single}
Yingying Zhang, Desen Zhou, Siqin Chen, Shenghua Gao, and Yi Ma.
\newblock Single-image crowd counting via multi-column convolutional neural
  network.
\newblock In {\em Proceedings of the IEEE conference on computer vision and
  pattern recognition}, pages 589--597, 2016.

\bibitem{zhu2019deformable}
Xizhou Zhu, Han Hu, Stephen Lin, and Jifeng Dai.
\newblock Deformable convnets v2: More deformable, better results.
\newblock In {\em Proceedings of the IEEE Conference on Computer Vision and
  Pattern Recognition}, pages 9308--9316, 2019.

\end{thebibliography}
}

\clearpage
\appendix
\onecolumn 
\newcount\iccvrulercount
\setcounter{page}{1}
\section*{Appendix}
\setcounter{table}{0}
\renewcommand{\thetable}{\Alph{table}}

\setcounter{figure}{0}
\renewcommand{\thefigure}{\Alph{figure}}

\section{Implementation Details}
In this section, we introduce in detail the implementations of ACDA.
We use PyTorch for all the implementations. In practice, the per-pixel adaptive convolutions can be efficiently implemented by first reorganizing input features with the \textit{Unfold} function in PyTorch to extract sliding local blocks from a batched input tensor. 
And then perform corresponding multiplications between sliding local blocks and the adaptive filters (dynamic atoms). All experiments are conducted on the same machine with 4$\times$2080Ti GPUs. The code will be released upon acceptance. 
\subsection{Visualization of filter decomposition}
\begin{figure}[h]
	\centering
	\includegraphics[width=0.5\linewidth]{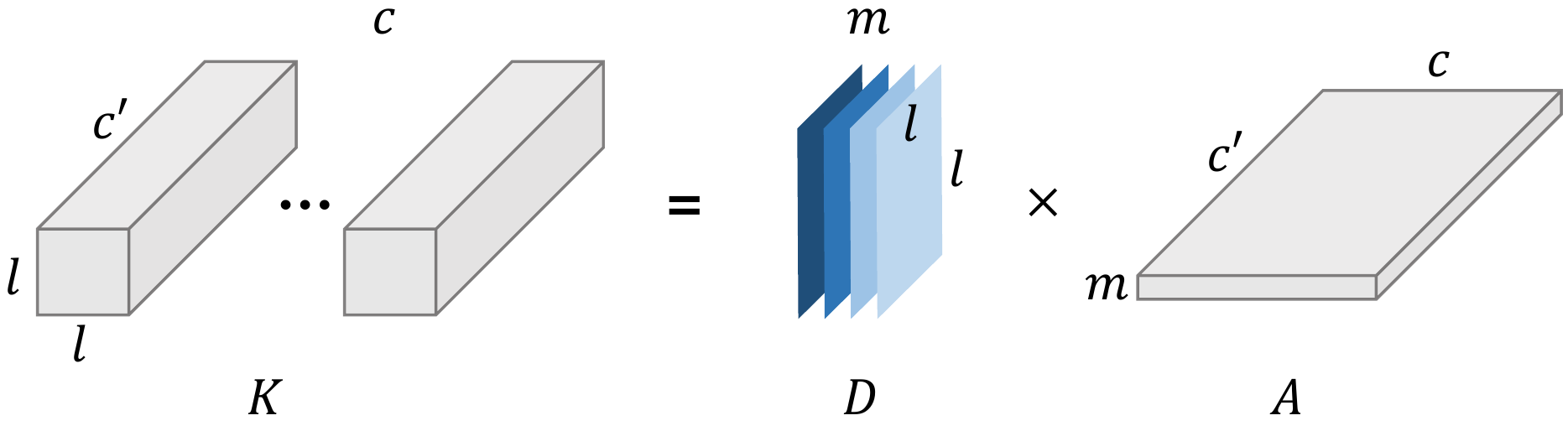}
	\caption{Illustration of decomposed convolutional filters with $m=4$ dictionary atoms. A convolutional filter $\filter \in \R^{c\times c^\prime \times l \times l}$ is decomposed over $m$ filter atoms $\atom \in \R^{m\times l \times l}$, linearly combined by coefficients $\coef \in \R^{c\times c^\prime \times m}$. }
	\label{fig:dcf}
\end{figure}

We provide a visualization of decomposing a convolutional filter $K$ over $m=4$ basis elements in Figure~\ref{fig:dcf}.

\subsection{Visualization of Multi-scale Fourier-Bessel Bases}
\begin{figure}[h]
	\centering
	\includegraphics[width=0.6\linewidth]{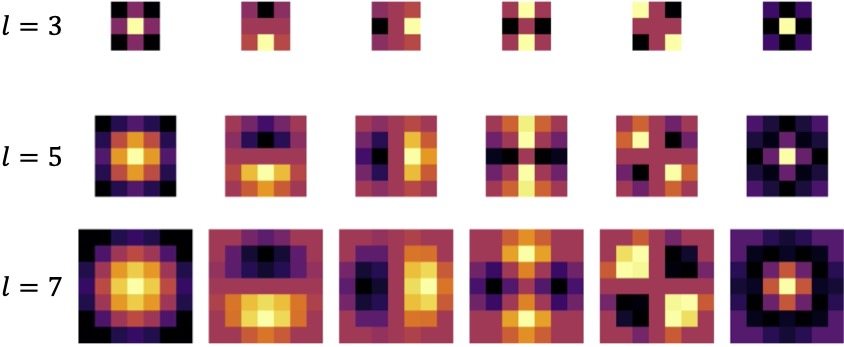}
	\caption{Visualization of the first six Fourier-Bessel basis elements at three scales of $l=3$, $l=5$, and $l=7$.}
	\label{fig:fb}
\end{figure}

Fourier-Bessel bases at three scales are visualized in Figure~\ref{fig:fb}. In practice, we truncate the first 6 basis elements at each scale. The small-size bases are padded with $0$ to match the spatial size of the largest bases, which allow parallel multiplications and convolutions.

\subsection{Convolutional Atom Generation Networks}
\label{phi}
To generate the dynamic atoms for per-pixel adaptive convolutional filters,
we parametrize the generation network $\Phi$ as a light weight 2-layer convolutional neural network.
The first layer uses $1 \times 1$ convolution to shrink the channel dimension of the input feature to 64 channels, and a $3\times3$ convolutional layer is then followed to output the prediction of dynamic atoms $\atom$ or basis coefficients $\alpha$, we consistently observe that this configuration delivers good results with very low cost, while increasing the size of $\Phi$ does not influence the performance significantly.

\subsection{Network Structures of Ad-ResNets}
\label{adres}
\paragraph{Dynamic bottleneck blocks}
Following ResNet \cite{resnet}, we construct dynamic bottleneck blocks by using $1\times 1$ convolutional layers to squeeze the restore the channel dimensions of feature maps, and applying ACDA on the intermediate features with fewer channels.
This configuration fully demonstrates the advantage of ACDA to alleviate the demand for filters with high channel numbers.
Therefore, networks with ACDA usually enjoy more compact model sizes with fewer parameters. 
We use ACDA with atom bases, which include three scales of Fourier-Bessel bases. As shown in Appendix Figure~\ref{fig:fb}, 
we use $m^\prime=6$ and $S = 3$, i.e., three sets of Fourier-Bessel bases with 6 basis elements in each set. We empirically observe that this configuration with only 18 atom basis elements is sufficient to deliver satisfying performance while only requires a 18-dim vector to be predicted at each spatial position by the dynamic atom generation network to reconstruct a dynamic filter atom. 
At each layer, we adopt a tiny two-layer CNN as the dynamic atom generation network, whose architecture is detailed in Appendix Section~\ref{phi}.
In practice, we use 6 dynamic filter atoms, i.e., $m=6$, so that the dynamic atom generation network output a $6 \times 18 = 108$-dim vector at each spatial position.

\paragraph{Ad-ResNet}
Without heavily tuning the network architectures, we follow ResNets and construct two variants of Ad-ResNet. The detailed network configurations are presented in Table~\ref{tab:adres}.
In crowd counting experiments, the first `Max pool' layer and the final `Output' layer are removed. Consecutive transposed convolution layers with batch normalization and ReLU activation are used for resolution recovery.

\newcommand{\blocka}[2]{\multirow{3}{*}{\(\left[\begin{array}{c}\text{Conv, 3$\times$3, #1}\\[-.1em] \text{3$\times$3, #1} \end{array}\right]\)$\times$#2}
}
\newcommand{\blockb}[3]{\multirow{3}{*}{\(\left[\begin{array}{c}\text{Conv, 1$\times$1, #2}\\[-.1em] \text{Conv, 3$\times$3, #2}\\[-.1em] \text{Conv, 1$\times$1, #1}\end{array}\right]\)$\times$#3}
}

\newcommand{\blockad}[3]{\multirow{3}{*}{\(\left[\begin{array}{c}\text{Conv, 1$\times$1, #2}\\[-.1em] \text{ACDA, 7$\times$7, #2}\\[-.1em] \text{Conv, 1$\times$1, #1}\end{array}\right]\)$\times$#3}
}

\newcommand{\blockads}[3]{\multirow{3}{*}{\(\left[\begin{array}{c}\text{Conv, 1$\times$1, #2}\\[-.1em] \text{ACDA, 5$\times$5, #2}\\[-.1em] \text{Conv, 1$\times$1, #1}\end{array}\right]\)$\times$#3}
}

\begin{table}[t]
	\begin{center} 
		\small
		\centering
		\caption{Network architectures of Ad-ResNets. `\text{Conv, 3$\times$3}, 64' denotes standard $3 \times  3$ convolutional layers with 64 output channels. We use ACDA with Fourier-Bessel atom bases. }
		\label{tab:adres}
		\resizebox{0.75\linewidth}{!}{
			\begin{tabular}{c |c | c | c}
				\toprule
				Layer name & Ad-ResNet-s &  Ad-ResNet-m & Ad-ResNet-l\\
				\midrule
				\multirow{2}{*}{Conv0} & \multicolumn{3}{c}{Conv 7×7, 64, stride 2} \\
				& \multicolumn{3}{c}{Max pool 3×3, stride 2}\\
				\midrule
				\multirow{3}{*}{Conv1\_x} & \blocka{64}{2} & \blockb{256}{64}{2} & \blockb{256}{64}{2}\\
				& & & \\
				& & & \\
				\midrule
				\multirow{3}{*}{Conv2\_x} & \blocka{128}{2} & \blockb{512}{128}{2} & \blockb{512}{128}{3}\\
				& & &\\
				& & &\\
				\midrule
				\multirow{3}{*}{Conv3\_x} & \blockad{256}{128}{2} & \blockad{512}{256}{2} & \blockad{512}{256}{8}\\
				& & &\\
				& & &\\
				\midrule
				\multirow{3}{*}{Conv4\_x} & \blockads{512}{256}{2} & \blockads{1024}{512}{2} & \blockads{1024}{512}{3} \\
				& & &\\
				& & &\\
				\midrule
				\multirow{2}{*}{Output} & \multicolumn{3}{c}{Average pool} \\
				& \multicolumn{3}{c}{1000-d FC layer}\\
				\bottomrule
		\end{tabular}}
	\end{center}
\end{table}

\subsection{Adaptivity and Diversity of Generated Atoms}
\label{diverse}
\begin{figure}
    \centering
    \begin{tabular}{r |c| c| c| c |c}
         \includegraphics[width=0.1\linewidth]{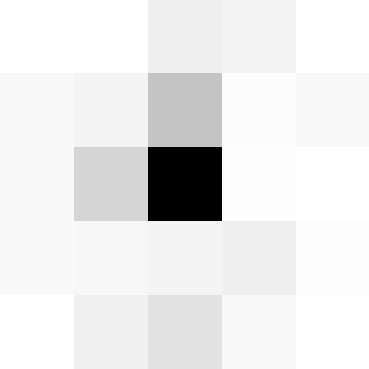}& 
         \includegraphics[width=0.1\linewidth]{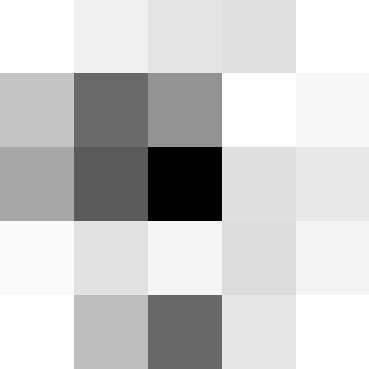}& 
         \includegraphics[width=0.1\linewidth]{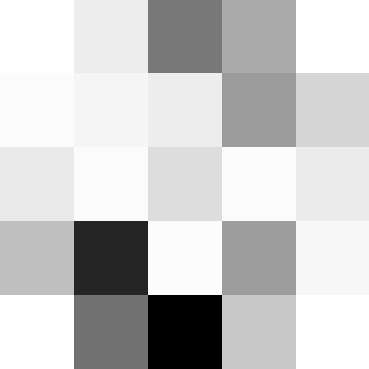}& 
         \includegraphics[width=0.1\linewidth]{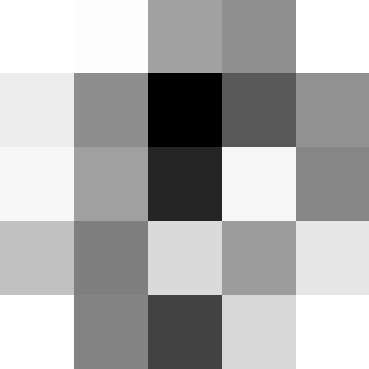}& 
         \includegraphics[width=0.1\linewidth]{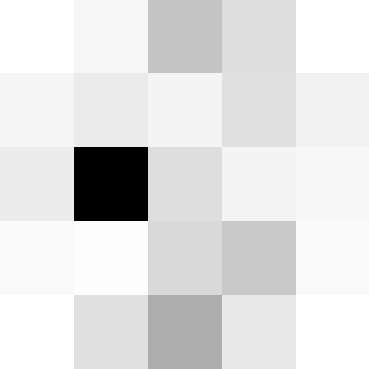}&
         \includegraphics[width=0.1\linewidth]{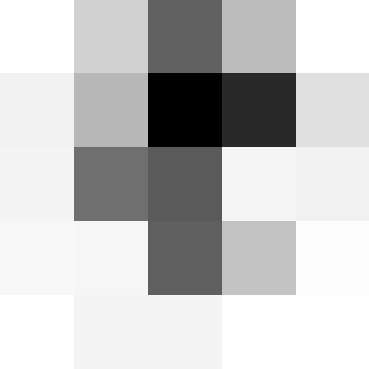}
    \end{tabular}
    \caption{We randomly select a layer and visualize the first predicted bases on a single ImageNet image at different spatial position. The bases are visually diverse.}
    \label{fig:vis}
\end{figure}

To show that the proposed atom generator predicts atoms adaptively, 
we calculate the average pairwise cosine distance of basis coefficients $\boldsymbol{\alpha}$ (averaged across scales) within the spatial position of a single image from ImageNet. On the entire ImageNet validation set, we obtain $0.68\pm0.30$ intra-image cosine distance of $\boldsymbol{\alpha}$, which jointly with Figure~\ref{fig:vis}, show the diversity of the predicted bases (filters) both quantitatively and qualitatively.

\subsection{Ablation Study}
\label{ablation}
We present ablation study on \textit{RealSR Final} $\times 3$ to validate the selection of the number of atoms $m$ and the scales $S$ of Fourier Bessel bases. All comparisons are in Table~\ref{tab:ab}.
$m > 7$ and $S > 3$ lead to minor performance improvements only. For better trade-offs between performance and practical costs, we adopt $m = 6$ and $S = 3$.

\begin{table}[h]
%	\vspace{-3.5mm}
	\begin{center} 
		\centering
		\caption{Ablation study on number of atom $m$ and scales $S$.}
		\label{tab:ab}
		\resizebox{0.9\linewidth}{!}{
			\begin{tabular}{c | c c c c c c c | c c c c c}
				\toprule
				& \multicolumn{7}{c|}{Number of atoms $m$} & \multicolumn{4}{c}{Number of scales $S$} \\
				\midrule
				& m = 3 & m=5 & m=7 & m=9 & m=11 & m=13 & m=15 & S=1 & S=2 & S=3 & S=4  & S=5 \\
				\midrule
				PSNR & 30.52 & 30.68 & 30.73 & 30.74 & 30.75 & 30.73 & 30.74 & 30.63 & 30.73 & 30.72 & 30.70 & 30.69 \\ 
				SSIM & 0.858 & 0.866 & 0.868 & 0.868 & 0.869 & 0.866 & 0.868 & 0.866 & 0.868 & 0.868 & 0.865  & 0.866\\
				\bottomrule
			\end{tabular}
		}
	\end{center}
	%	\vspace{-1mm}
\end{table}

\section{Qualitative Results on Crowd Counting}
\label{supp_cc}
\begin{figure}
	%	\centering
	\resizebox{\linewidth}{!}{%
		\small
		\begin{tabular}{c c c}
			%		\centering
			\hspace{-2mm}
			\includegraphics[width=\CTS]{./figs/results/crowd/000000_28input.jpg}
			\CHSS
			&
			\CHSS
			\includegraphics[width=\CTS]{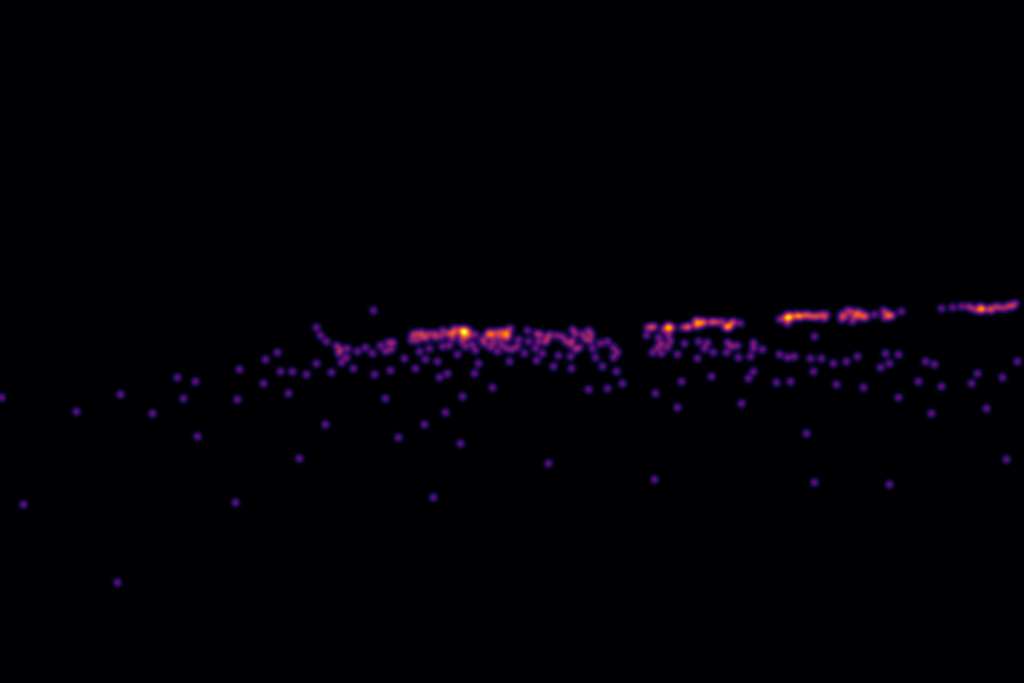}
			\CHSS
			&
			\CHSS
			\includegraphics[width=\CTS]{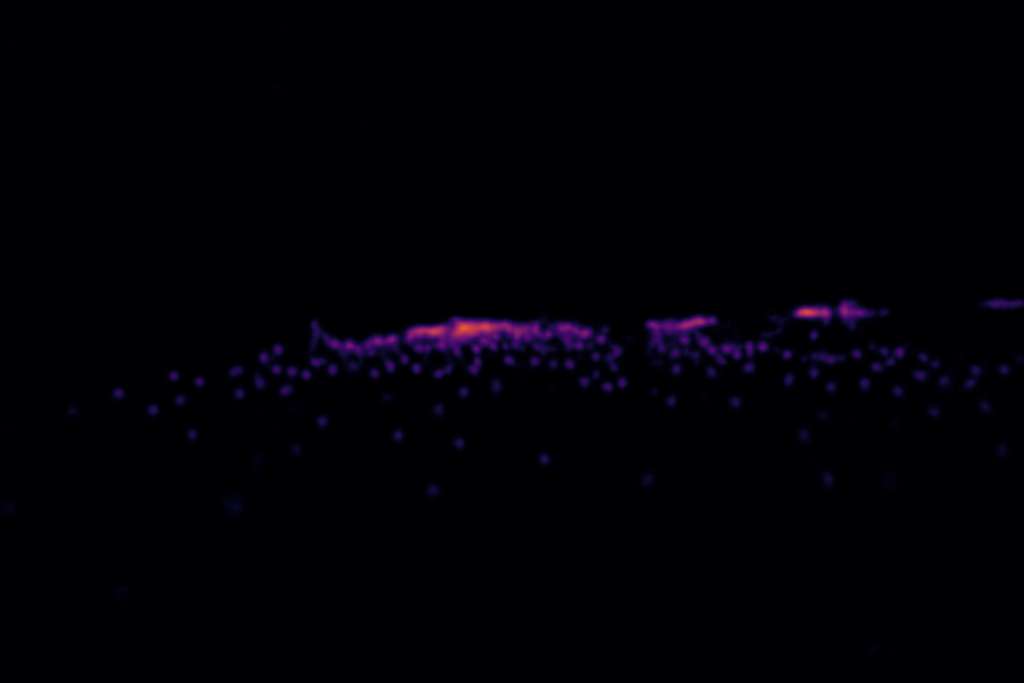}
			\hspace{-2mm}
			\\
			\vspace{1mm}
			
			Input & \hspace{-3mm}GT: 308 & Prediction: 289 \\
			
			\hspace{-2mm}
			\includegraphics[width=\CTS]{./figs/results/crowd/000000_35input.jpg}
			\CHSS
			&
			\CHSS
			\includegraphics[width=\CTS]{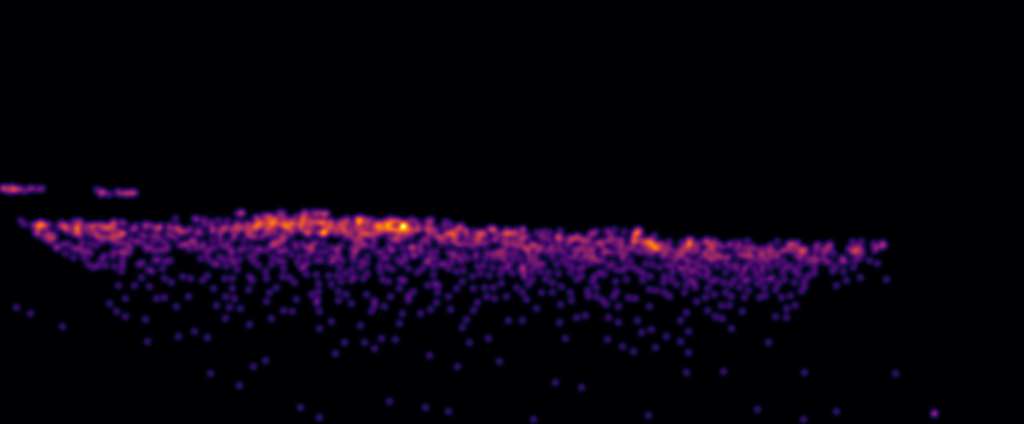}
			\CHSS
			&
			\CHSS
			\includegraphics[width=\CTS]{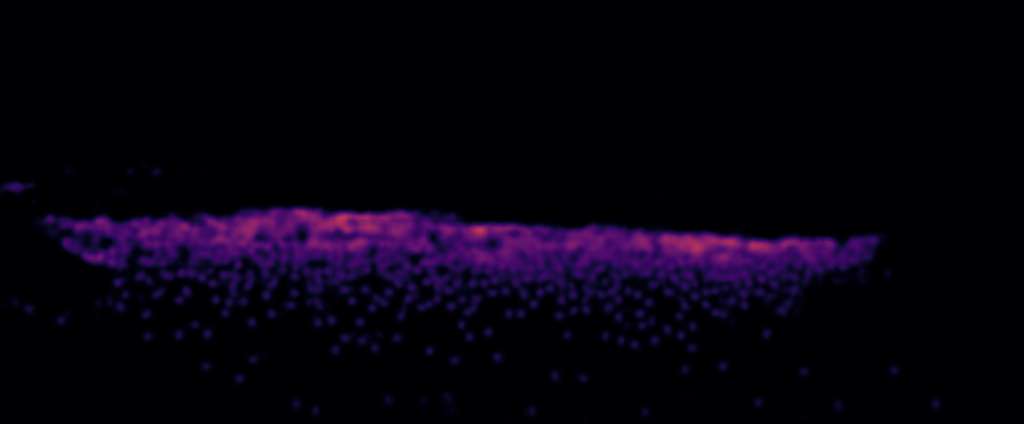}
			\hspace{-2mm}
			\\
			\vspace{1mm}
			
			Input & \hspace{-3mm}GT: 1265 & Prediction: 1179 \\
			
			\hspace{-2mm}
			\includegraphics[width=\CTS]{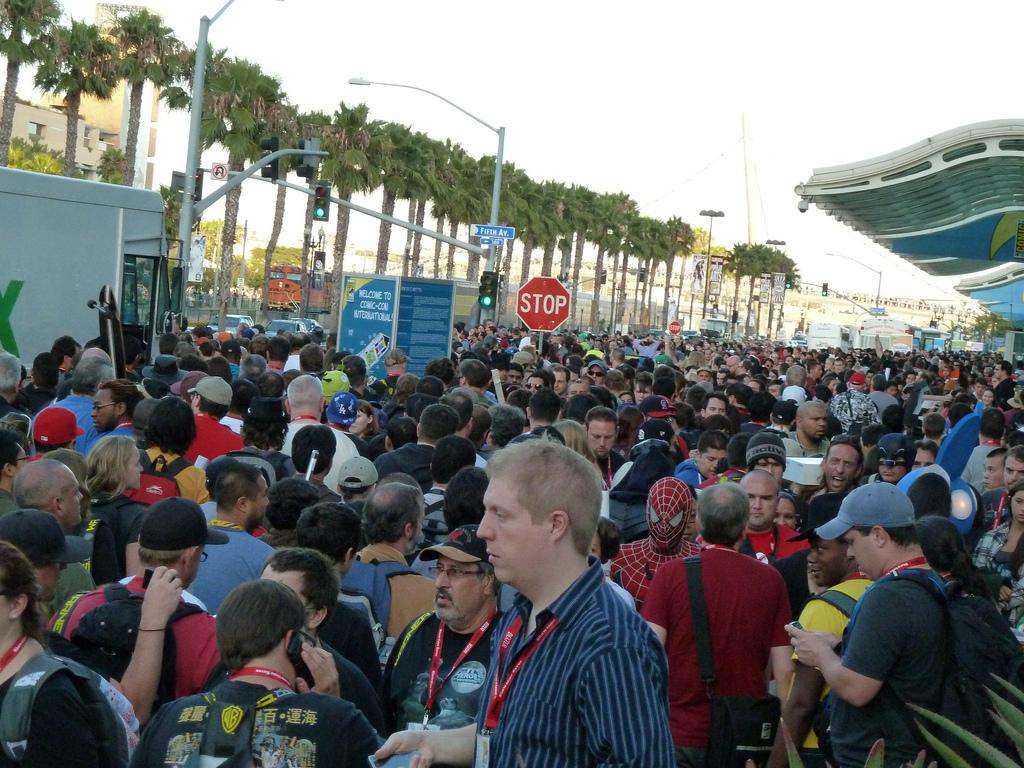}
			\CHSS
			&
			\CHSS
			\includegraphics[width=\CTS]{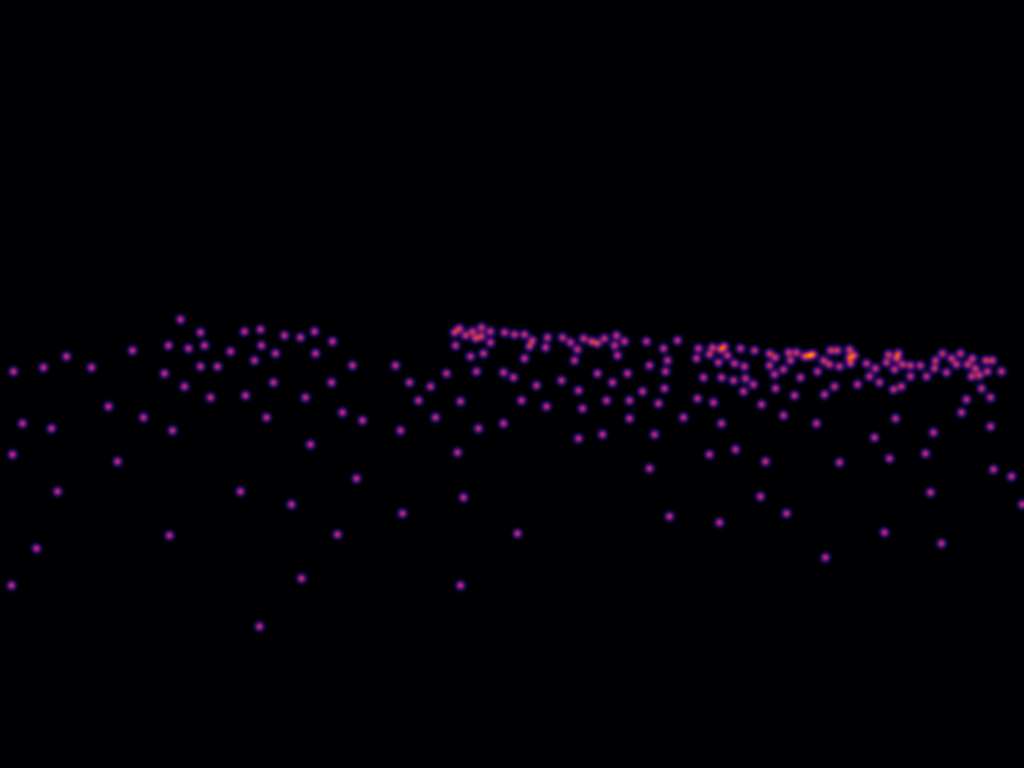}
			\CHSS
			&
			\CHSS
			\includegraphics[width=\CTS]{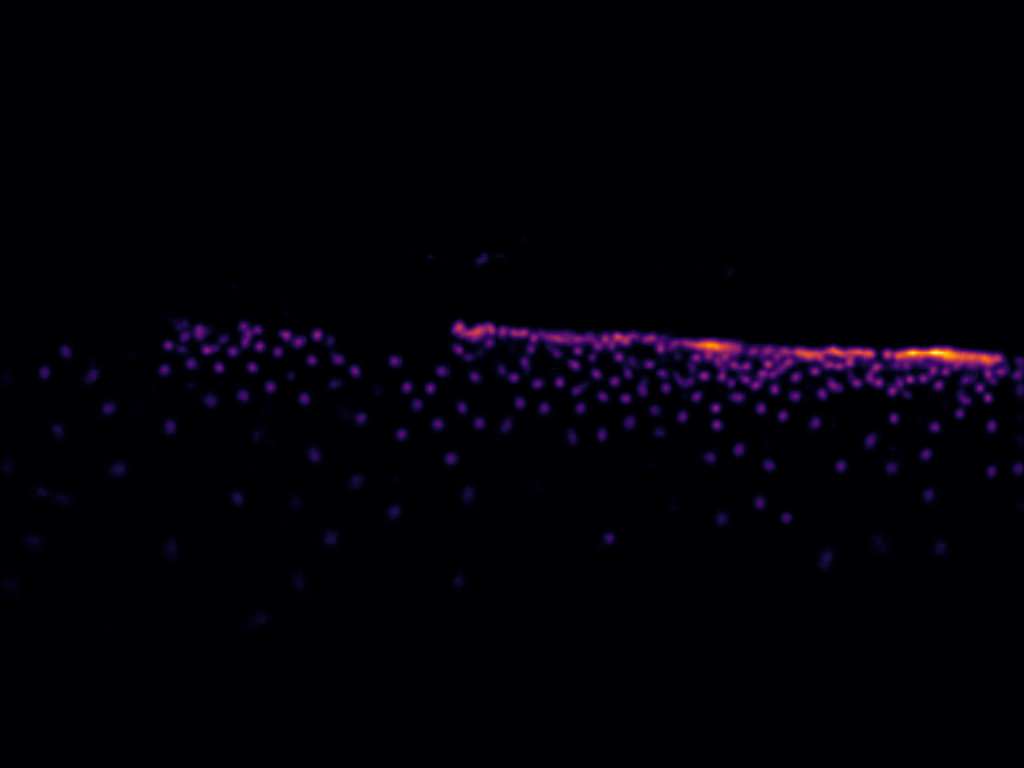}
			\hspace{-2mm}
			\\
			
			Input & \hspace{-3mm}GT: 235 & Prediction: 262 \\
			\vspace{1mm}
			
			\hspace{-2mm}
			\includegraphics[width=\CTS]{./figs/results/crowd/000000_108input.jpg}
			\CHSS
			&
			\CHSS
			\includegraphics[width=\CTS]{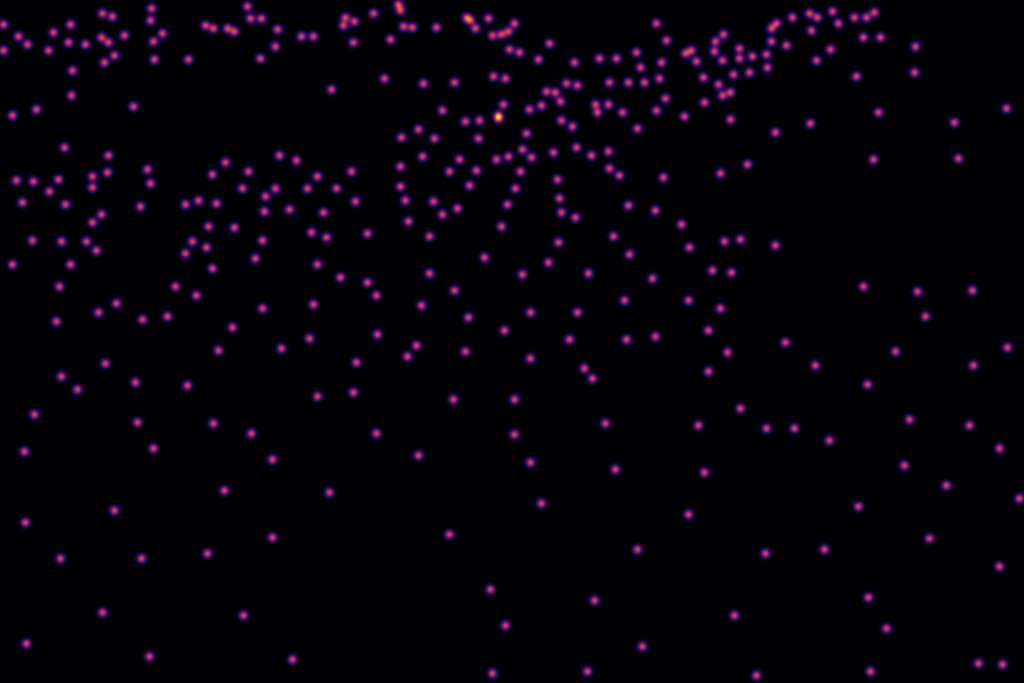}
			\CHSS
			&
			\CHSS
			\includegraphics[width=\CTS]{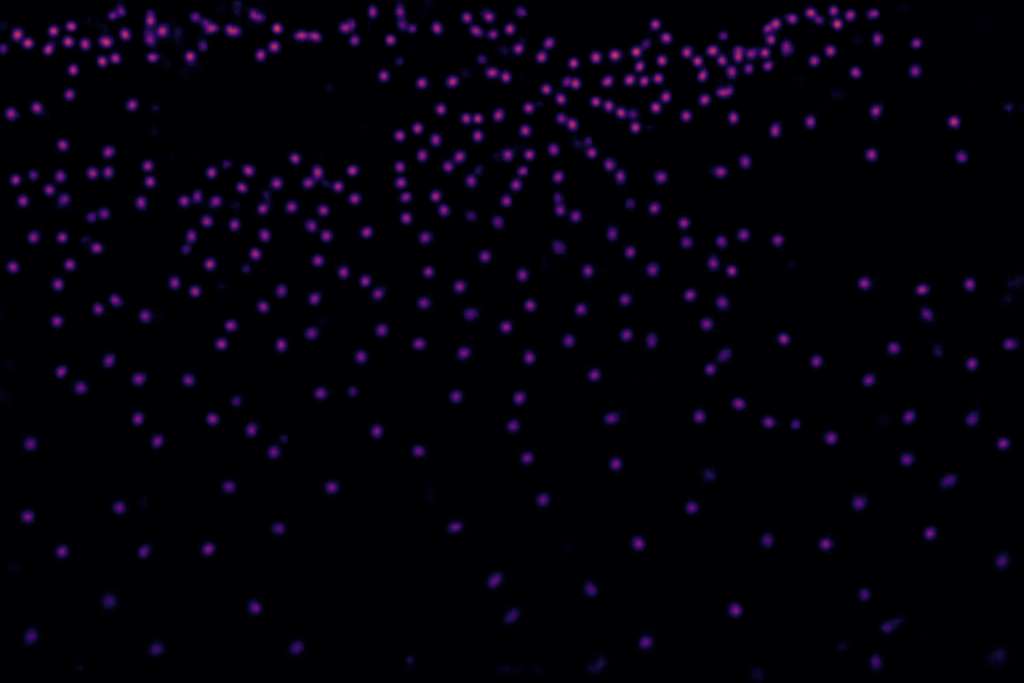}
			\hspace{-2mm}
			\\
			
			Input & \hspace{-3mm}GT: 378 & Prediction: 363 \\
			\vspace{1mm}
			
			\hspace{-2mm}
			\includegraphics[width=\CTS]{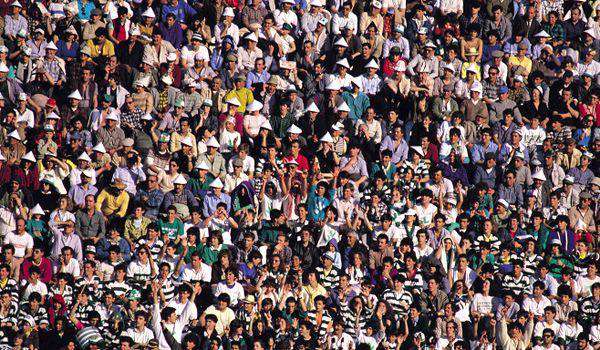}
			\CHSS
			&
			\CHSS
			\includegraphics[width=\CTS]{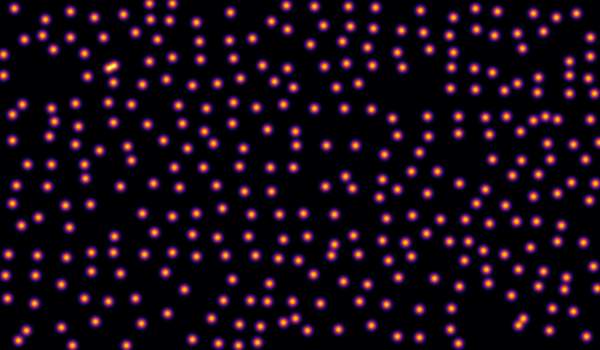}
			\CHSS
			&
			\CHSS
			\includegraphics[width=\CTS]{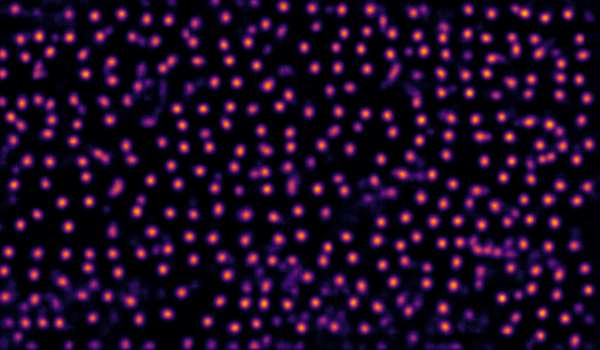}
			\hspace{-2mm}
			\\
			
			Input & \hspace{-3mm}GT: 314 & Prediction: 328 \\

	\end{tabular}}
	%	\vspace{2mm}
	\caption{Visualizations of the atom basis coefficients heatmaps. It is clearly shown that the adaptive convolutions tend to adopt large kernel sizes, i.e., with $7\times 7$ atom bases, when the objects in the target regions have larger spatial sizes, i.e., the closer objects. While $3\times3$ bases are preferred when the dynamic convolutions are applied on regions with dense objects.}
	\label{fig:viscc_supp}
\end{figure}

More qualitative results on crowd counting are presented in Figure~\ref{fig:viscc_supp}.

\section{Qualitative Results on Super-resolution}
\label{supp_sr}
\begin{figure}
	\centering
	\resizebox{0.75\linewidth}{!}{%
		\small
		\begin{tabular}{c c c c}
			%		\centering
			%			\hspace{-2mm}
			
			%			\\
			\includegraphics[width=\CTS]{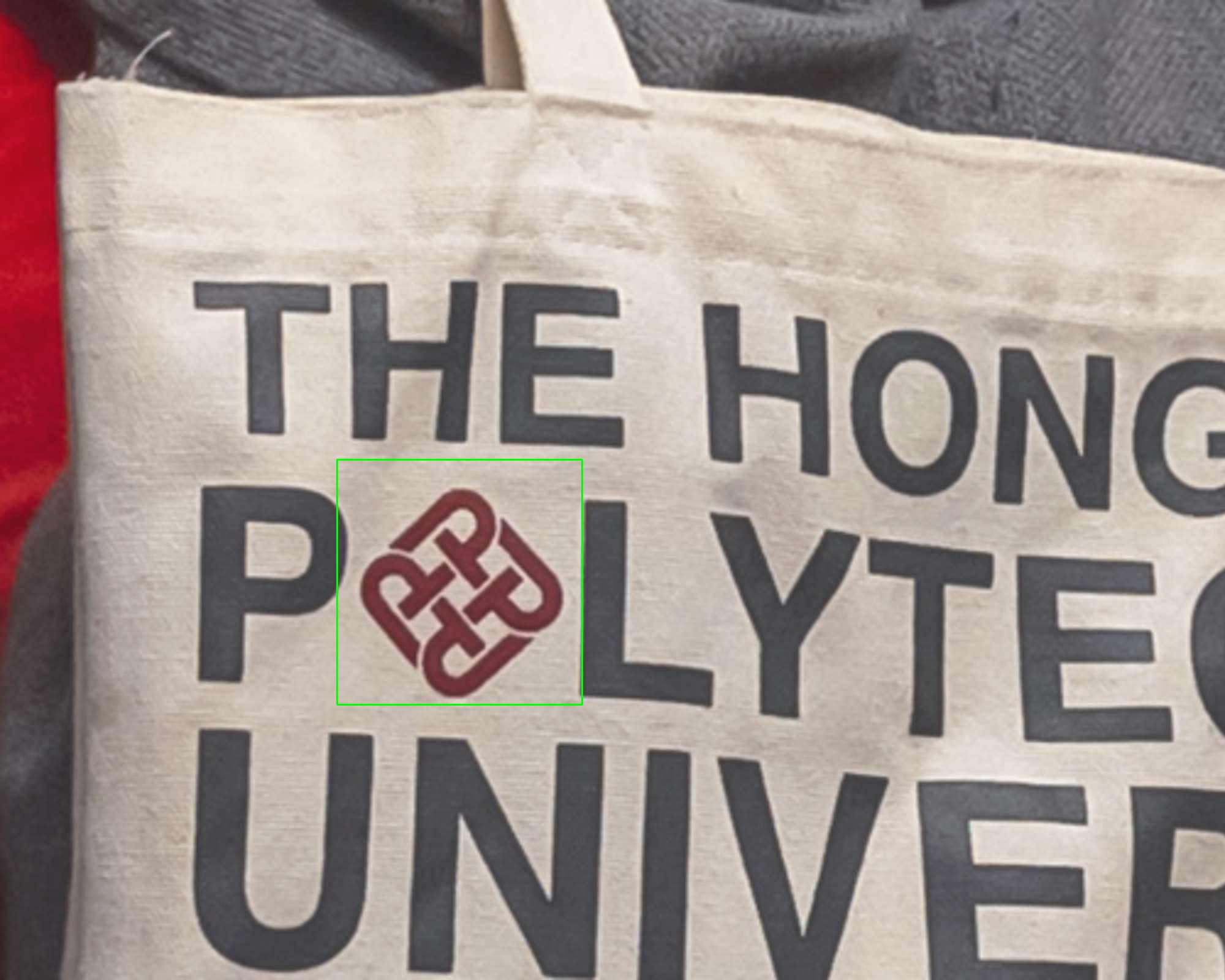}
			\sHSS
			&
			\sHSS
			\includegraphics[width=\CTS]{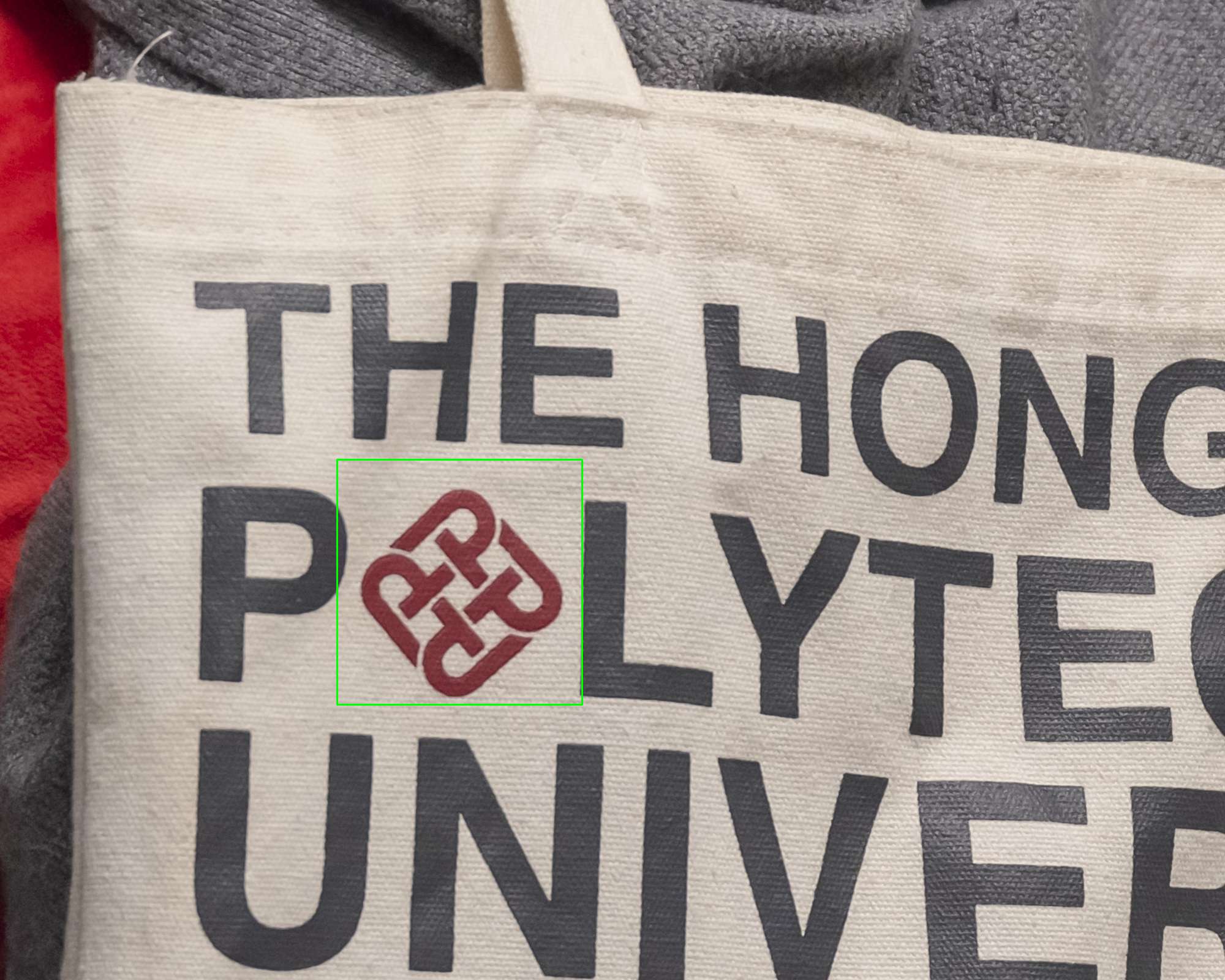}
			\sHSS
			&
			\sHSS
			\includegraphics[width=\CTS]{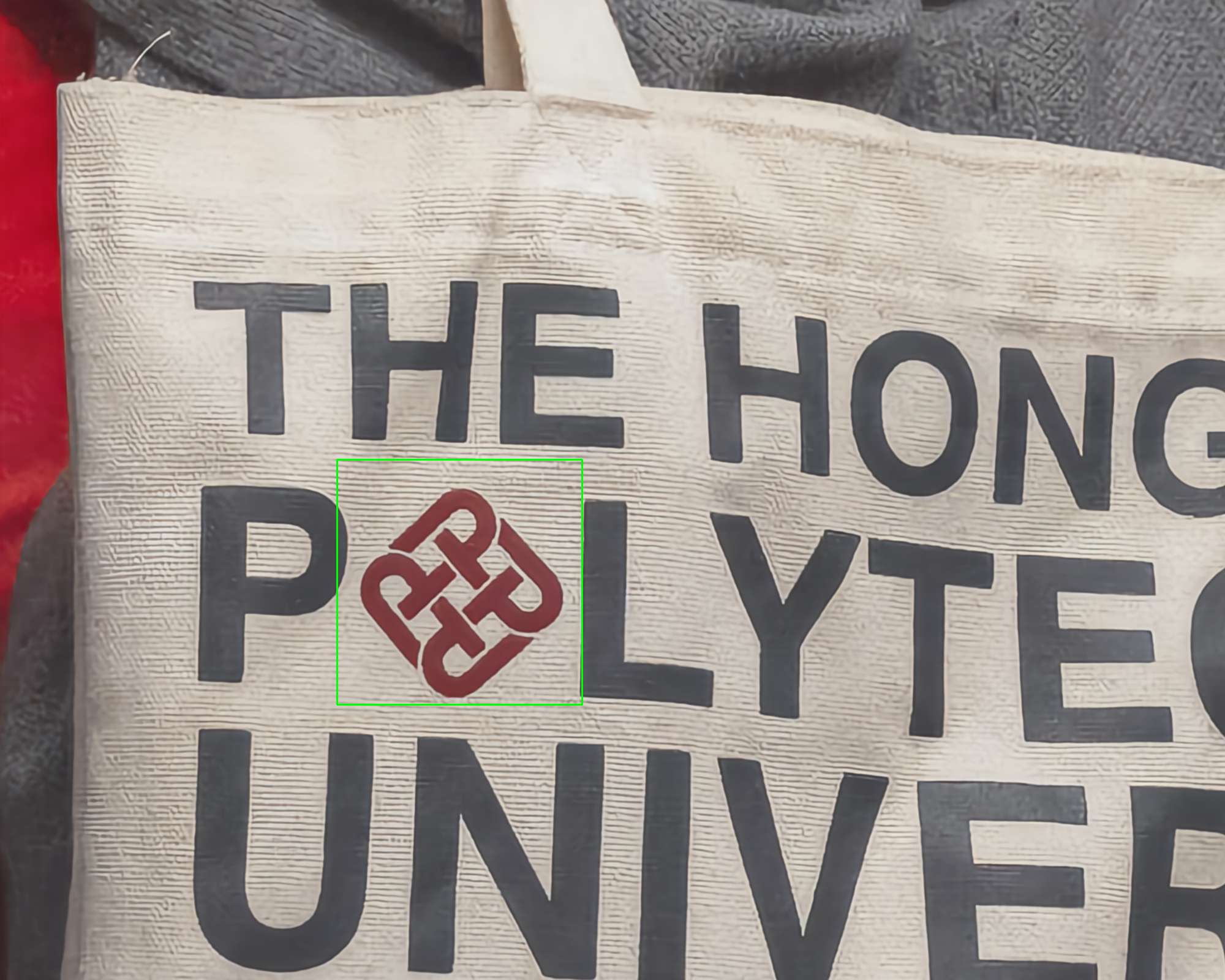}
			\sHSS
			&
			\sHSS
			\includegraphics[width=\CTS]{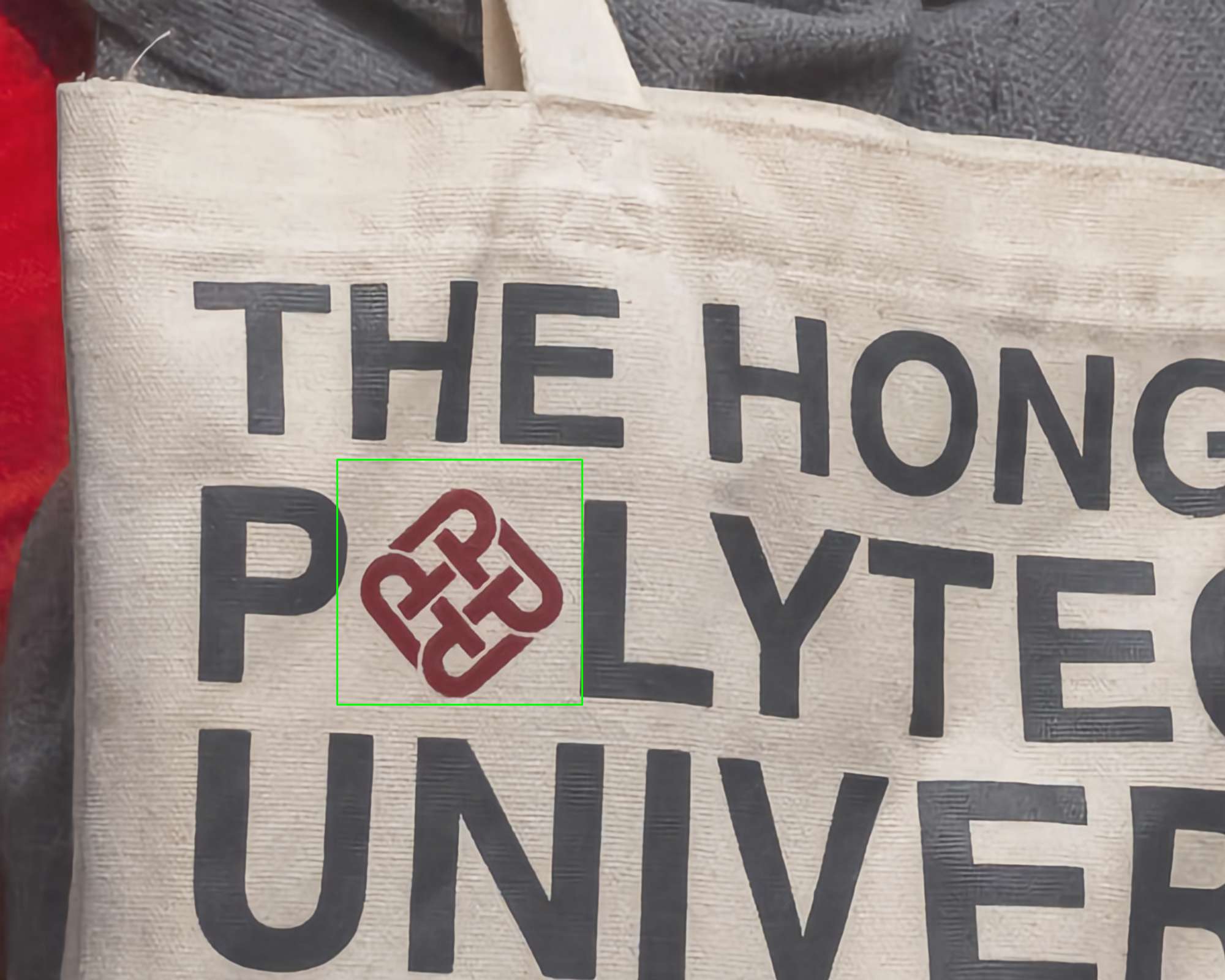}
			\hspace{-2mm}
			\\
			%			\vspace{1mm}
			\includegraphics[width=\CTS]{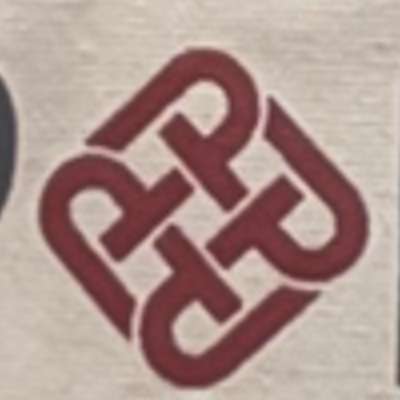}
			\sHSS
			&
			\sHSS
			\includegraphics[width=\CTS]{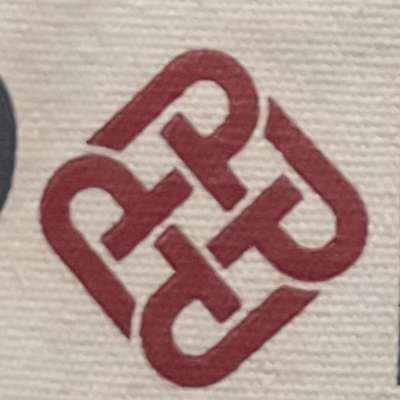}
			\sHSS
			&
			\sHSS
			\includegraphics[width=\CTS]{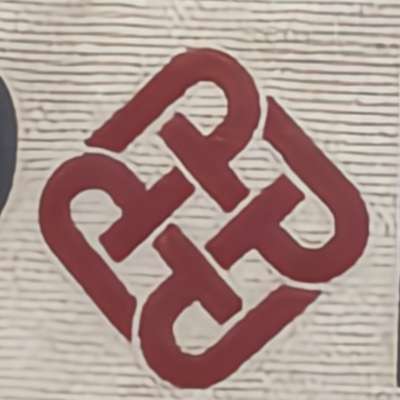}
			\sHSS
			&
			\sHSS
			\includegraphics[width=\CTS]{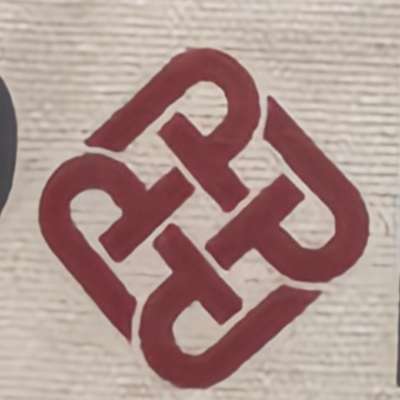}
			\hspace{-2mm}
			\\
			
			Low resolution & High resolution&  LP-KPN & Ours \\
			& & PSNR 22.20 & PSNR 27.56 \\
			
			\\
			%			\\
			\includegraphics[width=\CTS]{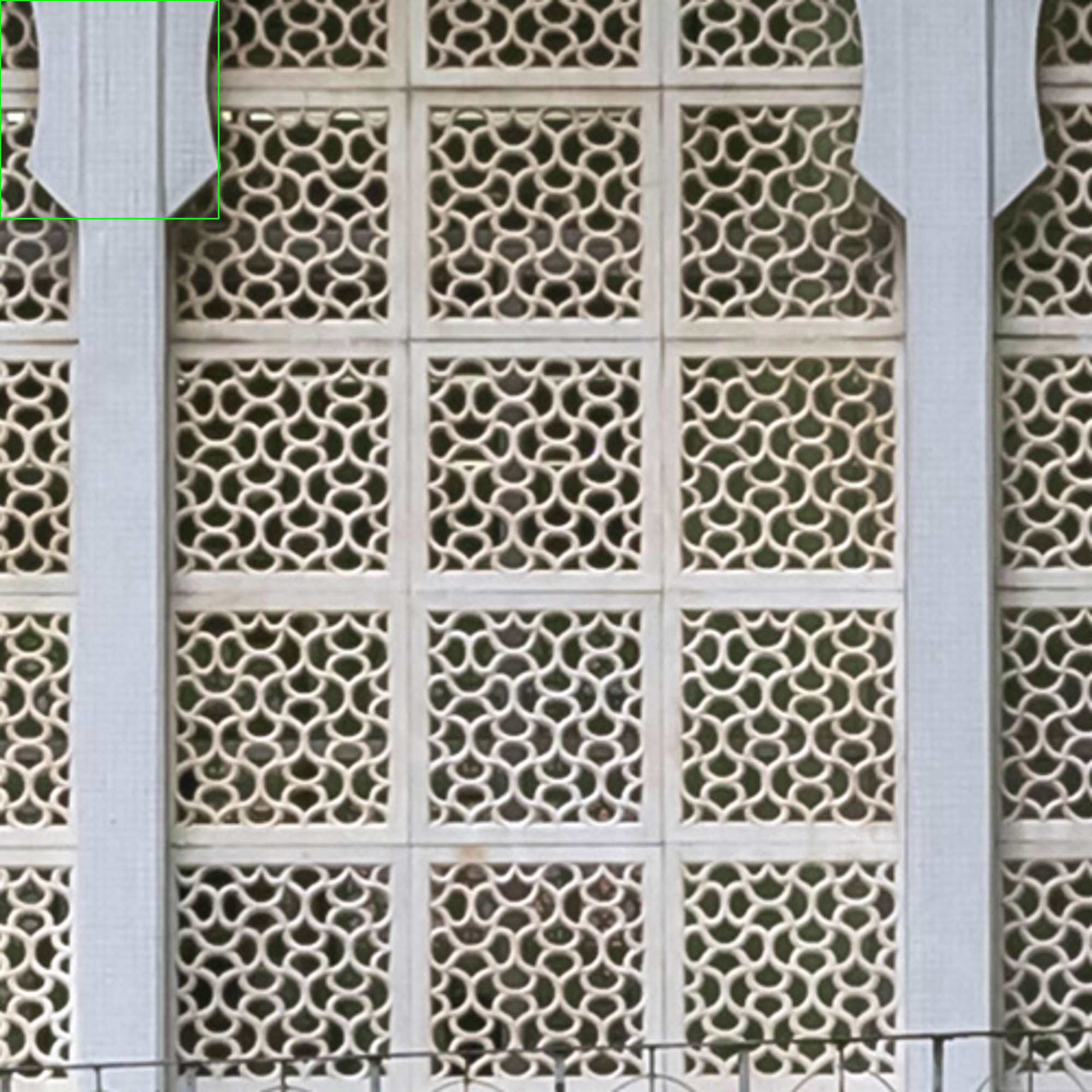}
			\sHSS
			&
			\sHSS
			\includegraphics[width=\CTS]{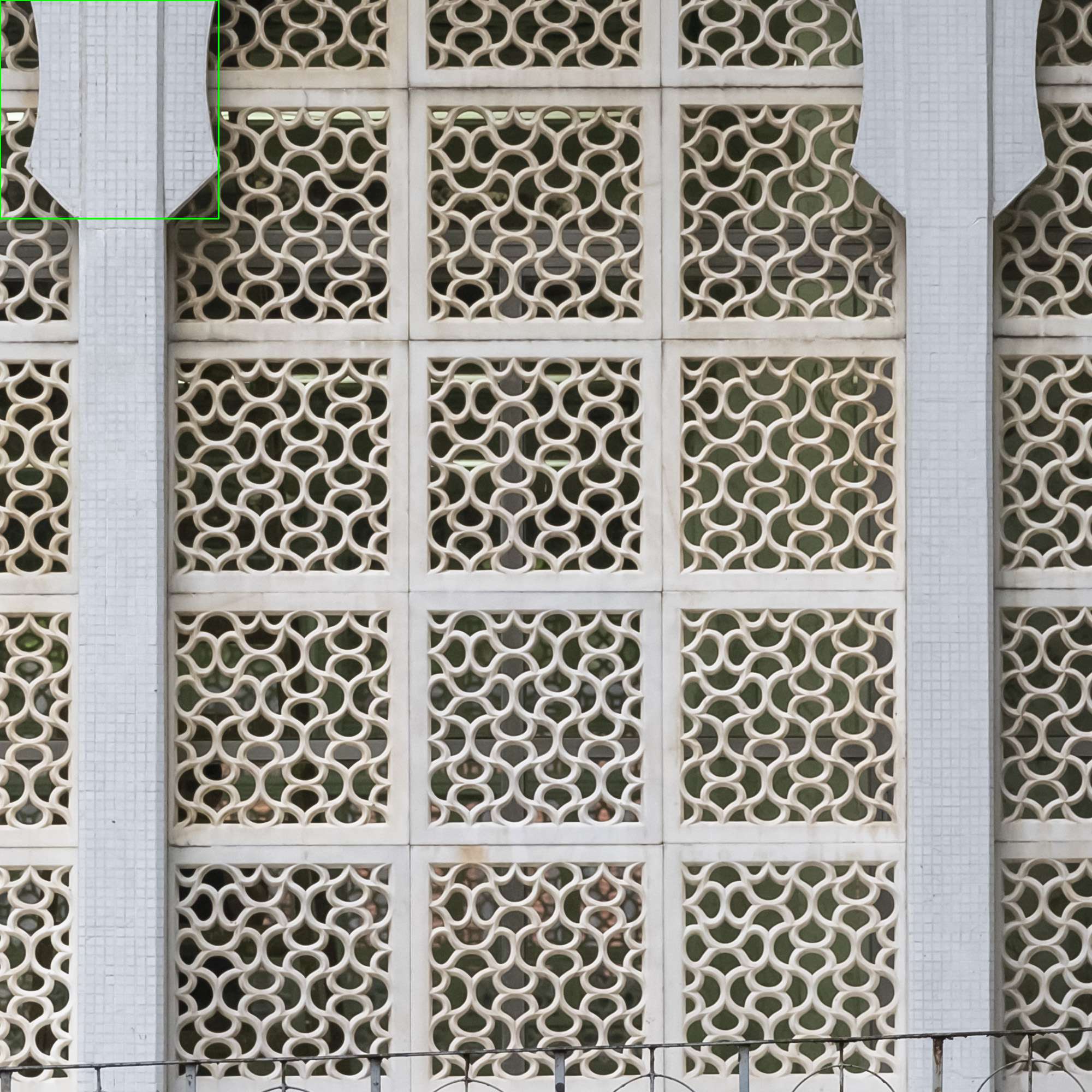}
			\sHSS
			&
			\sHSS
			\includegraphics[width=\CTS]{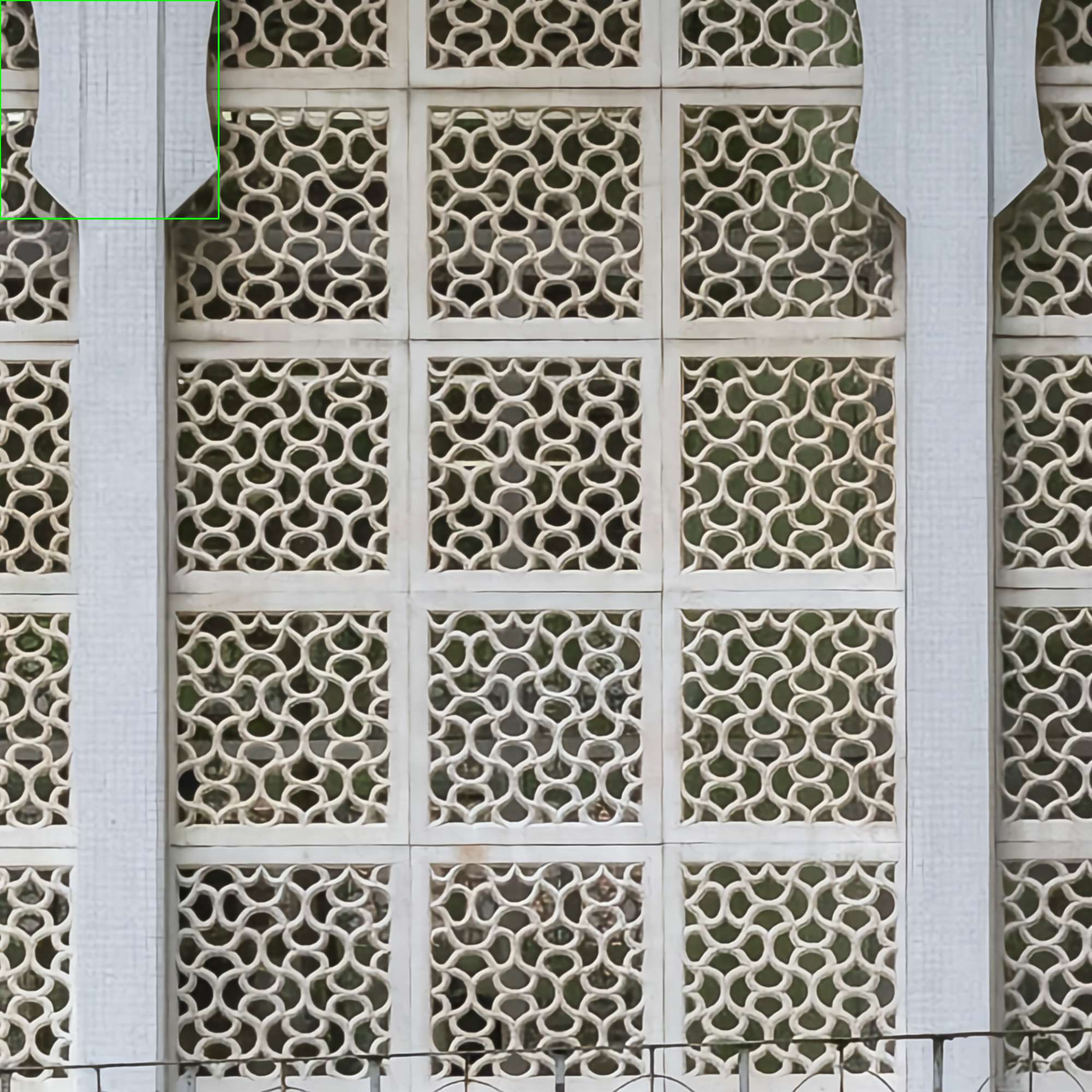}
			\sHSS
			&
			\sHSS
			\includegraphics[width=\CTS]{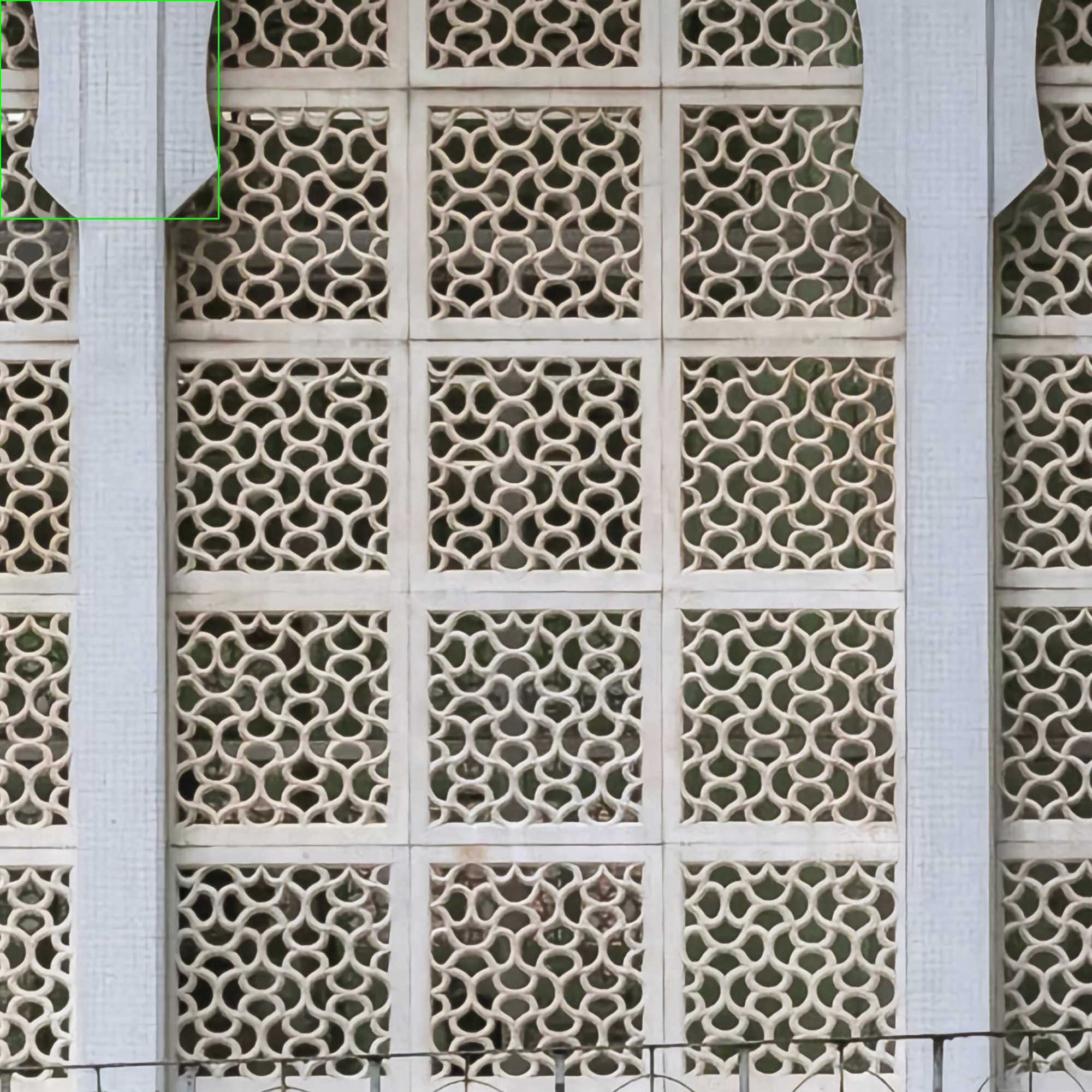}
			\hspace{-2mm}
			\\
			%			\vspace{1mm}
			\includegraphics[width=\CTS]{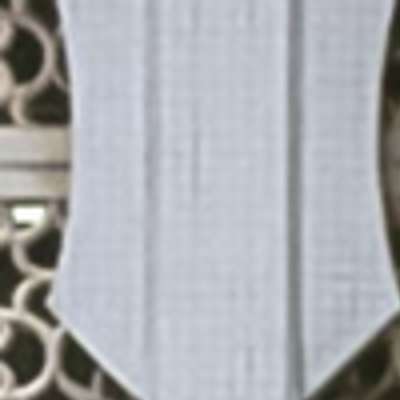}
			\sHSS
			&
			\sHSS
			\includegraphics[width=\CTS]{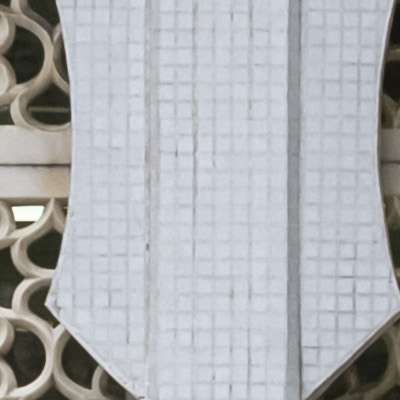}
			\sHSS
			&
			\sHSS
			\includegraphics[width=\CTS]{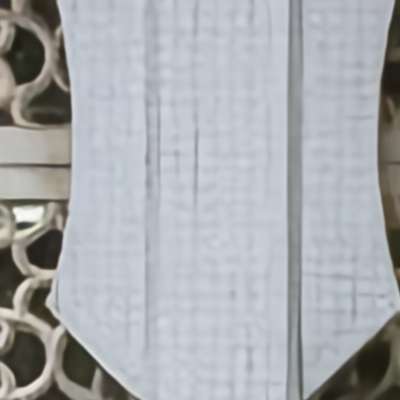}
			\sHSS
			&
			\sHSS
			\includegraphics[width=\CTS]{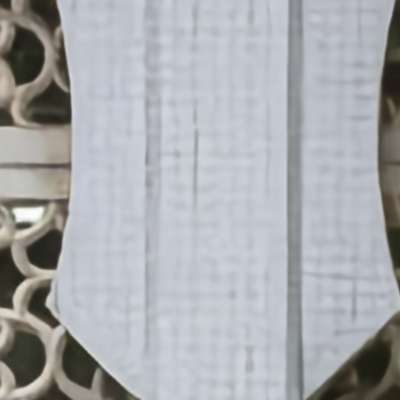}
			\hspace{-2mm}
			\\
			
			Low resolution & High resolution& LP-KPN & Ours \\
			& & PSNR 28.82 & PSNR 29.34 \\
			\\
			
			\includegraphics[width=\CTS]{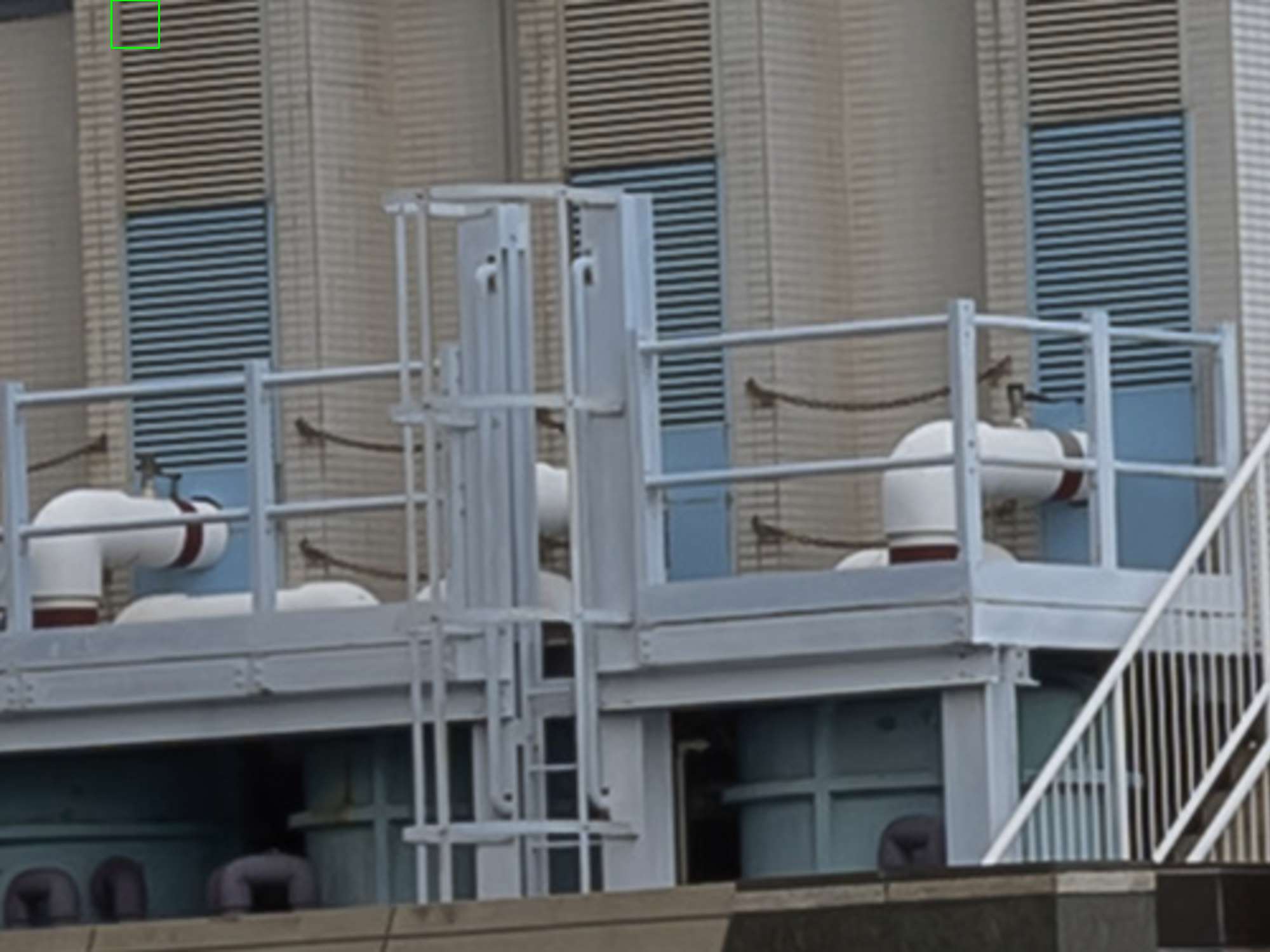}
			\sHSS
			&
			\sHSS
			\includegraphics[width=\CTS]{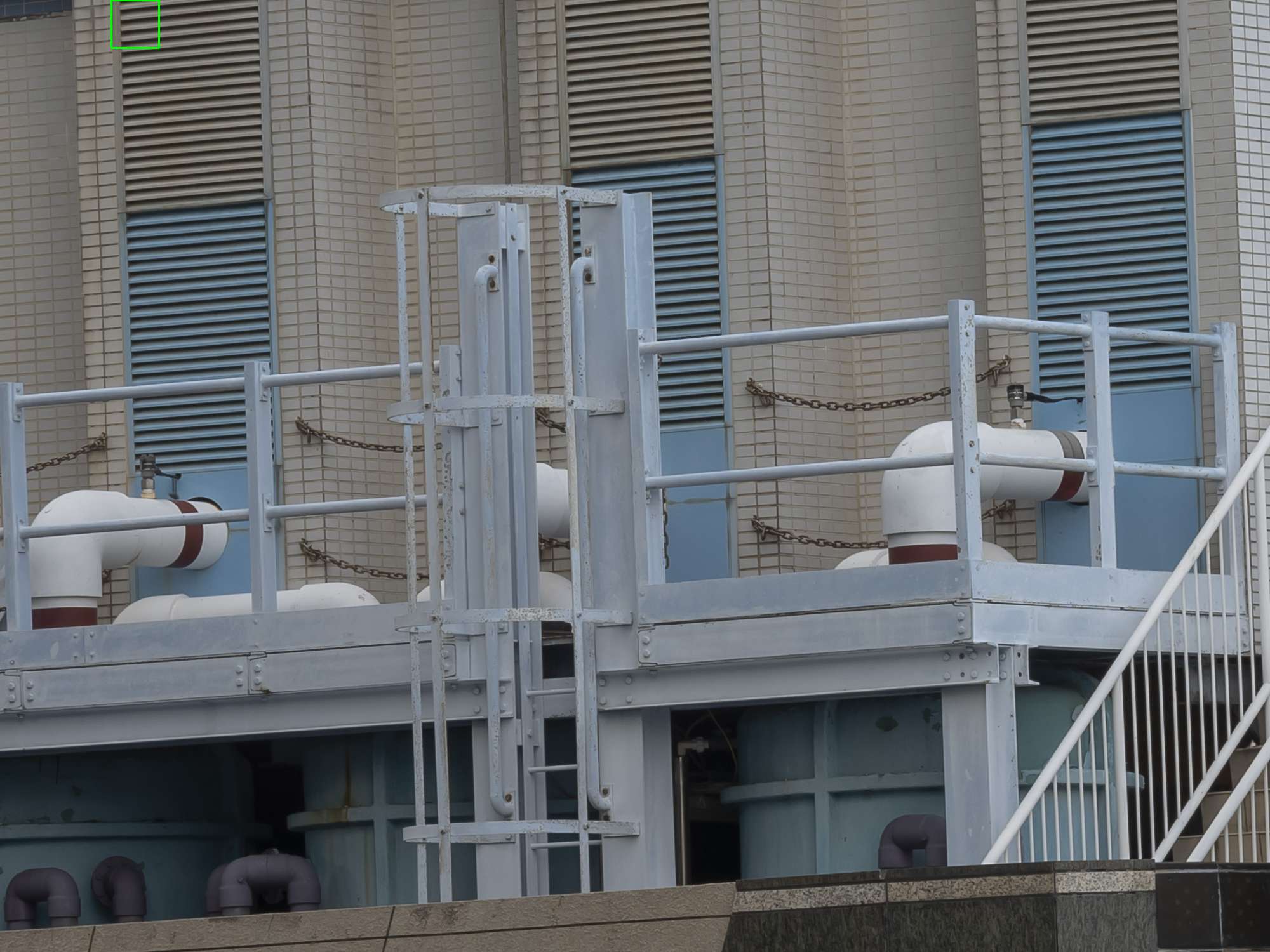}
			\sHSS
			&
			\sHSS
			\includegraphics[width=\CTS]{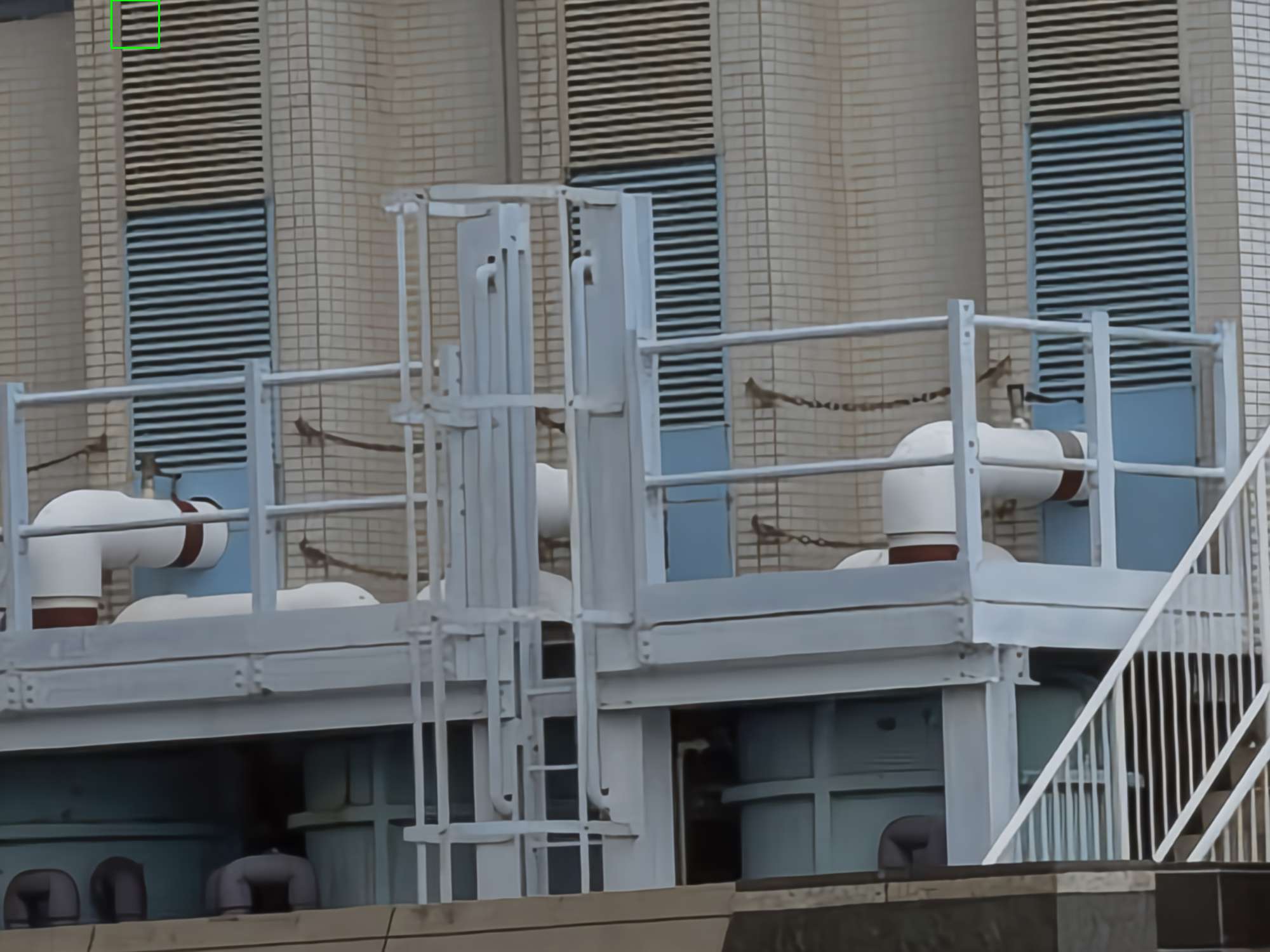}
			\sHSS
			&
			\sHSS
			\includegraphics[width=\CTS]{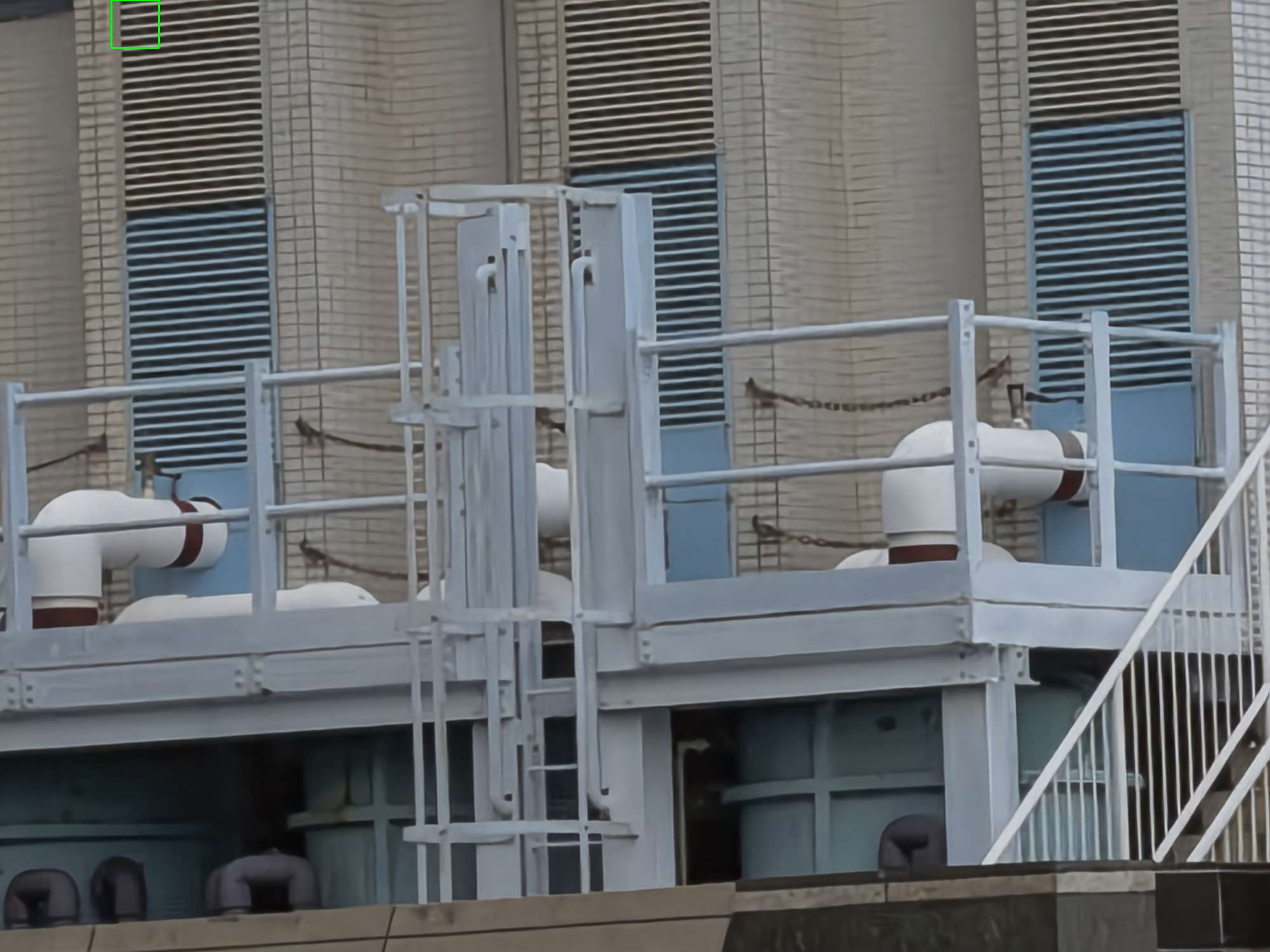}
			\hspace{-2mm}
			\\
			%			\vspace{1mm}
			\includegraphics[width=\CTS]{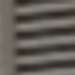}
			\sHSS
			&
			\sHSS
			\includegraphics[width=\CTS]{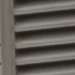}
			\sHSS
			&
			\sHSS
			\includegraphics[width=\CTS]{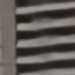}
			\sHSS
			&
			\sHSS
			\includegraphics[width=\CTS]{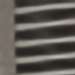}
			\hspace{-2mm}
			
			\\
			
			Low resolution & High resolution&  LP-KPN & Ours \\
			& & PSNR 26.24 & PSNR 26.89 \\
			%			\\
	\end{tabular}}
	\caption{Qualitative comparisons against LP-KPN. Although two methods deliver similar quantitative results, we consistently observe that LP-KPN suffers from strong artifacts. ACDA produce more faithful restoration results. We believe the adaptive convolutions operated in deep features help better capture image semantics, thus prevent over-sharp results with strong artifacts.}
	\label{fig:sr}
\end{figure}

We present qualitative results for super-resolution experiments on RealSR dataset in Figure~\ref{fig:sr}.

\section{Qualitative Results of Denoising}
\label{supp_dn}
We present qualitative results and comparisons in Figure~\ref{fig:denoise_more} and Figure~\ref{fig:denoise_comp}.

\newcommand{\STS}{0.22\linewidth}

\begin{figure}
	\centering
	\resizebox{0.9\linewidth}{!}{%
		%		\small
		\begin{tabular}{c c c | c c c}
			%		\centering
			%			\hspace{-2mm}
			\includegraphics[width=\STS]{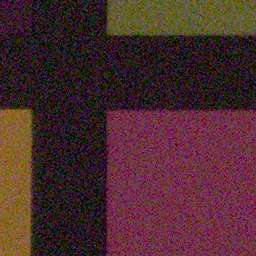}
			\sHSS
			&
			\sHSS
			\includegraphics[width=\STS]{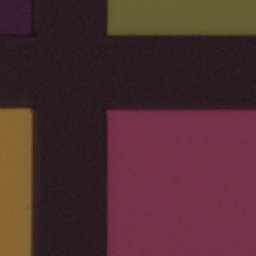}
			\sHSS
			&
			\sHSS
			\includegraphics[width=\STS]{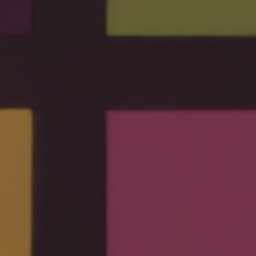}
			
			&
			
			\includegraphics[width=\STS]{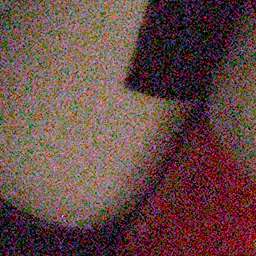}
			\sHSS
			&
			\sHSS
			\includegraphics[width=\STS]{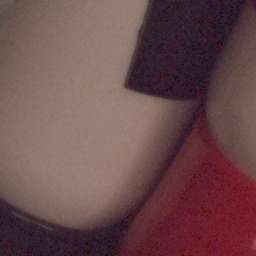}
			\sHSS
			&
			\sHSS
			\includegraphics[width=\STS]{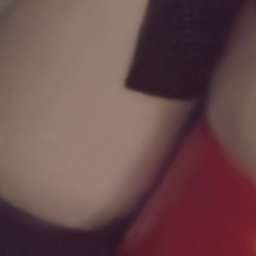}
			\hspace{-2mm}
			\\
			\\
			
			\includegraphics[width=\STS]{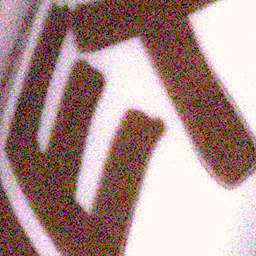}
			\sHSS
			&
			\sHSS
			\includegraphics[width=\STS]{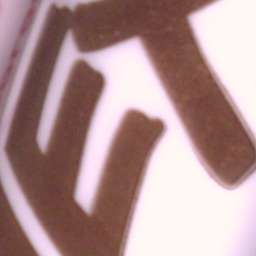}
			\sHSS
			&
			\sHSS
			\includegraphics[width=\STS]{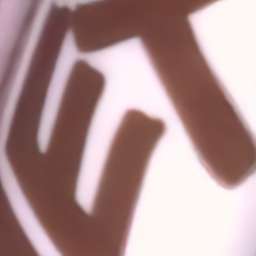}
			
			&
			
			\includegraphics[width=\STS]{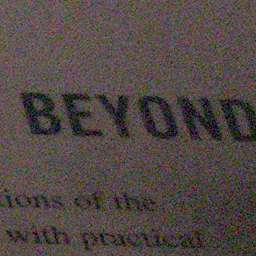}
			\sHSS
			&
			\sHSS
			\includegraphics[width=\STS]{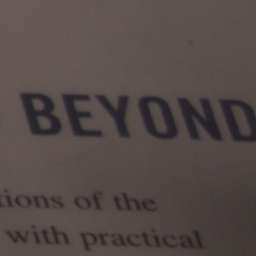}
			\sHSS
			&
			\sHSS
			\includegraphics[width=\STS]{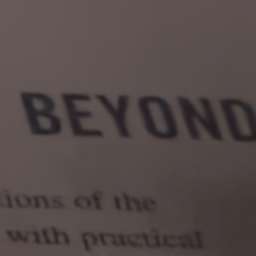}
			\hspace{-2mm}
			\\
			\\
			
			\includegraphics[width=\STS]{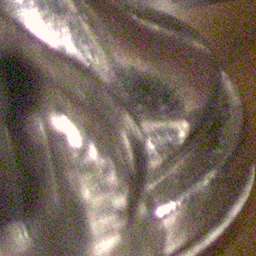}
			\sHSS
			&
			\sHSS
			\includegraphics[width=\STS]{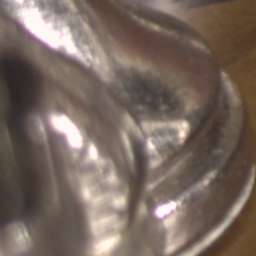}
			\sHSS
			&
			\sHSS
			\includegraphics[width=\STS]{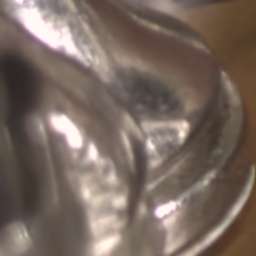}
			
			&
			
			\includegraphics[width=\STS]{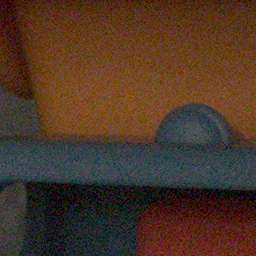}
			\sHSS
			&
			\sHSS
			\includegraphics[width=\STS]{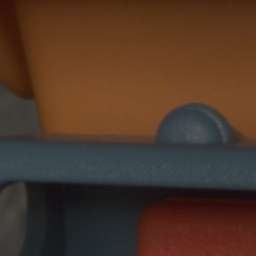}
			\sHSS
			&
			\sHSS
			\includegraphics[width=\STS]{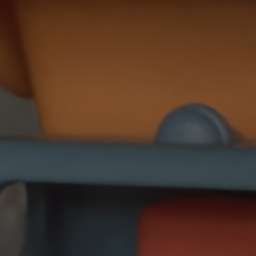}
			\hspace{-2mm}
			\\
			\\
			
			\includegraphics[width=\STS]{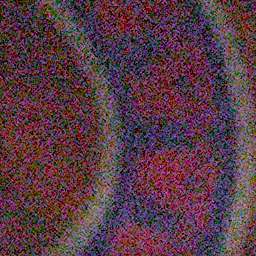}
			\sHSS
			&
			\sHSS
			\includegraphics[width=\STS]{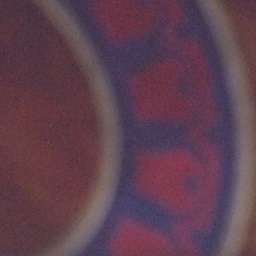}
			\sHSS
			&
			\sHSS
			\includegraphics[width=\STS]{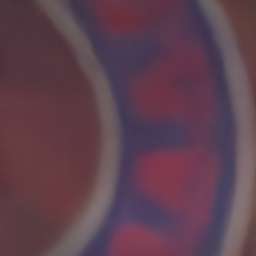}
			
			&
			
			\includegraphics[width=\STS]{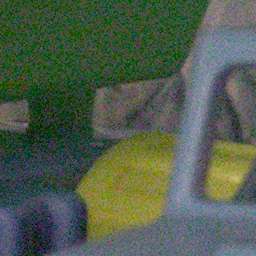}
			\sHSS
			&
			\sHSS
			\includegraphics[width=\STS]{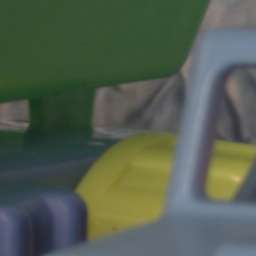}
			\sHSS
			&
			\sHSS
			\includegraphics[width=\STS]{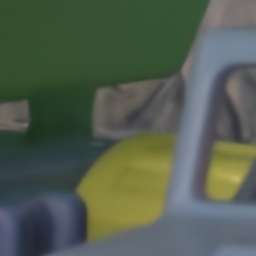}
			\hspace{-2mm}
			
			\\
			\\
			
			\includegraphics[width=\STS]{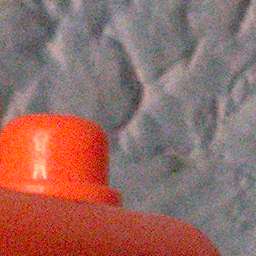}
			\sHSS
			&
			\sHSS
			\includegraphics[width=\STS]{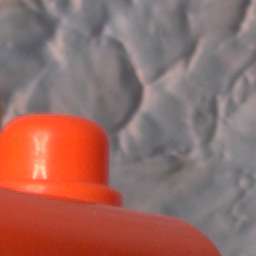}
			\sHSS
			&
			\sHSS
			\includegraphics[width=\STS]{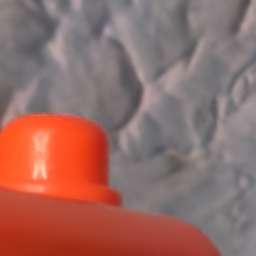}
			
			&
			
			\includegraphics[width=\STS]{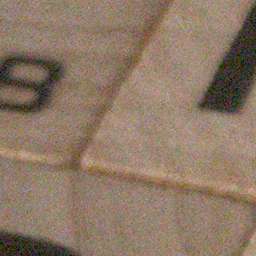}
			\sHSS
			&
			\sHSS
			\includegraphics[width=\STS]{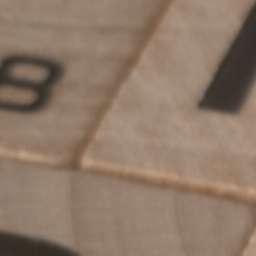}
			\sHSS
			&
			\sHSS
			\includegraphics[width=\STS]{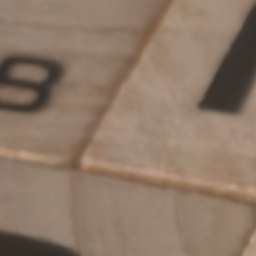}
			\hspace{-2mm}
			
	\end{tabular}}
	\caption{Qualitative results on SIDD. }
	\label{fig:denoise_more}
\end{figure}

\begin{figure}
	\centering
	\resizebox{\linewidth}{!}{%
		%		\small
		\begin{tabular}{c c c c c c c}
			%		\centering
			%			\hspace{-2mm}
			\includegraphics[width=\STS]{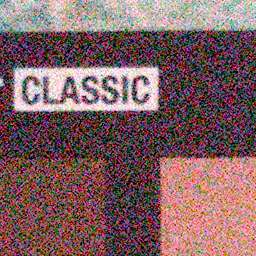}
			\sHSS
			&
			\sHSS
			\includegraphics[width=\STS]{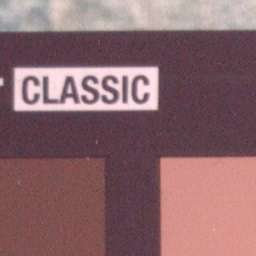}
			\sHSS
			&
			\sHSS
			\includegraphics[width=\STS]{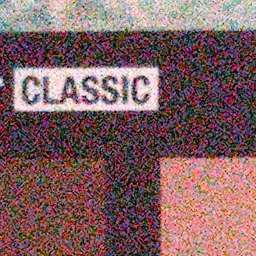}
			\sHSS
			&
			\sHSS
			\includegraphics[width=\STS]{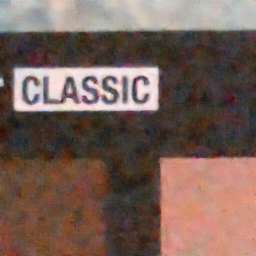}
			\sHSS
			&
			\sHSS
			\includegraphics[width=\STS]{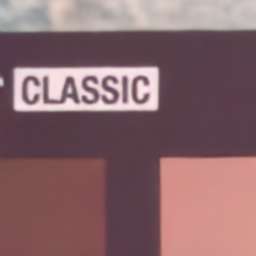}
			\sHSS
			&
			\sHSS
			\includegraphics[width=\STS]{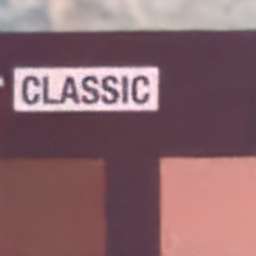}
			\sHSS
			&
			\sHSS
			\includegraphics[width=\STS]{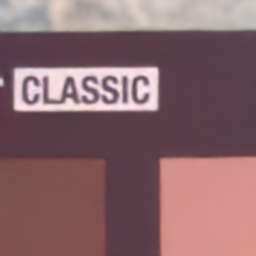}
			\hspace{-2mm}
			\\
			
			Noisy & Groundtruth & DnCNN & CBDNet & VDNet & ACDA + MSE & ACDA + VDNet \\
			& & PSNR 18.18 & PSNR 21.13 &PSNR  32.64 & PSNR 32.13 & PSNR 32.76 \\
			\\
			\\
			
			\includegraphics[width=\STS]{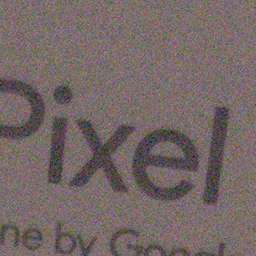}
			\sHSS
			&
			\sHSS
			\includegraphics[width=\STS]{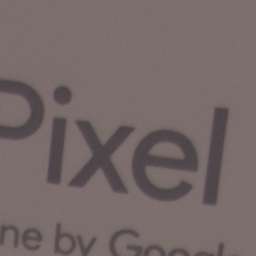}
			\sHSS
			&
			\sHSS
			\includegraphics[width=\STS]{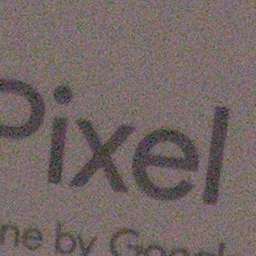}
			\sHSS
			&
			\sHSS
			\includegraphics[width=\STS]{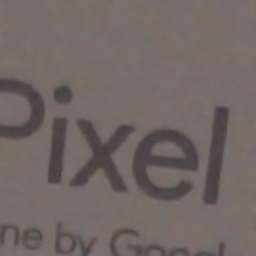}
			\sHSS
			&
			\sHSS
			\includegraphics[width=\STS]{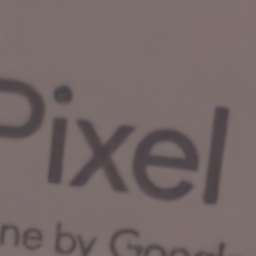}
			\sHSS
			&
			\sHSS
			\includegraphics[width=\STS]{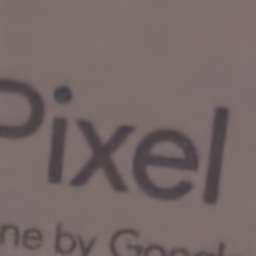}
			\sHSS
			&
			\sHSS
			\includegraphics[width=\STS]{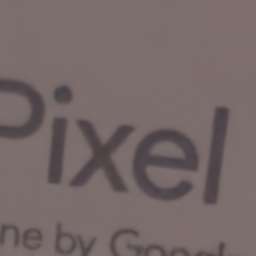}
			\hspace{-2mm}
			\\
			
			Noisy & Groundtruth & DnCNN & CBDNet & VDNet & ACDA + MSE & ACDA + VDNet \\
			& & PSNR 28.41 & PSNR  35.18 &PSNR  40.82 & PSNR 40.22 & PSNR 40.99 \\
	\end{tabular}}
	\caption{Qualitative comparisons against multiple methods on SIDD. }
	\label{fig:denoise_comp}
\end{figure}
\end{document}